%% file: ms.tex
\documentclass[12pt]{report}

\usepackage[a4paper, total={16cm, 23cm}]{geometry}
\usepackage{amsmath}
\usepackage{amssymb}
\usepackage{isomath}
\usepackage{graphicx}
\usepackage{longtable}
\usepackage{subcaption}
\usepackage[round]{natbib}
\usepackage[chapter]{algorithm}
\usepackage[noend]{algpseudocode}
\usepackage[hyphens]{url}
\usepackage[breaklinks]{hyperref}
\usepackage[section]{placeins}

\def\theAuthor{Marc Tanti}
\def\theMonth{March}
\def\theYear{2019}
\def\theTitle{On Architectures for Including Visual Information in Neural Language Models for Image Description}
\def\theSubtitle{}
\def\theFaculty{Institute of Linguistics and Language Technology}
\def\theUniversity{University of Malta}

\newcommand\ie{{\it i.e.\ }}

\newcommand\kw[1]{{\it #1}}

\newcommand\figuretitle[1]{{\centering\small\textbf{#1}\par\vspace{5pt}}}
\newcommand\chapterwithfootnote[2]{\chapter[#1]{#1\raisebox{.3\baselineskip}{\normalsize\footnotemark}}\footnotetext{#2}}

\DeclareMathOperator{\ReLU}{ReLU}
\DeclareMathOperator{\sig}{sig}
\DeclareMathOperator{\softmax}{softmax}
\DeclareMathOperator{\concat}{+\!\!+}
\DeclareMathOperator{\tensorprod}{\otimes}
\DeclareMathOperator{\dotprod}{\cdot}

\makeindex

\begin{document}
	\pagenumbering{gobble}
	
	\input{tex/chp0_preface1}
	
	\pagenumbering{roman}
	
	\input{tex/chp0_preface2}

	\pagenumbering{arabic}
	
	\input{tex/chp1_introduction}
	
	\input{tex/chp2_background}
	
	\input{tex/chp3_architectures}
	
	\input{tex/chp4_groundedness}
	
	\input{tex/chp5_transferlearning}
	
	\input{tex/chp6_conclusion}

	\bibliographystyle{plainnat}
	\bibliography{bibliography}
\end{document}

%% file: tex/chp0_preface1.tex

\begin{titlepage}
	\begin{center}
		\vspace*{1cm}
		\textbf{\theTitle}
		
		\vspace{0.5cm}
		\theSubtitle
		
		\vspace{1.0cm}
		\textbf{\theAuthor}
		
		\vspace{2cm}
		\theFaculty

		\vspace{0.5cm}
		\includegraphics[scale=0.6]{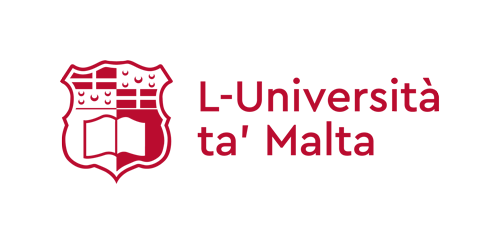}

		\vspace{0.5cm}
		{\large {\sc \theUniversity}}

		\vspace{1cm}
		{\Large \theMonth, \theYear}

		\vfill
		Submitted in partial fulfilment of the requirements for the degree of Ph.D.
	\end{center}
\end{titlepage}

\clearpage


\section*{Statement of originality}

I, the undersigned, declare that this is my own work, except where acknowledged and referenced.

\vspace{3cm}

\noindent
Student name: \theAuthor\\
Signature of student:

{\hspace{1.5cm}}

\noindent
Date: 14\textsuperscript{th} July 2019

\clearpage

%% file: tex/chp0_preface2.tex
\begin{abstract}
	A neural language model is a neural network that can be used to generate a sentence by suggesting probable next words given a partially complete sentence (a prefix). A recurrent neural network reads in the partial sentence and produces a hidden state vector which represents information about which words can follow. If a likely word from those suggested is selected and attached to the sentence prefix, another word after that can be selected as well, and so on until a complete sentence is generated in an iterative word by word fashion.
	
	Rather than just generating random sentences, a neural language model can instead be conditioned into generating descriptions for images by also providing visual information apart from the sentence prefix. This visual information can be included into the language model through different points of entry resulting in different neural architectures. We identify four main architectures which we call init-inject, pre-inject, par-inject, and merge.
	
	We analyse these four architectures and conclude that the best performing one is init-inject, which is when the visual information is injected into the initial state of the recurrent neural network. We confirm this using both automatic evaluation measures and human annotation.
	
	We then analyse how much influence the images have on each architecture. This is done by measuring how different the output probabilities of a model are when a partial sentence is combined with a completely different image from the one it is meant to be combined with. We find that init-inject tends to quickly become less influenced by the image as more words are generated. A different architecture called merge, which is when the visual information is merged with the recurrent neural network's hidden state vector prior to output, loses visual influence much more slowly, suggesting that it would work better for generating longer sentences.
	
	We also observe that the merge architecture can have its recurrent neural network pre-trained in a text-only language model (transfer learning) rather than be initialised randomly as usual. This results in even better performance than the other architectures, provided that the source language model is not too good at language modelling or it will overspecialise and be less effective at image description generation.
	
	Our work opens up new avenues of research in neural architectures, explainable AI, and transfer learning.
\end{abstract}

\clearpage


\section*{Dedication}

To my parents, for all the love they have given me.

\clearpage


\section*{Supervisor(s)}

\noindent
Dr. Albert Gatt,\\
Institute of Linguistics and Language Technology\\
University of Malta

\vspace{1cm}

\noindent
Prof. Kenneth P. Camilleri,\\
Faculty of Engineering\\
University of Malta

\clearpage


\section*{Acknowledgements}

This doctorate would not have been possible were it not for the many people who helped me along the way. I would first and foremost like to thank my two supervisors Dr.~Albert Gatt and Prof.~Kenneth P. Camilleri for their endless wisdom, encouragement, and time checking and correcting my work and ideas, as well as for their support to help me advance academically in my field. I would also like to thank the researchers who have discussed their work with me or taken the time to answer my questions: Barbara Plank, Sebastian Ruder, Junhua Mao, and Mert K{\i}l{\i}{\c{c}}kaya. Furthermore, I'd like to thank my examiners for their comments and the fruitful discussion: Dr.~Lonneke van der Plas, Dr.~Ing.~Reuben A. Farrugia, Prof.~Adrian Muscat, and Dr.~Raffaella Bernardi.

Support coming from academia is invaluable, but so is support from friends and family. I want to thank my mother Ruth, my father Joe, and my beautiful partner Analise for their constant love, support, and faith in me. I would also like to thank my best friend Robert for encouraging me to start my doctorate and for the many discussions we had about artificial intelligence. I also thank the persons responsible for the ENDEAVOUR scholarship scheme for supporting me throughout my studies.

Finally, I would like to thank the five annotators who took the time to manually evaluate my models. You have given this thesis a human touch thanks to your input.

\clearpage


\begin{figure}[H]
	\centering
	\includegraphics[scale=0.55]{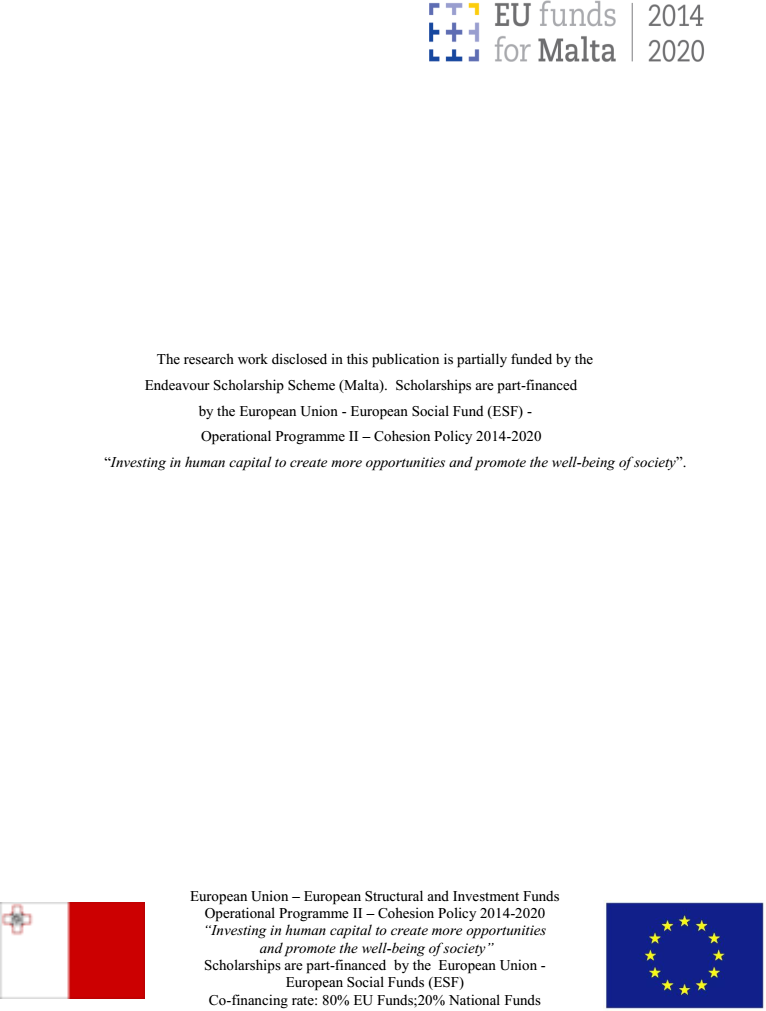}
\end{figure}

\clearpage


\setcounter{tocdepth}{2}
\tableofcontents
\listoffigures
\listoftables

\clearpage

%% file: tex/chp1_introduction.tex
\chapter{Introduction}
\clearpage

\section{Introduction}

Image caption generation is the task of automatically generating a description of an image using a computer. Text can be generated automatically using a kind of neural network called a neural language model that learns how to string words together into sentences after being trained on examples of well-written sentences. The problem is how to control the meaning of what gets generated, such as how to make the generated text a description of a specific image such that different images lead to different descriptions.

In order to do so, the neural language model needs to be influenced by visual information somehow. The question is, where should the visual information go in a neural language model? Existing image caption generators have included the visual information into the language model in different ways. Our objective is to make a systematic analysis of what these different ways are and their merits.

All the code used for the experiments in the thesis is available online\footnote{See: \url{https://github.com/mtanti/mtanti-phd}}.

\clearpage

\section{Motivations for the grounding problem}

The question of how to connect language to the world is called the grounding problem. In this case, we are interested in grounding language in perceptual data in order to be able to describe images. This is a fundamental AI challenge which has been addressed by cognitive scientists \citep{Harnad1990} and AI practitioners \citep{Roy2005} alike.

\begin{figure}
	\centering
	\includegraphics[scale=1.0]{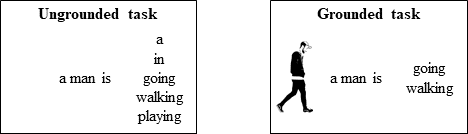}
	\caption{
		\label{fig:in_grounding}
		An illustration showing how the inclusion of vision helps with language tasks. Clipart image taken from \url{https://openclipart.org/detail/298220/man-walking}
	}
\end{figure}

Consider a system which predicts the next word in a half-finished sentence, such as for predictive text. The number of word candidates that can follow a partial sentence like ``a man is \dots'' is extremely large. Suppose however that apart from the partial sentence there is also an image which the sentence is supposed to be describing. In this case, the number of possible candidate words to follow the partial sentence will be drastically reduced since the image acts as a filter of which sentences are valid. This is shown in Figure~\ref{fig:in_grounding}.

Grounding serves as a form of guidance or goal to a text generation system in order to specify what it is that the system should describe or talk about. It also introduces several AI challenges to the task. For example, the level of abstraction needed to understand the image needs to be determined (is it a cluster of lines and colours, several individual people together, or a crowd?). Likewise, any non-salient parts of the image need to be ignored (do we need to describe the grass in the photo or is there more interesting information to focus on?). It is important to address these challenges in order to be able to communicate effectively to human beings.

There are several applications for visual grounding such as robots or chat bots that can describe what they are seeing from a camera, systems for describing images to people with impaired vision, summarising videos using short descriptions, providing extra information (in the form of images) to text understanding systems about potentially ambiguous sentences, and searching for images using text. Here are some specific fields in AI that require grounding in vision:

\begin{itemize}
	\item Visual question answering: generating answers for questions about images \citep{Antol2015,Fukui2016}.
	\item Multimodal translation: generating translations of sentences that are accompanied by an image \citep{Specia2016,Groenroos2018}.
	\item Grounded inference: classifying whether a sentence is true given a premise sentence and a premise image \citep{Vu2018} or just a premise image \citep{Xie2019,Lai2018}.
	\item Text-to-image generation: generating an image based on a textual description \citep{Xu2017,Mansimov2016}
\end{itemize}

We focus on image captioning in particular because it lends itself well to the analysis of visual grounding. It requires only a single input (the image) which will make analysis of the different possible ways to include the image into the neural network easier. With multiple inputs there would be many more possible combinations of input locations which would force us to only consider a subset.

\clearpage

\section{Research questions}

Our work investigates the following questions with regards to how to ground language in vision. First, the general questions:

\begin{enumerate}
	\item Given that neural language models require a representation of what has already been generated in order to work, should the visual information be encoded together with the words as a single mixed representation or should the visual and linguistic information be kept separate?
	
	\item Given that neural language models generate sentences one word at a time, should the visual information be introduced once at the beginning only or should it be reintroduced for every word being generated?
\end{enumerate}

More specific questions we will be trying to answer:

\begin{enumerate}
	\setcounter{enumi}{2}
	\item What are the merits of different neural architectures that include visual information at different locations in the neural language model? Is there a best architecture?
	
	\item Are there architectures that are influenced less by visual information than others? Furthermore, are there architectures that become influenced less by visual information as more words in the sentence are generated?
	
	\item In an architecture that keeps its visual and linguistic parts separate, is it possible to train the parts separately?
\end{enumerate}

Our goal is not to reach a new state of the art image caption generator, but to analyse the different possible architectures to condition a language model on visual information. We also avoid using the now-commonplace attention mechanism which gives better results but which limits the kind of architectures that can be used (attention mechanisms are explained in Subsection~\ref{sec:attention_mechanisms}). Our work here will set the foundation for further analysing more complex caption generators.

\clearpage

\section{Roadmap}

The remainder of this thesis is structured as follows. The next chapter will give an explanation of terms and methods related to neural image caption generators, starting from neural networks, then moving to neural language models, and finally image caption generators. It also establishes four possible main image caption generator architectures found in the literature which will be analysed in the remainder of the thesis.

The three chapters after that will be about three sets of experiments, each designed to answer one of the above three specific research questions. These are as follows:
\begin{itemize}
	\item Chapter 4 deals with evaluating the performance of each architecture using different evaluation measures (research question 3).
	
	\item Chapter 5 deals with analysing the amount of visual influence on each architecture (research question 4).
	
	\item Chapter 6 deals with evaluating the viability of separately training the visual and linguistic parts of the architectures that can be trained in this way (research question 5).
\end{itemize}

Finally we give our conclusions and future work in the last chapter.

\clearpage

\section{Publications related to this thesis}

We have already published preliminary work that was later improved for this thesis.

\begin{itemize}
	\item The work in chapter 4 has been published in the International Conference on Natural Language Generation \citep{Tanti2017} and in the Natural Language Engineering Journal \citep{Tanti2018}.
	
	\item The work in chapter 5 was published in the European Conference on Computer Vision \citep{Tanti2019b}.
	
	\item Finally the work in chapter 6 was not yet published at the time of writing this thesis but was put on Arxiv as a pre-print paper \citep{Tanti2019}.
\end{itemize}

\clearpage

%% file: tex/chp2_background.tex
\chapter{Background}
\clearpage

\section{Introduction}

In this chapter we give a general introduction to all the necessary background topics needed to understand this thesis. We will go from the most general topic to the most specific, highlighting important keywords using \kw{italic type}.

We start from a general explanation of how neural networks work. This allows us to establish the terminology and notation that will be used throughout the rest of the thesis.

This is followed by how neural networks can encode sentences. Here we explain that recurrent neural networks encode sequences of words rather than generate them, which is important to understand how one of the image caption generator architectures works.

The next section deals with how to predict the next word in an encoded partial sentence in order to generate sentences using neural language models. This is to prepare the ground for caption generators, as caption generators are not an indivisible entity but are a neural language model with an image added to it. This section also explains that there are two ways a neural language model can be used: a basic method and an efficient method. The most commonly used method is the efficient method and is the one which we use in this thesis but the other method is more powerful as will be discussed in the last section of this chapter.

The last section deals with how the sentences being generated can be controlled into being captions for images. This is where we give a review of the literature on caption generators and is the main subject of the thesis.

\clearpage

\section{Artificial neural networks}

\kw{Deep learning} \citep{LeCun2015} is a machine learning technique that reduces the problem of manual representation engineering as it can work directly on almost raw data. This technique consists of using artificial neural networks with multiple hidden layers, where each layer transforms a representation of the input into another representation until it becomes an output. \kw{Artificial neural networks} are functions that take in a tensor of numeric inputs and return another tensor of numeric outputs. They work using many neural units which are defined as follows:
\begin{align}
	y = f(\vectorsym{x} \dotprod \vectorsym{w} + b)
\end{align}
where $y$ is the scalar output of the neural unit, $\vectorsym{x}$ is the inputs vector to the neural unit, $\vectorsym{w}$ is the \kw{weights} vector, $b$ is the \kw{bias} scalar, $f$ is the non-linear \kw{activation function}, and $\dotprod$ is the dot product operator. Changing the weights and bias, collectively called \kw{parameters}, will change how the neural unit maps inputs to output.

Activation functions can be squashing functions such as \kw{sigmoid}:
\begin{align}
\sig(x) = \frac{1}{1 + e^{-x}}
\end{align}
or \kw{hyperbolic tangent}:
\begin{align}
\tanh(x) = \frac{e^{x} - e^{-x}}{e^{x} + e^{-x}}
\end{align}
or they can be unbounded functions such as the \kw{rectified linear unit}:
\begin{align}
	\ReLU(x) = \max(x, 0)
\end{align}

\begin{figure}
	\centering
	\includegraphics[scale=0.7]{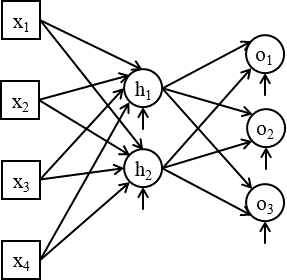}
	\caption{
		\label{fig:bg_neuralnet}
		An illustration of a neural network with a single hidden layer. The first column is the input layer which takes in vector $\vectorsym{x}$ of size 4. The second column is the hidden layer which transforms $\vectorsym{x}$ into vector $\vectorsym{h}$ of size 2. The last column is the output layer which transforms vector $\vectorsym{h}$ into vector $\vectorsym{o}$ of size 3. Squares are inputs, circles are neural units, connected arrows are weights, and dangling arrows are biases.
	}
\end{figure}

Computing multiple neural units for the same inputs leads to a vector of outputs. Such a group of neural units is called a \kw{layer}. Several layers can be used in series where the output vector of one layer serves as input to another layer, each transforming the vector of numbers into a different representation until the last layer makes the final transformation that is the desired output. The last layer is called the \kw{output layer} whilst the other layers are called \kw{hidden layers}. This is illustrated in Figure~\ref{fig:bg_neuralnet}.

We can extend the previous definition of a neural unit into the definition of a layer, specifically a \kw{fully connected feed-forward layer}, as follows:
\begin{align}
	\vectorsym{y} = f(\vectorsym{x} \tensorprod \matrixsym{W} + \vectorsym{b})
\end{align}
where $\vectorsym{y}$ is the output vector of the layer, $\matrixsym{W}$ is the weights matrix, $\vectorsym{x}$ is the inputs vector of the layer, $\vectorsym{b}$ is the bias vector, $f$ is the activation function, and $\tensorprod$ is the tensor product operator.

If the output of the neural network is to be used to decide on the class of the input, such as the type of object shown in an image or the topic of a document, then it is usually required that a probability for each class (a probability distribution) is returned in order to decide how likely the input is to belong to a particular class. This can be accomplished using a softmax function as an activation function in the output layer. \kw{Softmax} is defined as follows:
\begin{align}
	\softmax(\vectorsym{x})_i = \frac{e^{\vectorsym{x}_i}}{\sum_{j} e^{\vectorsym{x}_j}}
\end{align}
where $\vectorsym{x}$ is a vector of \kw{logits} (scores) for each class and $\softmax(\vectorsym{x})_i$ is the probability of the $i$\textsuperscript{th} class given by the softmax function. The result of the softmax is a vector of probabilities that sum to 1, one for each class.

An interesting feature of neural networks is their expressiveness. It is possible to prove that, given at least a single hidden layer with a sufficient number of neural units, a neural network is a universal approximating function, that is, it can be assigned parameters that make it approximate any particular function \citep{Hornik1989}. Furthermore, the number of neural units in the hidden layer required to express a function is exponentially larger than that of a neural network with two hidden layers \citep{Eldan2016}. The insight that adding more hidden layers makes neural networks more efficient at expressing functions led to deeper and deeper neural networks with more and more layers, which resulted in the field known as deep learning.

Whilst genetic algorithms and simulated annealing can be used to find the right parameters to make a neural network compute the desire function, the most common way to learn the parameters is by using gradient descent optimisation. Gradient-based techniques are expected to perform poorly on non-convex optimisation problems, such as neural network parameter optimisation, but it turns out that even if the parameter updates of gradient descent will lead to a local minimum, we can expect that most minima will be similar to each other given a large neural network, so this is not a significant problem \citep{Choromanska2015}. Gradients for a multi-layer neural network can be efficiently computed using the backpropagation algorithm \citep{Rumelhart1986}.

Of course, the more expressive a model is, the less likely it is to compute something meaningful to a person after training and the more likely it is to simply memorise the training set \citep{Zhang2016}, a problem known as \kw{overfitting}. In order to reduce the extent of overfitting, many \kw{regularisation} techniques were developed \citep[Chapter~7]{Goodfellow2016}, the most straightforward of which is called early stopping. \kw{Early stopping} is a technique where after every parameter update, the neural network is evaluated on a \kw{validation set}, which is disjointed from the training set and the test set. As soon as it stops improving on the validation set, training is stopped as this is evidence that the neural network has started memorising the training set rather than learning a useful function. Another example of regularisation involves constraining the possible values of the parameters of the neural network such as by weight decay (or L2 regularisation). \kw{Weight decay} encourages the neural network weights to be as small as possible by minimising the sum of the squares of all the weights in the neural network. This reduces the extent to which a neural unit can put too much importance on a single other neural unit in the previous layer. \kw{Dropout} \citep{Srivastava2014} is a more complex form of regularisation which approximates the effect of three different regularisation techniques:
\begin{itemize}
	\item averaging together the outputs of separately trained neural networks (ensembling),
	\item all of which share many parameters together (parameter sharing),
	\item and all of which are trained on different subsets of the training set (bagging).
\end{itemize}
This approximation is achieved by simply ignoring a random subset of neural units in a layer for each training item during training, forcing each neural unit to avoid relying on any other single neural unit.

As has been stated already, with deep learning the manual engineering needed on the input prior to being fed to the neural network is minimal. For example in text, the engineering decisions that are taken regarding the input are usually things like whether to process at the word level or the character level, the vocabulary size, and the maximum sentence length. Processing text after these decisions have been taken is relatively straight forward, as is elaborated upon in the next section.

\clearpage

\section{Deep learning approaches to encoding sentences}

To encode a sentence means to transform it into an abstracted representation that is easier to work with, which in the case of deep learning usually means transforming it into a vector or matrix. Traditionally sentences would be encoded using bag of words models in order to represent their meaning using vectors consisting of the number of times they use certain words or n-grams. With neural networks, there is no need for manually designing algorithms to encode sentences as this can be done automatically.

\subsection{Word embeddings}

Unless the sentence is viewed as a string of characters instead of words, the first step in encoding a sentence is usually to encode the individual words in a sentence, which in turn requires that a finite vocabulary of useful words be selected. The vocabulary usually consists of the most frequent $n$ words in the training set plus some pseudo-words that are added to the vocabulary for convenience. For example, any word in the input text that is not in the vocabulary is replaced with a pseudo-word called an \kw{unknown token}. Typically, another pseudo-word is introduced in the vocabulary called a \kw{pad token} which is used at the beginning or end of each sentence in the text in order to make all the sentences equal in length (equal to a determined maximum length).

\begin{figure}
	\centering
	\includegraphics[scale=0.8]{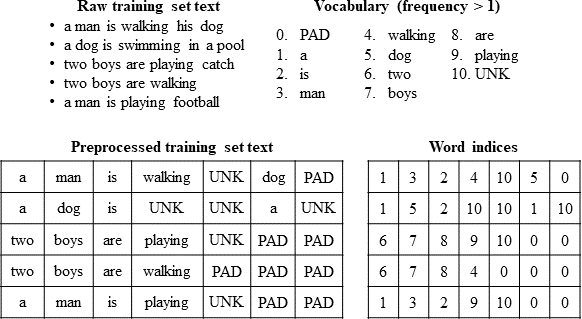}
	\caption{
		\label{fig:bg_text_preprocessing}
		How text is prepared before being used in a neural network. First, a vocabulary is extracted from the raw text, in this case only words that occur more than once make it to the vocabulary. The text is then processed to include only vocabulary words with any other words replaced by the unknown token. Sentences are also padded with pad tokens at the end so that all sentences are the same length. Finally, words are replaced with their indices in the vocabulary in order to produce a numerical matrix.
	}
\end{figure}

The words in the vocabulary now need to be replaced with numerical vectors since neural networks can only work with numbers. These vectors, called \kw{word embeddings}, should encode some representation of meaning about each word such that words that are used similarly in the training set will have similar vectors. The first step to doing this is to represent every word with an index which is its position in the vocabulary list. The first word in the vocabulary is replaced with the number 0, the second with 1, and so on, resulting in a regularly shaped numerical matrix as illustrated in Figure~\ref{fig:bg_text_preprocessing}.

\begin{figure}
	\centering
	\includegraphics[scale=0.6]{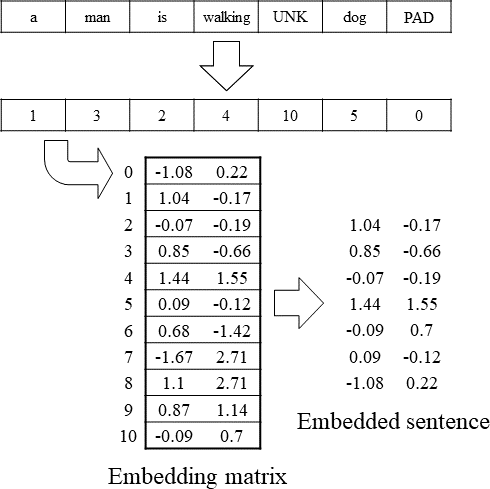}
	\caption{
		\label{fig:bg_embedding}
		An illustration showing how words in a sentence are converted into vectors using an embedding matrix.
	}
\end{figure}

Next, the neural network is fitted with an \kw{embedding matrix} which is a matrix where each row is the vector representing a different word in the vocabulary. Each word index is replaced with its corresponding row vector in the matrix, thus replacing the words with vectors. This is illustrated in Figure~\ref{fig:bg_embedding}. The numbers in the embedding matrix are optimised freely with the rest of the neural network parameters in order to find the optimal word embeddings according to the task being solved by the neural network.

Word embeddings are also useful in their own right. Once an embedding matrix has been trained, it is possible to transfer it to another neural network as is without needing to train it further, transferring the captured meaning along with it. This is so useful that surrogate tasks have been created just to be able to extract a useful embedding matrix that can be used for different tasks. An example of these off-the-shelf embedding matrices is word2vec \citep{Mikolov2013} where the surrogate task is to predict the words that are adjacent to a particular word in a sentence. Another example is GloVe \citep{Pennington2014} where the surrogate task is to make the dot-product of any two word vectors approximately equal to the weighted logarithm of the number of times that one of the words occurs in the context of the other in a corpus.

\subsection{Non-recurrent methods to sentence encoding}
\label{sec:nonrnn_sentence_encoding}

Inputting sentences is slightly more challenging than inputting words. The neural networks shown up to now can only accept fixed length vectors as inputs whilst sentences have a variable length.

\begin{figure}
	\centering
	\includegraphics[scale=0.6]{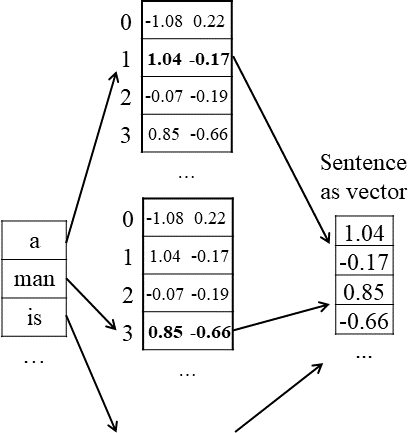}
	\caption{
		\label{fig:bg_sent_feedforward}
		An illustration showing the naive way to input a sentence into a neural network by concatenating the embedded words together into one vector and passing it to the next layer. Note that the embedding matrices for each word position need not be equal and can be optimised separately.
	}
\end{figure}

The naive solution is to decide on a maximum sentence length and then force all sentences to be that length. If a sentence is shorter than the maximum then pad words are used to fill in missing words. If the sentence is longer than the maximum then only a part of it can be used. After all words have been embedded they are then concatenated into a single vector. The embedding matrix can either be shared among all word positions in the sentence or a different one can be used for each position. An example of this is illustrated in Figure~\ref{fig:bg_sent_feedforward}. Not only is this wasteful, since it would need an excessive amount of parameters, the maximum sentence length is likely to be much larger than necessary most of the time, making a lot of parameters unnecessary most of the time.

One easy way to get around the large number of parameters is to simply average all the word vectors of the sentence together, creating a sentence vector that is the centroid of the words. This is called a deep averaging network. Its main flaw is that it ignores the order of the words, since the averaging operation is commutative, but there are experiments which show that it can work better than some more sophisticated ways of representing sentences \citep{Iyyer2015}.

\begin{figure}
	\centering
	\includegraphics[scale=0.6]{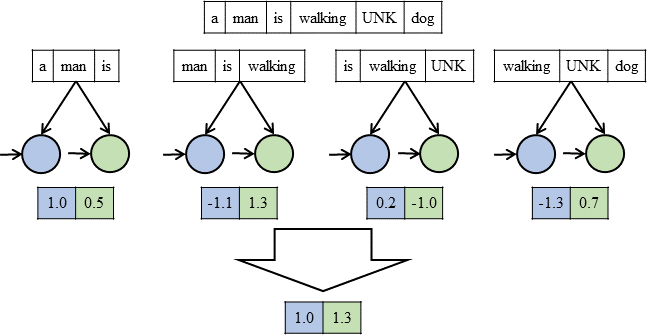}
	\caption{
		\label{fig:bg_1dconv}
		An illustration showing the convolutional and pooling operations of a one-dimensional convolutional neural network encoding a sentence. The filter works on three-word windows producing a feature vector of size two, and pooling is max-pooling. Each of the four displayed layers have the same parameters.
	}
\end{figure}

Another way to encode sentences is by using one-dimensional convolutional neural networks \citep{Collobert2011}. A convolutional neural network is more typically associated with encoding images and would need to be at least two-dimensional in order to process the rectangular shape of the image (more on this in Subsection~\ref{sec:convnets}). When applied to a sentence, there is only one dimension to process. A sentence is encoded by passing a sliding window over the word vectors, for example every three consecutive words, concatenating these word vectors together and passing them through a small fully connected layer, called a filter. Every window of words is passed through the same filter and transformed in the same way into separate feature vectors. Once the sentence is mapped into a collection of feature vectors, a pooling operation is performed where the collection of vectors is squashed into a single vector, usually by keeping only the maximum value of each corresponding element in the vectors. This pooled vector is the sentence encoding. The pooling allows the sentence to be long without requiring huge vector representations. This process is illustrated in Figure~\ref{fig:bg_1dconv}. Several filters can be used in sequence in order to extract higher level information from the previous feature vectors, creating a hierarchical encoding. Also, several filters can be used in parallel in order to extract different types of information and the resulting encodings can then be concatenated into a single vector encoding.

\subsection{Recurrent neural networks}
\label{sec:recurrent_neural_networks}

When encoding a variable length sentence, it is important to be able to transform a sentence of any length into a fixed vector size as this is what neural networks typically operate on. One-dimensional convolutional neural networks do this by using the pooling operator which squashes the many fixed-size filter vectors into a single fixed-size vector; the pooling operator however is not a trainable operator and might not be the optimal way to mix the vectors together given a particular task and dataset. \kw{Recurrent neural networks} (RNN) do not have this problem.

\begin{figure}
	\centering
	\includegraphics[scale=0.6]{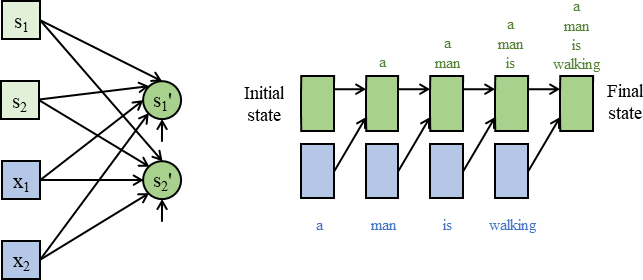}
	\caption{
		\label{fig:bg_srnn}
		An illustration showing what a simple RNN is made up of and how it processes a sequence of inputs (a sentence in this case). Given a state vector $s$ and an input vector $x$, the RNN layer will combine them to produce a new state vector $s'$ that can in turn be fed back to the RNN layer with another input. A default initial state is used for the first input, such as an all-zeros vector. Each state contains information describing all the inputs visited before that point (a prefix of the sentence) with the final state representing the full sentence.
	}
\end{figure}

An RNN is a neural network layer that maintains a memory, called a \kw{hidden state vector} or simply \kw{state}. This memory is a fixed vector that can be used to remember past inputs. In other words, the contents of this vector can be treated as a representation of a sequence of inputs. As a neural network layer, all the RNN does is take in two inputs: the new input to remember, such as a word in a sentence, and the previous state. Its output would then be the new state. Given that the state is produced from the new input and the previous state, this means that the RNN can be chained so that the new state is used as an input to itself. This is illustrated in Figure~\ref{fig:bg_srnn}. Of course the chaining implies that there has to be an \kw{initial state} that does not come from the RNN. This initial state can be a constant, such as an all-zeros vector, or it can come from some other neural network layer. What is important is that it is the same layer with the same parameters that is used for every word and so the sentence can be very long whilst the neural network remains the same size. After each word, the state represents information about all the words seen up to that point, which form a prefix of the sentence. The \kw{final state} will represent the full sentence. This type of RNN is called a \kw{simple recurrent neural network}, also known as an Elman network \citep{Elman1990}.

Formally, a simple RNN is defined as follows:
\begin{align}
	\matrixsym{S}_t = f((\matrixsym{S}_{t-1} \concat \matrixsym{X}_{t}) \tensorprod \matrixsym{W} + \vectorsym{b})
\end{align}
where $\matrixsym{S}$ is a matrix of state vectors, $\matrixsym{X}$ is a matrix of input vectors, $\matrixsym{S}_t$ is the new state vector after $t$ inputs, $\matrixsym{S}_{t-1}$ is the previous state vector, $\matrixsym{S}_0$ is the initial state vector, $\matrixsym{X}_{t}$ is the new input vector, $\matrixsym{W}$ is the weight matrix, $\vectorsym{b}$ is the bias vector, $f$ is the activation function, $\concat$ is the vector concatenation operator, and $\tensorprod$ is the tensor product operator.

The parameters of an RNN can be trained by `unrolling' it, that is, replicating the layer that processes one input for as many times as the length of the longest sequence in the training set. The replicated layers would be no different than those of the fully connected feed-forward neural networks described before. In order to be able to roll the replicated layers back into one, the parameters of the layers need to be kept equal. This is done using the backpropagation through time algorithm \citep{Werbos1990}, which is an extension of the backpropagation algorithm, to efficiently handle this case.

An interesting feature about RNNs is that they are equivalent to a universal Turing machine, in the sense that they are expressive enough to simulate any computer program \citep{Siegelmann1995}. This fact should not distract from the reality that learning the parameters to give complex behaviour to an RNN is challenging. Simple RNNs suffer from the vanishing and exploding gradient problem \citep{Hochreiter2000} whose main consequence is that the neural network is only able to learn short sequences. As longer and longer sequences are used, the gradient of the final state with respect to the early inputs in the sequence can grow or shrink exponentially, leading to exploding or vanishing gradients during training which will either cause numerical overflows or early inputs getting completely ignored during training. These problems can be mitigated by using ReLU as an activation function, clipping the gradients so that they are never greater than a certain value \citep{Pascanu2013}, or by initialising the weights matrix of the state to the identity matrix \citep{Le2015}.

Many practitioners, including us, opt to instead use a more sophisticated type of RNN such as the \kw{long short-term memory} (LSTM) \citep{Hochreiter1997} or the \kw{gated recurrent unit} (GRU) \citep{Chung2014}. These RNNs solve the problem of vanishing and exploding gradients by changing the equation to produce the next state to one which more easily propagates the gradient across long sequences. The main template for the equation is as follows:
\begin{align}\label{eqn:rnn_template}
	\matrixsym{S}_{t} = \tanh((\matrixsym{S}_{t-1} \concat \matrixsym{X}_{t}) \tensorprod \matrixsym{W} + \vectorsym{b}) + \matrixsym{S}_{t-1}
\end{align}

Both the LSTM and the GRU implement the above template. They also include gating functions, that is, additional neural layers that are used to decide whether an activation value in some other layer, the gated layer, should be allowed through or be replaced with a zero. This is done by multiplying the activation values of the gated layer by a fraction between $0$ and $1$, which in turn is produced by the gate layer by using $\sig$ as an activation function.

The LSTM uses two hidden state vectors called the hidden state and the cell state. Usually it is the hidden state that is treated as representing the sequence. The LSTM is defined as follows:
\begin{align}
	\matrixsym{G^f}_t &= \sig((\matrixsym{H}_{t-1} \concat \matrixsym{X}_{t}) \tensorprod \matrixsym{W^f} + \vectorsym{b^f}) \\
	\matrixsym{G^i}_t &= \sig((\matrixsym{H}_{t-1} \concat \matrixsym{X}_{t}) \tensorprod \matrixsym{W^i} + \vectorsym{b^i}) \\
	\matrixsym{G^o}_t &= \sig((\matrixsym{H}_{t-1} \concat \matrixsym{X}_{t}) \tensorprod \matrixsym{W^o} + \vectorsym{b^o}) \\
	\matrixsym{C}_t &= \matrixsym{G^i}_t \odot \tanh((\matrixsym{H}_{t-1} \concat \matrixsym{X}_{t}) \tensorprod \matrixsym{W^c} + \vectorsym{b^c}) + \matrixsym{G^f}_t \odot \matrixsym{C}_{t-1} \\
	\matrixsym{H}_t &= \matrixsym{G^o}_t \odot \tanh(\matrixsym{C}_t)
\end{align}
where $\matrixsym{H}$ is a matrix of hidden state vectors, $\matrixsym{C}$ is a matrix of cell state vectors,  $\matrixsym{X}$ is a matrix of input vectors, $\matrixsym{G^\alpha}$ is a matrix of gate vectors, $\matrixsym{H}_t$ is the hidden state vector after $t$ inputs, $\matrixsym{H}_0$ is the initial hidden state vector, $\matrixsym{C}_t$ is the cell state vector after $t$ inputs, $\matrixsym{C}_0$ is the initial cell state vector, $\matrixsym{G^\alpha}_t$ is a gate vector after $t$ inputs, $\matrixsym{W^\alpha}$ is a weights matrix, $\vectorsym{b^\alpha}$ is a bias vector, $\odot$ is the element-wise vector multiplication operator, $\concat$ is the vector concatenation operator, and $\tensorprod$ is the tensor multiplication operator. The template in Equation~\ref{eqn:rnn_template} is used on $\matrixsym{C}_t$. It is gated at the two terms of the addition: $\matrixsym{G^f}$ is called the forget gate and gates parts of the previous cell state vector in order to choose whether a piece of state information should be propagated forward, and $\matrixsym{G^i}$ is called the input gate and gates parts of the new vector to add to the previous cell state vector. These gates are not based on the cell state directly (although there is a version of the LSTM that does this called a cell peep-hole LSTM \citep{Gers2002}) but only on the current input $\matrixsym{X}_t$ and the hidden state $\matrixsym{H}_t$. $\matrixsym{H}_t$ is a second state that is derived from the cell state and is just the $\tanh$ of the cell state gated by $\matrixsym{G^o}_t$, called the output gate.

A simplified version of the LSTM is the GRU, which uses only one state vector. It is defined as follows:
\begin{align}
	\matrixsym{G^f}_t &= \sig((\matrixsym{S}_{t-1} \concat \matrixsym{X}_{t}) \tensorprod \matrixsym{W^f} + \vectorsym{b^f}) \\
	\matrixsym{G^i}_t &= 1 - \matrixsym{G^f}_t \\
	\matrixsym{G^r}_t &= \sig((\matrixsym{S}_{t-1} \concat \matrixsym{X}_{t}) \tensorprod \matrixsym{W^r} + \vectorsym{b^r}) \\
	\matrixsym{S}_t &= \matrixsym{G^i}_t \odot \tanh(((\matrixsym{S}_{t-1} \odot \matrixsym{G^r}_t) \concat \matrixsym{X}_{t}) \tensorprod \matrixsym{W^s} + \vectorsym{b^s}) + \matrixsym{G^f}_t \odot \matrixsym{S}_{t-1} \label{eqn:gru}
\end{align}
In this case, the input gate is just the logical inverse of the forget gate. $\matrixsym{G^r}_t$ is called a reset gate and is used to turn parts of the previous state to zero when computing the new information to be added to the state. There is no output gate.

Other types of RNN include the recurrent highway network \citep{Zilly2017} which is a simple RNN but with multiple layers used to compute the next state, the tree LSTM \citep{Zhang2016a} which is an LSTM that takes in multiple input vectors rather than one in order to encode a tree of inputs, and neural Turing machines \citep{Graves2016} which keep several vectors as states and which choose which ones to update and read at every time step. It is worth mentioning that as of recently there is an alternative to RNNs called transformer networks \citep{Vaswani2017,Devlin2018} which encode text using only attention mechanisms (to be discussed in Subsection~\ref{sec:attention_mechanisms}), but these are beyond the scope of this work as we focus on RNNs.

The next section will discuss how to use RNNs in order to implement neural language models which can generate text.

\clearpage

\section{Neural language models}

The neural networks described thus far are able to accept a sentence as input. This section discusses how neural networks can be used to generate novel natural language sentences.

\subsection{Language models}

A language model is a function that takes in a sentence and returns the probability that the sentence belongs to a language that is being modelled by the language model. The language model finds the following probability:
\begin{align}
	P(w_{1 \dots n}) = P(w_1, w_2, \dots, w_n)
\end{align}
where $w_{1 \dots n}$ is a sentence with $n$ words and $w_i$ is the $i$\textsuperscript{th} word in the sentence.

By using the chain rule, this probability can be broken down into the following product of probabilities:
\begin{align}
	P(w_{1 \dots n}) &= P(w_1) \times P(w_2|w_1) \times P(w_3|w_1, w_2) \times \dots \times P(w_n|w_1, \dots, w_{n-1}) \\
	&= \prod_{i=1}^n{P(w_i|w_1, \dots, w_{i-1})}
\end{align}
where $P(w_i|w_1, \dots, w_{i-1})$ is the probability of using word $w_i$ after the sentence prefix $w_1 \dots w_{i-1}$. For example, given the sentence prefix `a man is walking his \dots', what is the probability that the next word in that prefix is `dog'?

Henceforth we will focus on $P(w_i|w_1 \dots w_{i-1})$ rather than $P(w_{1 \dots {n-1}})$. Given a prefix $w_1 \dots w_{i-1}$, a language model might predict the probability of $w_{i}$ for every word in the vocabulary at once. This is usually accomplished using a softmax function.

Softmax is useful in language models because it first makes all the scores positive by taking their exponent before normalising them. Since the exponent function never equals zero, softmax will not give a probability of zero to any word, which means that there is no need for smoothing to handle unlikely word sequences.

A language model is just a sequence classifier that probabilistically classifies each word in the vocabulary using a prefix of a sentence as input. It might be confusing to think of language models as classifiers when there are multiple possible next words given the same prefix (a classifier usually picks one class only). This is a classification problem where what is important is not deciding a correct class but the probabilities associated with the classes. The probability given to a word should be proportional to the number of times the word was found to follow the prefix in the training set.

\subsection{Text generation}

\begin{figure}
	\centering
	\includegraphics[scale=0.5]{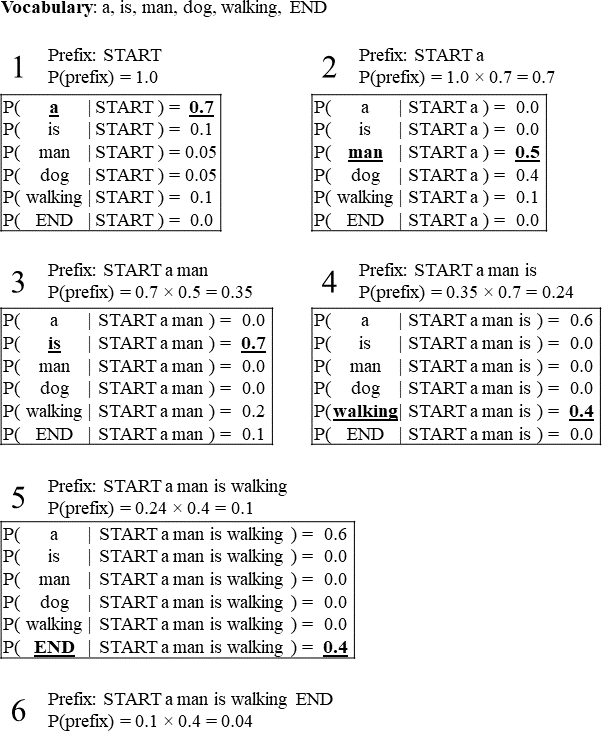}
	\caption{
		\label{fig:bg_text_generation}
		An illustration showing how a language model can be used to generate a sentence over a vocabulary of words. The prefix is extended with another word, where the word is chosen based on its probability according to the language model. Each word is chosen randomly using the given probabilities (which might not be the most likely word). After multiplying all the probabilities of the words in the generated prefix together, the probability of the sentence is found (in this case the probability is $0.04$).
	}
\end{figure}

Assuming that a function that computes $P(w_i|w_1 \dots w_{i-1})$ is available, a sentence can be generated over a vocabulary of words by repeatedly selecting a likely word to follow a given prefix and appending the selected word to the end of the prefix until the prefix becomes a complete sentence. This is illustrated in Figure~\ref{fig:bg_text_generation}.

Another two pseudo-words (apart from the unknown token and pad token) are added to the vocabulary in order to be able to begin and end the text generation process. The \kw{start token} is artificially added to the beginning of every sentence. Given that every sentence starts with this token, the probability of the start token is $1$ and thus does not need to be calculated. On the other hand, the \kw{end token} is artificially added to the end of every sentence and so as soon as the end token is selected, the sentence is fully generated. These two pseudo-words can be removed after the sentence is generated.

If the aim of the generation is to sample a sentence based on its probability (the probability of a particular sentence being generated is equal to the probability of the sentence according to the language model), then the next word to follow a prefix can be selected using roulette wheel selection. This is when a word is randomly chosen with a probability equal to that given by the language model. This is useful for generating likely sentences at random.

On the other hand, it is sometimes desirable to generate the most likely sentence according to the language model rather than a random sentence. This is essentially a search problem where the task is to find a path in a probability tree from the root to a leaf such that the probability of the path is maximised. This probability tree has the root node being the start token and the leaves are the end token. The probability of any node in the tree is calculated by the language model given the node's ancestors (the sentence prefix). Given a reasonably sized vocabulary, which can be in the hundreds of thousands, the arity of the probability tree is too large to be solved using exact algorithms in a practical amount of time and memory; so instead, approximation algorithms are used which find a highly likely path that is close to being the most likely.

\begin{figure}
	\centering
	\includegraphics[scale=0.5]{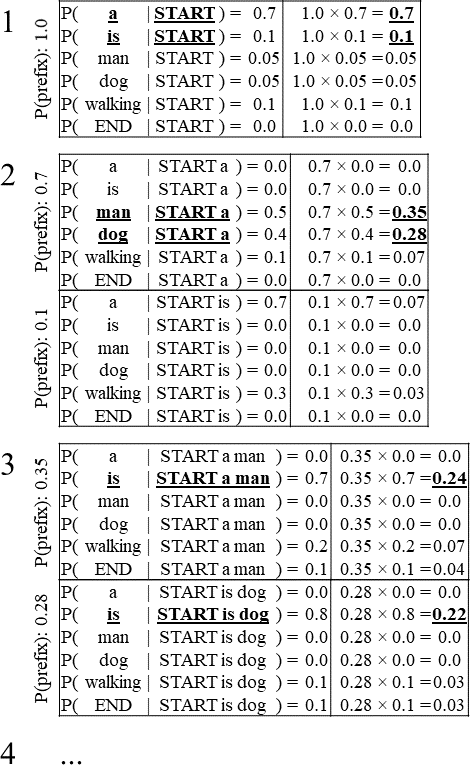}
	\caption{
		\label{fig:bg_beamsearch}
		A partial illustration showing how beam search works. The beam width is $2$. After each step, the two most probable candidate prefixes are selected to move to a new beam. When the most probable prefix in the beam ends with the end token, that prefix is returned as the most probable sentence.
	}
\end{figure}

The simplest and fastest method is greedy search, where the most likely next word is always chosen and no other possible path is explored. A better approach is to extend this algorithm into a \kw{beam search}, which is illustrated in Figure~\ref{fig:bg_beamsearch}. Greedy search is a beam search with a \kw{beam width} of one, because only one path is explored. A beam width of two would explore two possible paths as follows:

Given the start token as a prefix, the most likely two words are selected, which leads to two different prefixes that are added to a list called a beam. Each prefix in the beam is fed to the language model in order to see which words can follow each of the two prefixes. The probability of each word is multiplied by the probability of the prefix in order to see what the probability of the new prefix will be after appending each word. The two most likely new prefixes are chosen and added to a new beam. This is repeated until the most likely prefix in the beam has an end token, in which case that prefix is returned as the most likely sentence. The greater the beam width, the more likely that the generated sentence is the most probable sentence, but also the slower the process and the more memory is needed.

\subsection{Non-recurrent methods to language modelling}

In Subsection~\ref{sec:nonrnn_sentence_encoding} we described a naive way to encode a sentence by concatenating all the embedded words in the sentence and passing a single giant vector into a neural network. This idea is partly used in one of the earliest neural language models proposed \citep{Bengio2003}. The trick is to only use a small number of words, such as five, instead of a whole sentence. In order to predict the probability of the next word given a prefix, the language model makes a Markov assumption that only the previous five words are needed to predict the next word, similar to how n-gram language models work. This means that the neural network only needs to handle a truncated prefix of embedded word vectors concatenated together. A softmax layer at the output would then predict the probability of each word in the vocabulary being the next word after the truncated prefix.

This idea was further elaborated with log-bilinear models \citep{Mnih2007} which reduce the neural network to a simple linear layer (with no activation function) that transforms the concatenated word vectors into a query vector in the embedding space. The probability of a particular word matching the query vector, and hence being the next word in the truncated prefix, is determined by the dot product of the candidate word's vector and the query vector, called a score. Softmax is then applied to the scores in order to normalise them into probabilities.

\subsection{Recurrent neural language modelling}

The language models described in the previous subsection dealt with truncated prefixes only, requiring many more parameters added for every extra word the prefix is made to include. Recurrent neural networks can help with this problem as RNNs can encode arbitrary length prefixes using the same number of parameters.

\subsubsection{Basic language modelling}
\label{sec:bg_basic_lm}

\begin{figure}[b]
	\centering
	\includegraphics[scale=0.7]{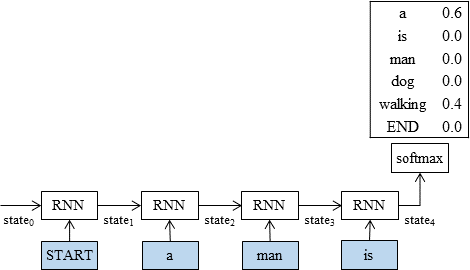}
	\caption{
		\label{fig:bg_rnnlangmod_disc}
		An illustration showing how an RNN language model processes a prefix of words to produce a probability for each word in the vocabulary.
	}
\end{figure}

An RNN language model uses an RNN in order to process a sequence of embedded word vectors and then pass the final state to a layer with a softmax activation function. The layer takes the encoded prefix and classifies the next word. This is illustrated in Figure~\ref{fig:bg_rnnlangmod_disc}.

The way this neural network is trained is by taking a corpus of text and breaking down each sentence into all possible prefixes together with the next word after each prefix. The neural network is then trained to predict the next word of all the prefixes by nudging the probability of the correct next word a little higher. Given that the softmax forces the sum of the probabilities to be equal to $1$, this implies that the other words get a slight reduction in their probability. Nonetheless, by doing this repeatedly, the probabilities of words that are not found to follow the prefix get reduced to be almost zero whilst the rest of the words get a probability that is proportional to the number of times they were found to follow the prefix.

\begin{figure}
	\centering
	\includegraphics[scale=0.7]{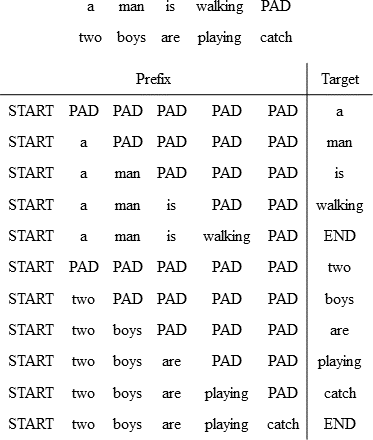}
	\caption{
		\label{fig:bg_langmoddataset_disc}
		An illustration showing how the dataset should be organised for a language model architecture like in Figure~\ref{fig:bg_rnnlangmod_disc}. Note that only two sentences are shown above.
	}
\end{figure}

The basic way to make use of this model is to format the dataset as illustrated in Figure~\ref{fig:bg_langmoddataset_disc}. Separating each prefix as an independent training item makes training very slow and memory wasteful due to all the padding that would be needed.

\subsubsection{Efficient language modelling}
\label{sec:bg_efficient_lm}

\begin{figure}
	\centering
	\includegraphics[scale=0.7]{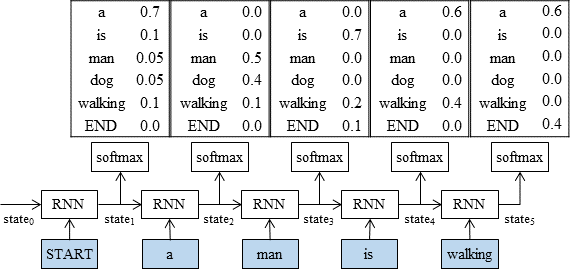}
	\caption{
		\label{fig:bg_rnnlangmod_cont}
		An illustration showing a more efficient way to train an RNN language model by passing in a whole sentence as input instead of a prefix and then using each state to predict the next word at that point. The probabilities after each prefix are then optimised. Note that the softmax layers are the same layer replicated for each state.
	}
\end{figure}

A more efficient alternative is to use the full sentence as an input and make a prediction after every word using the corresponding state, thus being able to optimise for several prefixes at once. This is shown in Figure~\ref{fig:bg_rnnlangmod_cont}.

\begin{figure}
	\centering
	\includegraphics[scale=0.7]{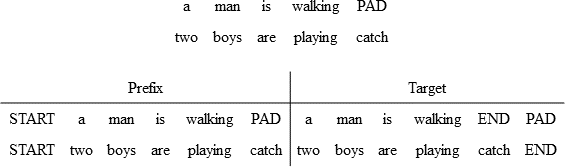}
	\caption{
		\label{fig:bg_langmoddataset_cont}
		An illustration showing how the dataset should be organised for a language model architecture like in Figure~\ref{fig:bg_rnnlangmod_cont}.
	}
\end{figure}

The dataset can now be organised in the form shown in Figure~\ref{fig:bg_langmoddataset_cont}. Note how the target in each row is now a sequence of words rather than a single word. This is basically the same sentence as in the prefix but instead of putting a start token at the beginning there is an end token at the end.

This method is much more efficient both in terms of speed and space. It also seems to be the most common way to train a language model and is the method we use in our experiments. During generation at test time, it is not possible to predict all the words of the sentence at once since the input prefix is not provided and needs to be constructed one word at a time. Therefore, at every time step during generation, the model will re-predict all the words that were already generated and these will need to be ignored as only the last predicted word will be useful. This word will then be added to the prefix and, again, only the last predicted word will be used.

As will be seen later, there are certain things that can only be done using the basic language modelling method. We will discuss these things in Section~\ref{sec:neural_image_caption_generation}.

\subsubsection{Training the language model}

The type of language model training where a ground truth prefix is provided at training time is called teacher forcing \citep{Williams1989}. The problem with this is that after training, the neural network is expected to generate a sentence from scratch without any guidance from a training set. If an incorrect word is selected at some point in the sentence generation process, the erroneous word is added to the prefix which might throw off all subsequent word probabilities, resulting in compounded errors.

Scheduled sampling \citep{Bengio2015} attempts to mitigate this problem by incorporating prefixes where some of the words were replaced with what would have been selected had the neural network been used as is to generate. Professor forcing \citep{Lamb2016} takes this approach further by generating some sentences in the middle of training, reading the states that result after inputting the ground truth sentences and the generated sentences and using generative adversarial training \citep{Goodfellow2014} to make the distribution of the two categories of states indistinguishable, making the RNNs learn to avoid veering off too much when an inconsistent word is introduced to the prefix.

\subsubsection{Other information}

There is a misconception that RNNs are sequence generators rather than encoders \citep{Tanti2017}. The role of RNNs in text generators is to encode prefixes of text in order to know what word should be predicted next. The prediction is done by the softmax layer and the generation is done by the whole neural network as part of the beam search algorithm. In the next section we will see that there are ways to condition RNN language models into providing descriptions for images by feeding visual information to the softmax layer directly, leaving the RNN unaware of what image is being described. This would not be possible if the generation was driven by the RNN.

Finally, there is the question of whether the way an RNN encodes a sentence prefix is sensible. The RNN treats the prefix as a linear structure rather than a hierarchical one which is based on syntactic or dependency parse trees. \citet{Frank2011} describe an experiment in which they find that linear models predict human reading time of text more accurately than hierarchical models, which suggests that humans rely more on linear processes to predict the next word being read than hierarchical processes. This is evidence that linear models like RNNs are linguistically valid in the context of predicting the next word.\\

\noindent The next section will explain how to take these language models, which are `blind', and adapt them into image caption generators by providing them with a visual input.

\clearpage


\section{Neural image caption generation}
\label{sec:neural_image_caption_generation}

In the previous section we have discussed how neural networks can be used to generate sentences, but the text that would be generated is unconditioned, that is, it does not have a way to specify `what' to generate. It would be more useful to be able to enforce a particular meaning in the generated sentence, that is, to control what the generated sentence is supposed to say by supplying some extra input. This is called `conditioning the language model' and the idea is to assign different probabilities to the same sentence given different desired meanings.

Some examples of conditioned language models are
\begin{itemize}
	\item machine translation where the generated target sentence needs to have the same meaning as the given source sentence \citep{Bahdanau2014},
	\item abstractive summarisation where the generated target text needs to be shorter than the source text whilst still retaining as much information as possible \citep{Nallapati2016},
	\item question answering systems where the generated sentence is an answer to a given question about a given body of data \citep{Sukhbaatar2015}.
\end{itemize}

In this work we are interested in a particular language model conditioning known as \kw{image caption generation} \citep{Bernardi2016,Hossain2019} where the input conditioning the language model is an image and the desired text is a description of the high-level content of the image.

\subsection{Automatic caption generation}

Although we refer to the task as image caption generation, technically the captions we typically encounter in print and social media are text that complements an image with extra information that is not available from the image itself, such as what a speaker is saying or what is the significance of the scene in the photo. These sort of descriptions are beyond the scope of this work. We instead focus on what \citet{Hodosh2013} refer to as concrete and conceptual image descriptions. Whereas abstract descriptions include things such as the mood being conveyed by the image (a romantic setting or a depressing scene), concrete descriptions are only limited to the entities in the image together with their attributes, relationships, and events they participate in. Whereas perceptual descriptions mention things like predominant colour or shapes, conceptual descriptions focus on more human-friendly concepts which is useful in tasks such as to aid visually-impaired people or to provide searchable text related to the image. We further focus on generic descriptions where no proper names of people or places are used.

Being able to automatically generate descriptions of images requires interaction between computer vision, the extraction of high level information from visual media, and natural language generation, the conversion of a non-linguistic representation into human readable text. Some early caption generation systems bridged these two fields by using sentence templates with blanks in them which are to be filled with words that are extracted from the image using object and attribute detectors \citep{Kulkarni2011,Mitchell2012,Elliott2013}. Other early systems would instead use a database of known image-description pairs. Given an image to be described, the system would then search the database for the most similar image in order to use its description \citep{Ordonez2011,Gupta2012,Mason2014}. Parts of several descriptions can be stitched together in order to create some variety \citep{Kuznetsova2014}. A variation on this approach is to learn a multimodal space using neural networks. A multimodal space is a vector representation that can represent both images and sentences such that the more similar in content the image or sentence is, the more similar the vectors will be (according to some distance measure). The system would then convert an image into a vector and look for the most similar vector of a sentence or convert a sentence into a vector and look for the most similar vector of an image \citep{Socher2014}.

Both template-based and retrieval-based methods, although intuitive, are not able to give novel descriptions as they are limited to the pre-programmed templates or database of existing captions. The third category of caption generators would be able to generate novel descriptions. This consists of a neural language model that is conditioned on an image. Even so, \citet{Devlin2015} report that out of all the captions generated by their conditioned language model, only about 33\% were unique whereas their retrieval-based method returned captions where about 37\% of them were unique. On the other hand, in our experiments, over 80\% of captions were unique so it is possible to obtain novel captions, even if maybe somewhat stereotypical.

Typically, conditioned language models use image features extracted from a hidden layer of another neural network that was pre-trained to perform object recognition. This vector of activation values is then supplied as an input to the neural language model which will condition the generated sentence. In this work we focus on this category of caption generators and thus will go into more detail on it below.

\subsection{Convolutional neural networks for images}
\label{sec:convnets}

An image is a tensor of pixel values where each value indicates a spot of colour. Grey scale images use a two-dimensional tensor (the width and height of the image) with each value indicating the brightness of the grey. On the other hand, colour images indicate colour by mixing different intensities of red, green, and blue and thus require a three dimensional tensor where the extra dimension stores the three colour channel values\footnote{In reality there are several different colour models available to indicate colour in an image other than the red-green-blue method. These include hue-saturation-value and the CIE 1931 XYZ color space.}. In general we consider an image to be a rectangular array of vectors where each vector contains the channels of each pixel (can be one channel for grey scale, three channels for reg-green-blue pixels, and so on).

Unlike sentences, images can be resized so that all the images in the training set have the same tensor shape meaning there is no need to pad an image in the same way that a sentence needs to be padded. Furthermore, unlike words, pixels are possible to work with directly without needing to embed them. This is because the values in the pixels represent an intensity of a colour rather than an arbitrary position in a vocabulary. Instead of embedding, the pixels are preprocessed by centering, that is, their values are altered such that the mean pixel value in all images in the training set is zero (which is done by subtracting the mean pixel value from all pixels in all images).

\begin{figure}
	\centering
	\includegraphics[scale=0.7]{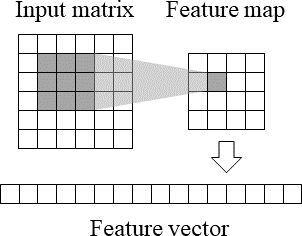}
	\caption{
		\label{fig:bg_2dconv}
		An illustration showing a convolutional neural network being used on an image. A convolutional layer performs a process where a sliding window called a receptive field (the three-by-three grey grid on the left) is moved along the image pixels, where pixels are vectors of channels, and the pixels in the window are passed through a filter to produce a single vector on a feature map. Each vector in the feature map represents a patch of the image. The feature map is then passed through another convolutional layer to produce another feature map (as if it were another image) until finally the feature map is flattened into a vector called a feature vector which is then passed through a normal fully connected neural network.
	}
\end{figure}

Just like encoding sentences, naively encoding an image using a normal fully connected layer that takes each pixel in a reasonably sized image as input has the same problem as that described in Subsection~\ref{sec:nonrnn_sentence_encoding}: it has far too many parameters. Instead, a two-dimensional \kw{convolutional neural network} (CNN) is used in order to be able to process an image with a minor amount of parameters. The 1D CNN has already been discussed in Subsection~\ref{sec:nonrnn_sentence_encoding}. A 2D CNN uses a rectangular sliding window called a receptive field that goes over a patch of pixels (vectors) in the image and uses a small fully connected layer called a filter to transform the patch of pixel values into a single vector. The vectors derived from all patches that were processed in the image are organised into a rectangular shape called a feature map such that vectors in the map come from correspondingly positioned patches in the image. The feature map is considered the activation values of a hidden layer. This map is then convolved again into another feature map and so on for each hidden layer. At the end, the last feature map is flattened into a single dimensional vector which can then be fed to a normal fully connected neural network layer in order to do something with the image representation, such as recognise the object shown in the image. This process is illustrated in Figure~\ref{fig:bg_2dconv}.

Feature maps can sometimes be considered a higher level representation of the image. This representation can be informative enough to be a useful for use in other tasks such as caption generation, similar to how word embeddings can be transferred to other tasks. Unfortunately, feature maps can be very large tensors, so an alternative is to use the activations of the fully connected layers instead which provide a compressed fixed-size vector that represents the image. One important advantage of representing images with feature maps rather than feature vectors is that feature maps preserve the locality of the features; that is, you can trace a feature on a feature map back to a region of pixels in the image, something which cannot be done with feature vectors. This makes them useful to process different regions of the image differently (which allows for more advanced image captioning techniques like attention, which will be discussed later).

CNNs are not typically trained from scratch for generating image representations that are optimised for caption generation. More commonly, the image representations are either extracted as is from a pre-trained CNN that was optimised to perform object recognition or the pre-trained CNN is further fine-tuned with more training on the caption generation task. This is because object recognition datasets such as ImageNet \citep{Russakovsky2015} contain millions of images to train on whilst caption generation datasets only contain a few hundreds of thousands of images, which is not enough to learn a generic image representation. The next subsection will discuss how to actually use the image representation for caption generation.

Well-known CNNs include LeNet \citep{Lecun1998}, AlexNet \citep{Krizhevsky2012}, VGGNet \citep{Simonyan2014}, GoogLeNet \citep{Szegedy2015}, and ResNet \citep{He2016}.

\subsection{Combining convolutional neural networks with neural language models}
\label{sec:combining_cnns_with_nlms}

\begin{figure}
	\centering
	\begin{subfigure}{\textwidth}
		\centering
		\includegraphics[scale=0.8]{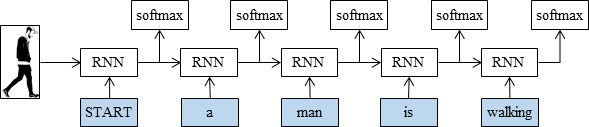}
		\caption{
			\label{fig:bg_cnnlm_init}
			The init-inject architecture.
			\vspace{15pt}
		}
	\end{subfigure}
	\begin{subfigure}{\textwidth}
		\centering
		\includegraphics[scale=0.8]{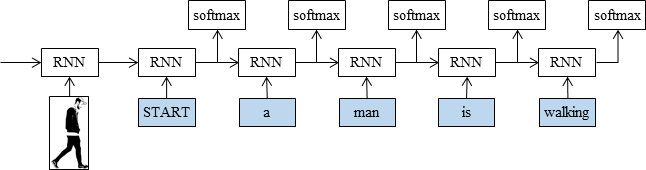}
		\caption{
			\label{fig:bg_cnnlm_pre}
			The pre-inject architecture.
			\vspace{15pt}
		}
	\end{subfigure}
	\begin{subfigure}{\textwidth}
		\centering
		\includegraphics[scale=0.8]{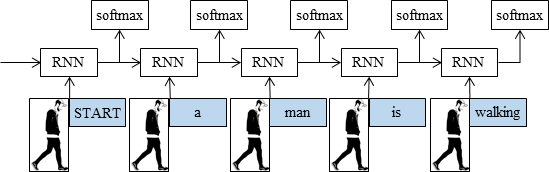}
		\caption{
			\label{fig:bg_cnnlm_par}
			The par-inject architecture.
			\vspace{15pt}
		}
	\end{subfigure}
	\begin{subfigure}{\textwidth}
		\centering
		\includegraphics[scale=0.8]{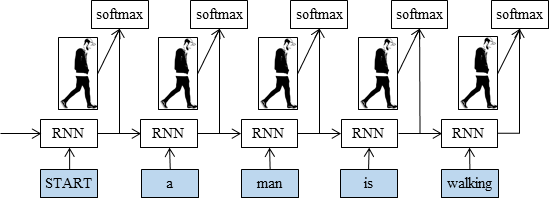}
		\caption{
			\label{fig:bg_cnnlm_merge}
			The merge architecture.
		}
	\end{subfigure}
	\caption{
		\label{fig:bg_cnnlm}
		The four main ways to connect an image representation to a neural language model in caption generation. Clipart image taken from \url{https://openclipart.org/detail/298220/man-walking}.
	}
\end{figure}

The types of caption generators we are interested in are the ones which take an image that has been converted into a feature representation and is then somehow fed into an RNN-based neural language model. A survey of the literature, which will be elaborated upon in Subsection~\ref{sec:litrev}, reveals that there are four broad categories for how this is done. Sometimes more than one method is used at once. In this work, these categories are called \kw{init-inject}, \kw{pre-inject}, \kw{par-inject}, and \kw{merge}. They are illustrated in Figure~\ref{fig:bg_cnnlm}.

\begin{itemize}
	\item Init-inject: Init-inject is when the RNN's initial hidden state vector is set to be the image vector (or some vector derived from the image vector). It requires the image vector to have the same size as the RNN hidden state vector.
	
	\item Pre-inject: Pre-inject is when the first input to the RNN is the image vector (or some vector derived from the image vector). The word vectors of the description prefix come later. The image vector is thus treated as a first word in the prefix and it requires the image vector to have the same size as the word vectors.
	
	\item Par-inject: Par-inject is when the image vector (or some vector derived from the image vector) serves as input to the RNN in parallel with the word vectors of the description prefix, such that the word vectors are combined with the image vector into a single input before being passed to the RNN. The image vector does not need to be included with every word, although it usually is, and can instead be included with the first word only (this is not pre-inject as the image is not injected on a separate time step).
	
	\item Merge: Merge is when the RNN is not exposed to the image vector (or some vector derived from the image vector) at any point. Instead, the image is introduced into the language model after the prefix has been encoded by the RNN.
\end{itemize}

It is possible to think about a fifth architecture called post-inject. This is when the image vector is injected as the last word in every prefix. Whilst it is possible to do so, it would also require that the language model be trained using the basic language model method described in Subsection~\ref{sec:bg_basic_lm}. This type of architecture is never mentioned in the literature and so we will ignore it.

We make a broad distinction between the inject architectures and the merge architecture which is that the inject architectures all inject the image information into the RNN's state in some way or another. This forces the RNN to accommodate two types of information in the state: the prefix of the description and the image. The merge architecture on the other hand leaves this separate visual information outside of the RNN and thus leaves the RNN part to act as a normal text-only language model which encodes the prefix.

Another important difference between architectures is that par-inject and merge both require the image vector to be replicated at every time step in the description whilst init-inject and pre-inject only allow the image vector to be inserted once at the start. This means that it is possible to use a different image representation at every time step for par-inject and merge such that the visual information evolves over time. An example of using different image representations at every time step is in attention mechanisms.

\subsection{Attention mechanisms}
\label{sec:attention_mechanisms}

Having a dynamic image representation which adapts according to what word needs to be generated next, referred to as attention, allowed caption generators to reach performance levels that would have probably been unattainable otherwise. Rather than having a bottleneck due to requiring a single vector to contain all the information needed to describe the whole image, attention mechanisms allow the image vector to only encode information that is necessary for the next word only. This means that the image vector can be smaller and better able to capture fine-grained information.

Attention in neural networks was first used in machine translation \citep{Bahdanau2014}. Early neural machine translation models required that the full source sentence be first represented as a single vector that is used to inform a language model on which words to use throughout the whole target sentence generation process \citep{Sutskever2014}. This requires the source sentence vector to encode enough information to be able to generate every word in the target sentence correctly, which becomes harder to do as the sentences become longer. Attention gets around this problem by producing a different source sentence representation for each word that needs to be generated, which is obtained by focussing on only a few words in the source sentence whilst ignoring those that do not contribute to the word that needs to be generated next.

This focus is achieved by multiplying each word vector by an element in a softmax vector, such that elements in the softmax that are close to zero will zero out the corresponding word vectors. By inspecting what these softmax vectors were before each word was generated, it is possible to check which words were being attended to, that is, what the neural network was `looking at' when it generated a particular word. The neural network would know which words to focus on based on the current state of the language model's RNN which stores information about what has been generated thus far and, by extension, what needs to be generated next.

\begin{figure}
	\centering
	\includegraphics[scale=0.8]{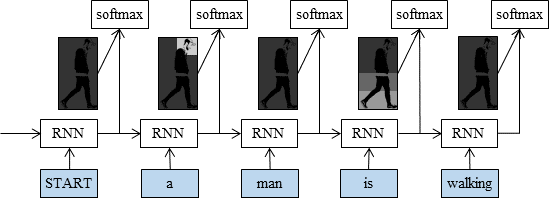}
	\caption{
		\label{fig:bg_attention}
		An illustration of how the image representation is altered per time step when attention is used. Each image representation filters out regions of the image that are not important for predicting the next word. For predicting `man', the face of the person would be focussed on the most. On the other hand, for predicting `walking', the legs would be focussed on the most. The focussing is more concentrated the smaller the number of regions being focussed on.
	}
\end{figure}

Attention can be applied to caption generation \citep{Xu2015,Rennie2017} by focussing on different regions of the image rather than words, depending on what kind of word needs to be generated next in the description. Once this word-specific image representation has been encoded, it is then inserted into the language model via par-inject or merge, with a different representation per time step. The process is illustrated in Figure~\ref{fig:bg_attention}. Init-inject and pre-inject do not allow for this sort of behaviour unless the basic language modelling method described in Subsection~\ref{sec:bg_basic_lm} is used, which would be slow.

\subsection{A review of existing caption generators}
\label{sec:litrev}

With the different architectures described in Subsection~\ref{sec:combining_cnns_with_nlms} in mind, we next discuss a selection of recent contributions, placing them in the context of this classification. Table~\ref{tbl:litrev_summary} provides a summary of these published architectures.

\begin{longtable}{l|c|c|c|c|p{5cm}}
	\hline\hline
	Source &	Init &	Pre &	Par &	Mrg &	Remarks \\
	\hline\hline
	\citep{Chen2014} &	 &	 &	$\checkmark$ &	 &	 \\ \hline
	\citep{Chen2015} &	 &	 &	$\checkmark$ &	 &	 \\ \hline
	\citep{Devlin2015} &	$\checkmark$ &	 &	 &	 &	 \\ \hline
	\citep{Donahue2015} &	 &	 &	$\checkmark$ &	 &	 \\ \hline
	\citep{Gu2017} &	 &	 &	$\checkmark$ &	 &	Encodes prefix using a CNN. \\ \hline
	\citep{Hendricks2016} &	 &	 &	 &	$\checkmark$ &	 \\ \hline
	\citep{Hessel2015} &	 &	 &	$\checkmark$ &	 &	Based on \citep{Karpathy2015}. \\ \hline
	\citep{Karpathy2015} &	 &	 &	$\checkmark$ &	 &	Image is only included with the first word. \\ \hline
	{\citep{Krause2016}} &	 &	$\checkmark$ &	 &	 &	Image is passed through a separate RNN at every time step and the hidden state vectors are pre-injected. \\ \hline
	\citep{Liu2016}$\dagger$ &	$\checkmark$ &	 &	 &	 &	 \\ \hline
	\citep{Liu2016}$\dagger$ &	$\checkmark$ &	 &	$\checkmark$ &	 &	Par-injects image attributes. \\ \hline
	\citep{Liu2017} &	$\checkmark$ &	 &	 &	$\checkmark$ &	Translates image attributes into a description with attention mechanism. Init-injects whole attributes and merges attended attributes. \\ \hline
	\citep{Lu2016} &	 &	 &	$\checkmark$ &	$\checkmark$ &	Attention mechanism which par-injects whole image and merges the attended image. \\ \hline
	\citep{Ma2016} &	$\checkmark$ &	 &	 &	 &	Translates image attributes into a description. \\ \hline
	\citep{Mao2014} &	 &	 &	 &	$\checkmark$ &	 \\ \hline
	\citep{Mao2015a} &	 &	 &	 &	$\checkmark$ &	 \\ \hline
	\citep{Mao2015} &	 &	 &	 &	$\checkmark$ &	 \\ \hline
	\citep{Nina2015} &	 &	$\checkmark$ &	 &	 &	 \\ \hline
	\citep{Oruganti2016} &	 &	 &	$\checkmark$ &	 &	Image is passed through a separate RNN several times so that a different image hidden state vector is injected at each time step. \\ \hline
	\citep{Rennie2017}$\dagger$ &	 &	$\checkmark$ &	 &	 &	 \\ \hline
	\citep{Rennie2017}$\dagger$ &	 &	 &	$\checkmark$ &	 &	Attention mechanism which par-injects the attended image into the part of the LSTM that is input gated. \\ \hline
	\citep{Venugopalan2017} &	 &	 &	 &	$\checkmark$ &	Based on \citep{Hendricks2016}. \\ \hline
	\citep{Vinyals2015} &	 &	$\checkmark$ &	 &	 &	 \\ \hline
	\citep{Vinyals2017} &	 &	$\checkmark$ &	 &	 &	Based on \citep{Vinyals2015}. \\ \hline
	\citep{Wang2016a} &	 &	 &	$\checkmark$ &	 &	Generates two descriptions from the front and in reverse and then the most probable is picked. \\ \hline
	\citep{Wang2016} &	$\checkmark$ &	 &	 &	 &	 \\ \hline
	\citep{Wu2015} &	 &	$\checkmark$ &	 &	 &	Pre-injects image attributes. \\ \hline
	\citep{Xu2015} &	$\checkmark$ &	 &	$\checkmark$ &	$\checkmark$ &	Attention-based mechanism which init-injects the full image while the attended image is par-injected and merged. \\ \hline
	\citep{Yao2017}$\dagger$ &	 &	$\checkmark$ &	 &	 &	First two words are the image attributes and the image. \\ \hline
	\citep{Yao2017}$\dagger$ &	 &	$\checkmark$ &	$\checkmark$ &	 &	Either pre-inject is made with image attributes and par-inject is made with the image or vice versa. \\ \hline
	\citep{You2016} &	 &	$\checkmark$ &	$\checkmark$ &	$\checkmark$ &	 \\ \hline
	\citep{Zhou2016} &	 &	$\checkmark$ &	$\checkmark$ &	 &	The image is modified by the last generated word before being par-injected. \\
	\hline\hline
	\caption{
		\label{tbl:litrev_summary}
		A summary of caption generators that use the different conditioning methods. $\dagger$ means that the publication describes multiple systems which use different conditioning methods.
	}
\end{longtable}

\paragraph{Init-inject architectures:} Models conforming to the init-inject architecture treat the image vector as the initial hidden state vector of an RNN \citep{Devlin2015,Liu2016}. \citet{Wang2016} combine two RNNs in parallel, both initialised with the same image. Some systems treat their images as sequences using attributes and then translate these attributes into a description through init-injection \citep{Ma2016}.

On the other hand, \citet{Liu2017} instead translate objects detected in an image into a description. In this case, the translation is one which uses attention mechanisms. The object vectors are passed through an RNN so that its states can be attended and merged during the generation of the description. The final state of the objects RNN is init-injected. Other systems also use init-injection in attention mechanisms in order to provide a vector representing information about the whole image. For example \citet{Xu2015} initialise the RNN with the centroid of all image parts before attending to some parts as needed.

\paragraph{Pre-inject architectures:} Models conforming to the pre-inject architecture treat the image as though it were the first word in the prefix \citep{Vinyals2015,Vinyals2017,Nina2015,Rennie2017}. Image attributes are sometimes used instead of image vectors \citep{Wu2015,Yao2017}. \citet{Yao2017} also try passing an image as the first two words instead of just one word by using the image vector as the first word and image attributes as a second, or vice versa.

Just like init-inject, pre-inject is also used to provide information about the whole image in attention mechanisms \citep{You2016,Zhou2016}.

\citep{Krause2016} generate paragraph-length descriptions in two stages. First, an RNN is used to convert the image vector into a sequence of image vectors by incorporating the image at every time step. This sequence of vectors represents sentence topics, each of which is to be converted into a separate sentence by conditioning a language model using pre-inject.

\paragraph{Par-inject architectures:} Models conforming to the par-inject architecture input the image features into the RNN jointly with each word in the description prefix. It is by far the most common architecture used and has the largest variety of implementations. For example \citet{Donahue2015} do this with two RNNs in series and find that it is better to inject the image in the second RNN than the first. \citet{Yao2017} par-inject the image whilst pre-injecting image attributes (or vice versa); and \citet{Liu2016} par-inject attributes from the image whilst init-injecting the image vector. Other, less common instantiations include par-injecting the image, but only with the first word \citep{Karpathy2015,Hessel2015}; processing the prefix using a one-dimensional CNN and then passing the encoded prefix to an RNN together with the image vector \citep{Gu2017}; generating two descriptions using two different RNNs, one from the front and one in reverse, and then picking the most probable one \citep{Wang2016a}; and passing the words through a separate RNN, such that the resulting hidden state vectors are what is combined with the image vector \citep{Oruganti2016}.

Many times this architecture is used in order to pass a different representation of the same image with every word so that visual information changes for different parts of the sentence being generated. For example \citet{Zhou2016} perform element-wise multiplication of the image vector with the last generated word's embedding vector in order to attend to different parts of the image vector. \citet{Oruganti2016} pass the image through its own RNN for as many times as there are words in order to use a different image vector for every word. \citet{Chen2014,Chen2015} use a simple RNN to try to predict what the image vector looks like given a prefix. This predicted image is then used as a second image representation which is par-injected together with the actual image vector.

More commonly, modified image representations come from attention mechanisms \citep{You2016,Xu2015,Rennie2017}. \citet{Rennie2017} inject the image not as an input to the RNN but use a modified LSTM, which allows them to inject the attended image directly inside the input gated expression (the part of the LSTM which is multiplied by the input gate).

Like init-inject and pre-inject, par-inject is sometimes used to provide information about the whole image in attention mechanisms whilst the attended image regions are merged \citep{Lu2016}.

\paragraph{Merge architectures:} Rather than combining image features together with linguistic features from within the RNN, merge architectures delay their combination until after the description prefix has been vectorised \citep{Mao2014,Mao2015,Mao2015a}. \citet{Hendricks2016} and later \citet{Venugopalan2017} use a merge architecture in order to keep the image out of the RNN and thus be able to train the part of the neural network that handles images and the part that handles language separately, using images and sentences from separate training sets.

Some work on attention mechanisms also uses merge architectures with attention mechanisms by merging a different image representation at every time step. \citet{You2016} and \citet{Xu2015} merge as well as par-inject the attended visual regions; \citet{Lu2016} only merge the regions whilst par-injecting a fixed image representation; \citet{Liu2017} pass vector encoded objects detected in an image into an RNN and the final state is init-injected whilst the attended RNN states are merged.

Though they do not use an RNN and hence are not focussed on in this review, caption generators that use log-bilinear models usually merge the image with the prefix representation \citep{Kiros2014,Kiros2014a,Song2016}.

\subsection{Evaluation measures}

The best way to evaluate how well a caption generator performs is by asking human annotators to rate the quality of the captions it generates. Unfortunately, this is time consuming, and instant automatic evaluation measures are necessary during development. The most basic evaluation measure is the perplexity measure which was developed to evaluate language models in general. Perplexity is defined as
\begin{align}
	\text{perplexity}(s, P) &= 2^{\text{entropy}(s, P)} \\
	\text{entropy}(s, P) &= -\frac{1}{|s|} \sum_{i=2}^{|s|} \log_2 P(s_i | s_{1}\dots s_{i-1})
\end{align}
where $s$ is a sentence (caption), $P$ is a language model that predicts the probability of a word in a sentence given its preceding words, $s_i$ is the $i$\textsuperscript{th} word in the sentence, $|s|$ is the number of words in sentence $s$, and $s_1$ and $s_{|s|}$ are the start and end token respectively.

Given a reference sentence that was written by a human, the perplexity function measures how probable that sentence is according to the language model. The more probable a given correct sentence is according to the model, the more likely that sentence is to be generated by the model. Perplexity is simple and fast to calculate but it does not measure the quality of generated sentences, only how likely it is to generate given correct sentences. To measure the quality of actual generated sentences we will need to use caption quality metric functions. Unfortunately, perplexity does not correlate well with these functions \citep{Tanti2019a}.

Automatic sentence quality measures were originally developed for other text generation tasks such as machine translation and automatic summarisation, which were later adopted by the caption generation community. Basically, given a set of possible (human written) translations for a given source sentence, these functions would measure how similar the generated translation is to the reference sentences by breaking the sentences down into n-grams and counting how many n-grams the generated and reference sentences have in common. Examples of such measures are BLEU \citep{Papineni2002}, ROUGE \citep{Lin2004}, and METEOR \citep{Banerjee2005}. METEOR in particular makes use of a thesaurus in order to allow for matching synonyms between n-grams rather than rely on exact matching alone. In caption generation, rather than use a set of possible translations for a given source sentence, a set of possible descriptions of a given image is used instead.

More specific image description evaluation measures were eventually developed, notably CIDEr (Consensus-based Image Description Evaluation) \citep{Vedantam2015} and SPICE (Semantic Propositional Image Caption Evaluation) \citep{Anderson2016}. CIDEr also uses n-gram similarities but makes use of TF-IDF to measure the similarity and was tuned on captions specifically rather than general sentences. SPICE computes similarity between sentences from scene graphs \citep{Johnson2015}, which are graphs that specify the content of a scene in an image based on objects and their attributes and relationships. The idea is to measure the similarity of the sentence to the content of the image directly, however a scene graph would be difficult to extract from an image and so is instead approximated by parsing the reference sentences.

Lately, WMD (Word Mover's Distance) \citep{Kusner2015}, a function that was originally developed for general document similarity, has also been adapted for caption evaluation \citep{Kilickaya2017}. WMD measures the semantic distance between texts by measuring how many words they have in common, in no order; however, the words are matched using word2vec \citep{Mikolov2013} embeddings and their cosine distance. In order to measure how much word meaning the two texts have in common, Earth Mover's Distance is used which is a function that measures the minimum distance needed to move each word vector in one text to one of the word vectors in the other text. The words are weighted by their frequency in their corresponding text and stop words are removed.

We opted to use WMD as a representative caption quality measure because its authors found that it correlates well with human judgement whilst being robust to synonym swapping and other distraction tasks. Our decision is further confirmed because, as will be seen later, it corresponds the best to other ways of measuring the quality of caption generators.

\clearpage


\section{Conclusion}

While the literature on caption generation now provides a rich range of models and comparative evaluations, the different architectures described above have been given little attention in terms of performance comparisons. Work that has tested both par-inject and pre-inject, such as by \citet{Vinyals2015}, reports that pre-inject works better. The work of \citet{Mao2015} compares inject and merge architectures and concludes that merge is better than inject. Mao {\em et al.}'s comparison between architectures is however a relatively tangential part of their overall evaluation, and is based only on the BLEU metric \citep{Papineni2002}.

Answering the question of which architecture is best is difficult because different architectures perform differently on different evaluation measures, as shown for example by \citet{Wang2016}, who compared architectures with simple RNNs and LSTMs. Although the state of the art systems in caption generation all use inject-type architectures, it is also the case that they are more complex systems than the published merge architectures and so it is not fair to conclude that inject is better than merge based on a survey of the literature alone. This is what Research Question 3 asks in Chapter 1 of this thesis.

There are also several challenges worth mentioning in the field of caption generation in general. For example, it turns out that neural caption generators do a poor job of grounding. \citet{Shekhar2017} used a caption generator in order to determine if a word in a caption was replaced with an incorrect word (such that the caption does not describe the image any more). This was done by measuring how much the probability of the whole sentence according to the neural network goes down as a result of replacing the word. It was found that the model was only able to detect the incorrect word 45\% of the time whilst a `blind' language model could do so 25\% of the time. Similar techniques are used in this thesis in order to measure how much influence the image has on the model in order to answer Research Question 4. Another interesting observation was that simply providing the caption generator with a vector that describes what objects are in the image instead of a complex visual features representation results in better captions \citep{Wang2018}. This also means that caption generators can predict verbs using only the nouns rather than by looking at what is going on in the image. This is also relevant to Research Question 4.

In what follows, we present a systematic comparison between all the different architectures discussed above. We perform these evaluations using a common dataset and a variety of quality metrics, covering (a) the quality of the generated descriptions; (b) the linguistic diversity of the generated descriptions; and (c) the networks' capabilities to determine the most relevant image given a description.

\clearpage

%% file: tex/chp3_architectures.tex
\chapterwithfootnote{Architecture comparison}{An earlier version of the work shown in this chapter has been published \citep{Tanti2018}.}

\clearpage

\section{Aims}

In this chapter we will compare each of the four architectures shown in Figure~\ref{fig:bg_cnnlm} in the previous chapter using several metrics to see each one's pros and cons when trained under similar conditions.

The main contribution of these experiments is to present a systematic comparison of the different ways in which the conditioning of linguistic choices based on visual information can be carried out, studying their implications for caption generator architectures. Thus, rather than seeking new results that improve on the state of the art, we seek to determine, based on an exhaustive evaluation of inject and merge architectures on a common dataset, where image features are best placed in a caption generator.

From a scientific perspective, such a comparison would be useful for shedding light on the way language can be grounded in vision. Should images and text be intermixed throughout the process, or should they initially be kept separate before being combined in some multimodal layer? Many papers speak of RNNs as `generating' text. Is this the best way to view them or are RNNs better viewed as encoders which vectorise a linguistic prefix so that the next feed-forward layer can predict the next word, conditioned on an image? Answers to these questions would help inform theories of how caption generation can be performed. The architectures we compare provide different answers to these questions. Hence, it is important to acquire some insights into their relative merits.

From an engineering perspective, insights into the relative performance of different models could provide rules of thumb for selecting an architecture for the task of image captioning, possibly for other tasks as well, such as machine translation. This would make it easier to develop new architectures and new ways to perform caption generation.

\clearpage


\section{Experiments}

As a reminder, the four architectures being investigated (illustrated in Figure~\ref{fig:bg_cnnlm} and discussed in Subsection~\ref{sec:combining_cnns_with_nlms}) are:
\begin{itemize}
	\item Init-inject: Image as initial state of RNN.
	\item Pre-inject: Image as first word in RNN.
	\item Par-inject: Image concatenated to every word.
	\item Merge: Image concatenated to RNN hidden state vector.
\end{itemize}

\subsection{Datasets}
\label{sec:wi_dataset}

The datasets used for all experiments were the version of Flickr8K \citep{Hodosh2013}, Flickr30K \citep{Young2014}, and MSCOCO \citep{Lin2014} distributed by \citet{Karpathy2015}.\footnote{See: \url{http://cs.stanford.edu/people/karpathy/deepimagesent/}}. All datasets consist of images taken from Flickr\footnote{See: \url{https://www.flickr.com}} combined with between five and seven manually written captions per image. The provided datasets are split into a training, validation, and test set using the following number of images respectively:
\begin{itemize}
	\item Flickr8K - 6\,000 (75.0\%), 1\,000 (12.5\%), 1\,000 (12.5\%);
	\item Flickr30K - 29\,000 (93.5\%), 1\,014 (3.3\%), 1\,000 (3.2\%);
	\item MSCOCO - 82\,783 (89.2\%), 5\,000 (5.4\%), 5\,000 (5.4\%).
\end{itemize}

The images were vectorised into 4\,096-element vectors via the activation values of layer `fc7' (the penultimate layer) of the VGG OxfordNet 16-layer convolutional neural network \citep{Simonyan2014}, which was trained for object recognition on the ImageNet dataset \citep{Deng2009}. The convolutional neural network is fixed and not modified by the caption generator's training. The pre-trained VGG OxfordNet was obtained from Davi Frossard's VGG16 implementation.\footnote{See: \url{https://www.cs.toronto.edu/~frossard/post/vgg16/}}

The known vocabulary consists of all the words in the captions of the training set that occur at least 5 times. This amounts to 2\,532 words for Flickr8K, 7\,342 words for Flickr30K, and 8\,725 words for MSCOCO. Any other word which is not part of the vocabulary is replaced with the unknown token. In order to reduce the amount of different tokens, we preprocess all the captions in the datasets by lowercasing all characters, replacing strings of digits with a `NUM' token, and removing all non-alphanumeric non-space characters.

\subsection{Architecture}
\label{sec:architecture}

\begin{figure}[t]
	\centering
	\includegraphics[scale=1.0]{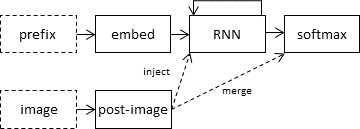}
	\caption{
		\label{fig:wi_architecture}
		An illustration of the main architecture schema that is instantiated in the four different architectures tested in the experiments. The neural network takes a prefix of a sentence, embeds each word, and encodes it via an RNN (a GRU). A 4\,096-element feature vector representing the image is projected into a smaller vector via a fully connected layer (post-image) which is then either mixed in with the RNN (inject) or concatenated to the RNN hidden state vector (merge). The mixed image-prefix multimodal vector is then passed to the softmax layer to predict the next word in the prefix. Only one of the dashed arrows is used depending on whether the architecture is one of merge or inject.
	}
\end{figure}

In order to make a fair evaluation, a basic schema was used to construct each architecture. A diagram illustrating the schema is shown in Figure~\ref{fig:wi_architecture}. The schema is based on the architecture described by \citet{Vinyals2015}, without the ensemble, which was chosen for its simplicity whilst still being the best performing system in the 2015 MSCOCO image captioning challenge.\footnote{See: \url{http://mscoco.org/dataset/\#captions-leaderboard}} Tensorflow v1.4\footnote{See: \url{https://www.tensorflow.org/}} was used to implement the neural networks.

\paragraph{Word embeddings:} Word embeddings, that is, the vectors that represent known words prior to being fed to the RNN, consist of vectors that have been randomly initialised. No precompiled vector embeddings, such as word2vec \citep{Mikolov2013} were used. Instead, the embeddings are trained as part of the neural network in order to learn the best word representations for the task.

\paragraph{Recurrent neural network:} The purpose of the RNN is to take a prefix of embedded words (together with the image vector in inject architectures) and produce a single vector that represents the sequence. A GRU \citep{Chung2014} was used in our experiments for the simple reason that it is a powerful RNN that only has one hidden state vector. By contrast, an LSTM has two state vectors (the hidden state and the cell state). This would make architecture comparisons more complex, as the presence of two state vectors raises the possibility of multiple versions of the init-inject architecture such as using one state but not the other or using both at once. By using an RNN with a single hidden state vector there is only one way to implement init-inject.

\paragraph{Image:} All images are input as 4\,096-element vectors. A fully connected layer compresses this vector into a smaller post-image vector.

\paragraph{Output:} Once the image and the caption prefix have been vectorised and mixed into a single vector, called a multimodal vector, the next step is to use them to predict the next word in the caption. This is done by passing the mixed vector through a fully connected layer with a softmax activation function that outputs the probability of each possible next word in the vocabulary. Based on this distribution, the next word to come after the prefix is selected.

\subsection{Hyperparameter tuning}
\label{sec:wi_hyperparameter_tuning}

For the results to be reliable, it is important to find the best (within practical limits) hyperparameters for each architecture so that we can judge the performance of the architectures when they are optimally tuned, rather than using one-size-fits-all hyperparameter settings which might cause some architectures to under-perform. The library Scikit-Optimize\footnote{See: \url{https://scikit-optimize.github.io/}} was used to perform hyperparameter tuning using Bayesian optimisation with a random forest model. It is also possible to use Gaussian processes instead of random forests but Gaussian processes are not as good \citep{Eggensperger2015}. The model is used to predict the expected improvement of a given hyperparameter combination. To train the model, it is initialised using 32 random hyperparameters together with their evaluated resulting performance after training the neural network. Following this, the model continues to be improved by exploring a sequence of 64 candidate hyperparameters that the model suggests will maximise the expected improvement, the result of each one being fed back to the model. At the end, the best hyperparameter combination found out of the 96 different hyperparameters ($32+64$) is used to set the neural networks during the experiments.

For each hyperparameter combination to be evaluated, we trained a neural network on the Flickr8K training set based on the hyperparameters and then generated captions for the MSCOCO validation set using beam search as a generation method (\citet{Mao2015} also used Flickr8K for hyperparameter tuning). The reason why Flick8K was used for training is to speed up the hyperparameter tuning process (it would take too long to train on MSCOCO with different hyperparameters 96 times). The reason why the MSCOCO validation set was used instead of the Flickr8K one is because the Flickr8K validation set is used for early stopping during hyperparameter evaluation. Thus it was preferable to use an alternative validation set during tuning to avoid subsequent evaluation of a model on data that had influenced its training. The Word Mover's Distance \citep{Kusner2015,Kilickaya2017}  metric, or WMD, was used to measure the quality of the generated captions. This process was performed twice for each hyperparameter combination and the average WMD resulting from the two independent train and generation sessions was used as a score for the hyperparameter combination. This makes the score more robust than if the model was only trained and evaluated once. The optimal hyperparameters found were then used in the experiments across all datasets.

The following hyperparameters are the ones that were tuned:

\paragraph{Weights initialisation method:} The probability distribution used to initialise all the weights. Can be the normal distribution or Xavier initialisation \citep{Glorot2010} with normal distribution.

\paragraph{Maximum initial weight:} The maximum absolute value of the initial weight beyond which it is clipped (for both positive and negative values). Can be between 1e-5 and 1.0.

\paragraph{Embed size:} The layer size of the embedding layer. Can be between 64 and 512.

\paragraph{RNN size:} The hidden state vector size of the RNN. Can be between 64 and 512.

\paragraph{Post-image size:} The layer size of the post-image layer. Can be between 64 and 512.

\paragraph{Post-image activation:} The activation function used on the post-image layer. Can be ReLU or none (the identity function).

\paragraph{Optimiser:} The optimiser used for training. Can be Adam \citep{P.Kingma2014}, RMSProp\footnote{See: \url{http://www.cs.toronto.edu/~tijmen/csc321/slides/lecture_slides_lec6.pdf}}, or AdaDelta \citep{Zeiler2012}. These were selected based on common practices in the literature: \citet{Rennie2017} used Adam, \citet{Karpathy2015} used RMSProp, and \citet{Mao2015a} used AdaDelta.

\paragraph{Learning rate:} The learning rate to use with the chosen optimiser. Can be between 1e-5 and 1.0.

\paragraph{Normalise image:} Whether to use the vector norm of the image feature vector or leave the image as is. Can be true or false.

\paragraph{Weight decay weight:} The weight assigned to the weight decay regularisation. Can be between 1e-10 and 0.1.

\paragraph{Image dropout rate:} The dropout rate applied to the image vector input. Can be between 0.0 and 0.5.

\paragraph{Post-image dropout rate:} The dropout rate applied to the post-image layer. Can be between 0.0 and 0.5.

\paragraph{Embed dropout rate:} The dropout rate applied to the embedding layer. Can be between 0.0 and 0.5.

\paragraph{RNN dropout rate:} The dropout rate applied to the hidden state vector of the RNN. Can be between 0.0 and 0.5.

\paragraph{Maximum gradient norm:} The maximum norm of the gradients beyond which it is clipped. Can be between 1.0 and 1000.0.

\paragraph{Minibatch size:} The minibatch size to use during training. Can be between 10 and 300.

\paragraph{Beam width:} The beam width to apply when using beam search to generate captions. Can be between 1 and 5.\\

\noindent The following hyperparameters are the ones that were fixed (not tuned):

\paragraph{RNN:} The RNN was set to be a GRU for all architectures in order to have a powerful RNN that only has one hidden state vector rather than two like the LSTM does. To initialise the RNN (except for init-inject which takes the image as an initial state), we use a learnable vector that is optimised together with the rest of the neural network.

\paragraph{Loss function:} The loss function used during training is the mean of the cross-entropy of each word in each caption in a minibatch.

\paragraph{Early stopping:} Training ends when the geometric mean of the perplexity on the validation set after a particular epoch is worse than it was after the previous epoch. Training does not terminate before then.

\paragraph{Caption generation:} The generated captions must be between 5 and 20 words long. Beam search will not end a sentence before there are at least 5 words in it and will abruptly stop a sentence that is 20 words long. A caption cannot have the same word twice next to each other and cannot have the unknown token in it.

\paragraph{Bias initialisation:} All biases are initialised to zeros.

\paragraph{Adam optimiser hyperparameters:} Other than the learning rate, the other hyperparameters used by the Adam optimiser were left as default, that is, $\beta_1=0.9$, $\beta_2=0.999$, and $\epsilon=$1e-08.\\

\noindent In Subsection~\ref{sec:wi_results:hyperparameters} we will discuss the optimal hyperparameters found. It is important to note that the init-inject and pre-inject architectures have an advantage over the other two architectures when it comes to hyperparameter searching since their search space is smaller because of the constraint that the post-image size must be equal to the RNN size (init-inject) or the embed size (pre-inject). This means that it is likely that these two architectures will end up with better hyperparameters, but this is also an advantage of the architectures.

\subsection{Evaluation metrics}

To evaluate the different architectures, the test set images are used to measure the architectures' quality using metrics that fall into four classes, described below.

\paragraph{Quality metrics:} These metrics quantify the quality of the generated captions by measuring the degree of overlap between generated captions and those in the test set. We use the MSCOCO evaluation code\footnote{See: \url{https://github.com/tylin/coco-caption}} which measures the standard evaluation metrics BLEU-(1,2,3,4) \citep{Papineni2002}, ROUGE-L \citep{Lin2004}, METEOR \citep{Banerjee2005}, CIDEr \citep{Vedantam2015}, and SPICE \citep{Anderson2016}. The evaluation code does not include WMD \citep{Kusner2015,Kilickaya2017} as a metric so we created a fork of the repository that also includes WMD\footnote{See: \url{https://github.com/mtanti/coco-caption}}.

\paragraph{Diversity metrics:} Even though we would expect conditioned language models to generate novel descriptions, it is possible for them to produce the same generic captions for different images \citep{Devlin2015} which would be undesirable. Therefore, apart from measuring the caption similarity to the ground truth, we also measure the diversity of the captions. To quantify the novelty of generated captions we use the following measures:
\begin{itemize}
\item the percentage of known vocabulary words used in all generated captions (indicates the extent of vocabulary exploitation),
\item the percentage of unique sentences generated (indicates the variety of sentences generated),
\item and the number of sentences that were copied from the training set (indicates the amount of novel sentences generated).
\end{itemize}
As a ceiling estimate of diversity, we compute the same metrics on the human-written test set captions themselves. For each group of human-written captions available for each image in the test set, we extract a random caption and apply these diversity metrics on these extracted captions.

\paragraph{Retrieval metrics:} Retrieval metrics are metrics that quantify how well the architectures perform when retrieving the correct image out of all the test set images in the test set given a corresponding caption. A conditioned language model can be used for retrieval by measuring the degree of relevance each image has to the given reference caption (according to the model). Relevance is measured as the probability of the whole caption given the image (by multiplying together each word's probability). Different images will usually result in different probabilities for the same caption. The more probable the caption is, the more relevant the image. We use the standard $R$@$n$ recall measures \citep{Hodosh2013} and report recall at 1, 5, and 10. Recall at $n$ is the percentage of captions whose correct image is among the top $n$ most relevant images. We also calculate the median rank of each correct image in the sorted list of retrieved images. Since this process takes time proportional to the number of captions multiplied by the number of images, the pool of possible reference captions to consider during retrieval only includes one caption out of the group of captions available for each image in order to reduce the evaluation time. The selected caption for each image was randomly chosen but was kept the same across models.

\paragraph{Probability metrics:} Apart from the retrieval metrics, probability metrics are also useful measures on their own. The higher the probabilities of the probability of the test caption given the test image, the higher the probability that the test caption is generated given the image. We measure both caption probability and caption perplexity and aggregate all the caption scores into a single score by using mean, median, and geometric mean.

\clearpage


\section{Results}

\subsection{Hyperparameter tuning results}
\label{sec:wi_results:hyperparameters}

In order to extract as much information as possible from the conducted experiments, we also analysed the results produced from the hyperparameter search for each architecture. The top five hyperparameter combinations for each architecture are shown in Table~\ref{tbl:wi_hyperparams}, where the `rank 1' column gives the best hyperparameters found and which were used in the experiments. The WMD is small because these neural networks were trained on Flickr8K whilst being evaluated on the MSCOCO validation set.

\begin{table}
	\centering
	\begin{subtable}{\textwidth}
		\centering
		\begin{tabular}{l|ccccc}
			Rank &	1 &	2 &	3 &	4 &	5 \\
			WMD &	0.0832 &	0.0830 &	0.0821 &	0.0813 &	0.0813 \\
			\hline
			weight init. method &	Normal &	Normal &	Xavier &	Normal &	Xavier \\
			max. init. weight &	3.50e-03 &	1.22e-05 &	2.92e-05 &	2.43e-02 &	2.36e-05 \\
			embed size &	502 &	123 &	315 &	202 &	150 \\
			RNN size &	418 &	132 &	396 &	179 &	488 \\
			post-image size &	418 &	132 &	396 &	179 &	488 \\
			post-image activation &	none &	none &	none &	none &	none \\
			optimiser &	Adam &	Adam &	RMSProp &	Adam &	RMSProp \\
			learning rate &	1.73e-03 &	1.17e-03 &	8.86e-04 &	1.35e-04 &	1.22e-03 \\
			normalise image &	true &	false &	false &	false &	true \\
			weight decay weight &	2.74e-06 &	6.28e-07 &	6.03e-06 &	3.13e-09 &	1.24e-07 \\
			image dropout rate &	0.01 &	0.06 &	0.04 &	0.12 &	0.01 \\
			post-image dropout rate &	0.34 &	0.46 &	0.38 &	0.30 &	0.48 \\
			embedding dropout rate &	0.34 &	0.22 &	0.37 &	0.33 &	0.40 \\
			RNN dropout rate &	0.11 &	0.12 &	0.06 &	0.15 &	0.00 \\
			max. gradient norm &	1.57e+00 &	1.63e+01 &	4.11e+00 &	8.56e+00 &	9.85e+00 \\
			minibatch size &	168 &	270 &	284 &	54 &	239 \\
			beam width &	3 &	5 &	2 &	2 &	5 \\
		\end{tabular}
		\caption{
			\label{tbl:wi_hyperparams_init}
			Top five hyperparameter combinations for the init-inject architecture.
			\vspace{15pt}
		}
	\end{subtable}
	
	\begin{subtable}{\textwidth}
		\centering
		\begin{tabular}{l|ccccc}
			Rank &	1 &	2 &	3 &	4 &	5 \\
			WMD &	0.0800 &	0.0790 &	0.0777 &	0.0772 &	0.0762 \\
			\hline
			weight init. method &	Xavier &	Xavier &	Normal &	Normal &	Xavier \\
			max. init. weight &	5.04e-04 &	2.16e-05 &	1.50e-01 &	3.50e-05 &	5.03e-05 \\
			embed size &	206 &	112 &	200 &	220 &	95 \\
			RNN size &	498 &	408 &	391 &	480 &	312 \\
			post-image size &	206 &	112 &	200 &	220 &	95 \\
			post-image activation &	none &	none &	none &	ReLU &	none \\
			optimiser &	RMSProp &	Adam &	Adam &	RMSProp &	Adam \\
			learning rate &	5.54e-04 &	4.66e-03 &	1.42e-03 &	6.39e-04 &	8.64e-04 \\
			normalise image &	false &	true &	true &	false &	true \\
			weight decay weight &	1.77e-09 &	3.00e-07 &	3.29e-06 &	1.39e-06 &	3.34e-06 \\
			image dropout rate &	0.26 &	0.14 &	0.21 &	0.47 &	0.29 \\
			post-image dropout rate &	0.29 &	0.20 &	0.42 &	0.11 &	0.29 \\
			embedding dropout rate &	0.29 &	0.08 &	0.26 &	0.24 &	0.26 \\
			RNN dropout rate &	0.09 &	0.22 &	0.08 &	0.14 &	0.48 \\
			max. gradient norm &	1.65e+00 &	2.54e+01 &	1.08e+01 &	9.16e+00 &	1.91e+01 \\
			minibatch size &	85 &	122 &	21 &	183 &	16 \\
			beam width &	5 &	4 &	3 &	1 &	2 \\
		\end{tabular}
		\caption{
			\label{tbl:wi_hyperparams_pre}
			Top five hyperparameter combinations for the pre-inject architecture.
		}
	\end{subtable}
\end{table}
\begin{table}
	\ContinuedFloat
	\centering
	\begin{subtable}{\textwidth}
		\centering
		\begin{tabular}{l|ccccc}
			Rank &	1 &	2 &	3 &	4 &	5 \\
			WMD &	0.0827 &	0.0827 &	0.0815 &	0.0814 &	0.0809 \\
			\hline
			weight init. method &	Xavier &	Normal &	Xavier &	Xavier &	Normal \\
			max. init. weight &	6.25e-02 &	7.75e-04 &	1.35e-02 &	2.74e-02 &	7.64e-03 \\
			embed size &	493 &	461 &	511 &	446 &	502 \\
			RNN size &	451 &	494 &	432 &	174 &	418 \\
			post-image size &	359 &	212 &	446 &	155 &	465 \\
			post-image activation &	none &	none &	ReLU &	ReLU &	none \\
			optimiser &	Adam &	RMSProp &	RMSProp &	RMSProp &	Adam \\
			learning rate &	2.54e-04 &	5.77e-04 &	4.95e-04 &	4.17e-04 &	4.39e-04 \\
			normalise image &	false &	false &	false &	false &	false \\
			weight decay weight &	2.18e-10 &	4.96e-10 &	2.01e-10 &	1.10e-08 &	7.71e-09 \\
			image dropout rate &	0.23 &	0.42 &	0.37 &	0.37 &	0.16 \\
			post-image dropout rate &	0.33 &	0.19 &	0.42 &	0.08 &	0.19 \\
			embedding dropout rate &	0.01 &	0.01 &	0.01 &	0.03 &	0.05 \\
			RNN dropout rate &	0.19 &	0.25 &	0.03 &	0.17 &	0.16 \\
			max. gradient norm &	6.77e+01 &	9.99e+01 &	4.22e+01 &	1.12e+02 &	2.36e+01 \\
			minibatch size &	143 &	118 &	141 &	151 &	149 \\
			beam width &	2 &	4 &	5 &	4 &	3 \\
		\end{tabular}
		\caption{
			\label{tbl:wi_hyperparams_par}
			Top five hyperparameter combinations for the par-inject architecture.
			\vspace{15pt}
		}
	\end{subtable}
	
	\begin{subtable}{\textwidth}
		\centering
		\begin{tabular}{l|ccccc}
			Rank &	1 &	2 &	3 &	4 &	5 \\
			WMD &	0.0838 &	0.0826 &	0.0821 &	0.0816 &	0.0816 \\
			\hline
			weight init. method &	Xavier &	Xavier &	Xavier &	Xavier &	Xavier \\
			max. init. weight &	1.96e-01 &	6.62e-01 &	8.24e-02 &	2.80e-02 &	1.66e-01 \\
			embed size &	276 &	193 &	103 &	420 &	221 \\
			RNN size &	227 &	155 &	229 &	443 &	80 \\
			post-image size &	268 &	417 &	454 &	213 &	260 \\
			post-image activation &	ReLU &	ReLU &	ReLU &	ReLU &	ReLU \\
			optimiser &	Adam &	Adam &	Adam &	Adam &	Adam \\
			learning rate &	2.64e-04 &	3.02e-04 &	2.71e-04 &	2.16e-04 &	3.90e-04 \\
			normalise image &	false &	false &	false &	false &	false \\
			weight decay weight &	3.01e-07 &	1.37e-10 &	5.79e-07 &	2.24e-07 &	6.30e-05 \\
			image dropout rate &	0.02 &	0.37 &	0.15 &	0.33 &	0.37 \\
			post-image dropout rate &	0.21 &	0.20 &	0.26 &	0.17 &	0.18 \\
			embedding dropout rate &	0.01 &	0.09 &	0.36 &	0.36 &	0.11 \\
			RNN dropout rate &	0.28 &	0.26 &	0.24 &	0.22 &	0.26 \\
			max. gradient norm &	6.86e+02 &	5.00e+01 &	1.34e+00 &	4.70e+02 &	3.02e+00 \\
			minibatch size &	237 &	110 &	148 &	215 &	182 \\
			beam width &	5 &	3 &	3 &	3 &	3 \\
		\end{tabular}
		\caption{
			\label{tbl:wi_hyperparams_merge}
			Top five hyperparameter combinations for the merge architecture.
		}
	\end{subtable}
	\caption{
		\label{tbl:wi_hyperparams}
		The top five hyperparameter combinations found for each architecture. An explanation of each hyperparameter is given in Subsection~\ref{sec:wi_hyperparameter_tuning}.
	}
\end{table}

The WMD of the best hyperparameters found for the merge architecture is higher than that of all the other architectures. No hyperparameter was set the same across all architectures and ranks. The following are architecture specific consistencies:

\paragraph{Init-inject:} The only hyperparameter that was set the same across all ranks in this architecture is the post-image layer not using an activation function. Using Pearson correlation on the whole 96 hyperparameters explored we found that the hyperparameters that correlated the most with WMD were using AdaDelta as an optimiser ($r=-0.476$) and image normalisation ($r=-0.228$).

\paragraph{Pre-inject:} No hyperparameter was set the same across all ranks in this architecture. Using Pearson correlation on the whole 96 hyperparameters explored we found that the hyperparameters that correlated the most with WMD were using AdaDelta as an optimiser ($r=-0.445$), using RMSProp as an optimiser ($r=0.349$), and RNN size ($r=0.254$).

\paragraph{Par-inject:} The learning rate should be around 4.4e-4, the image vector should not be normalised, the embedding layer should have a size of about 500 and have a very low dropout rate, and the minibatch size should be about 150. Using Pearson correlation on the whole 96 hyperparameters explored we found that the hyperparameters that correlated the most with WMD were using AdaDelta as an optimiser ($r=-0.393$), using RMSProp as an optimiser ($r=0.357$), weight decay weight ($r=-0.302$), embedding dropout rate ($r=-0.296$), embed size ($r=0.287$), and the maximum initial weight ($r=0.270$).

\paragraph{Merge:} The weight initialisation method should be Xavier, the optimiser should be Adam with a learning rate of around 3.0e-4, the image vector should not be normalised, the post-image layer should have a ReLU activation function, and the RNN dropout rate should be around 0.25. Using Pearson correlation on the whole 96 hyperparameters explored we found that the hyperparameters that correlated the most with WMD were using AdaDelta as an optimiser ($r=-0.579$), using Adam as an optimiser ($r=0.378$), and the maximum initial weight ($r=0.234$).\\

In general, these results suggest that AdaDelta is not recommended as an optimiser. In order to provide a better analysis of some of these hyperparameters, we plotted a scatter plot that compares the layer sizes to the resultant WMD values. All 96 hyperparameter combinations that were explored during the hyperparameter tuning process were plotted. It might be argued that since only 32 of the 96 hyperparameters were chosen randomly whilst 64 were chosen by Baysian optimisation, there is a risk of some bias in the plotted data points. Unfortunately, using just 32 data points is too small a sample size to get a meaningful plot, so we opted to use all the available hyperparameter combinations.

\begin{figure}
	\centering
	\begin{subfigure}{0.4\textwidth}
		\centering
		\includegraphics[scale=0.5]{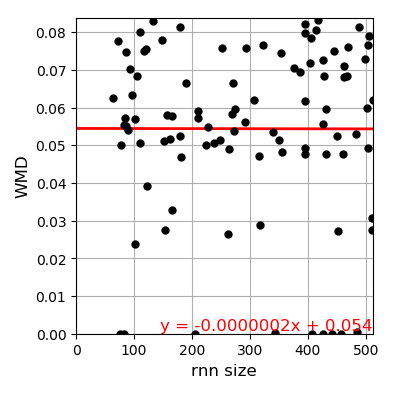}
		\caption{
			\label{fig:wi_results_hyp_rnnsize_init}
			The init-inject architecture.
		}
	\end{subfigure}
	\quad
	\begin{subfigure}{0.4\textwidth}
		\centering
		\includegraphics[scale=0.5]{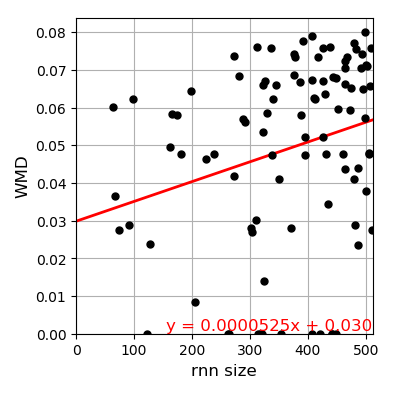}
		\caption{
			\label{fig:wi_results_hyp_rnnsize_pre}
			The pre-inject architecture.
		}
	\end{subfigure}
	
	\begin{subfigure}{0.4\textwidth}
		\centering
		\includegraphics[scale=0.5]{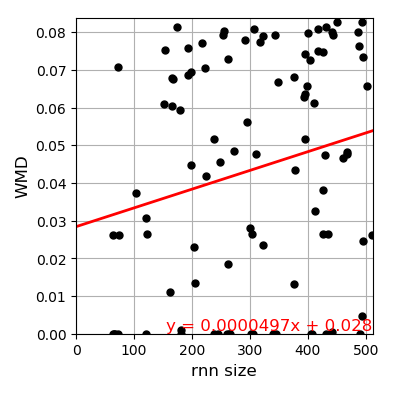}
		\caption{
			\label{fig:wi_results_hyp_rnnsize_par}
			The par-inject architecture.
		}
	\end{subfigure}
	\quad
	\begin{subfigure}{0.4\textwidth}
		\centering
		\includegraphics[scale=0.5]{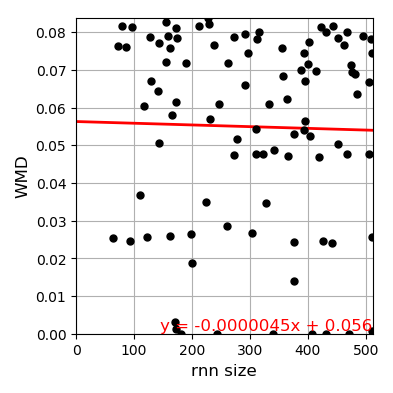}
		\caption{
			\label{fig:wi_results_hyp_rnnsize_merge}
			The merge architecture.
		}
	\end{subfigure}
	\caption{
		\label{fig:wi_results_hyp_rnnsize}
		Scatter plots of how the RNN size relates to the WMD for each architecture. A linear trend line is fitted and shown as well.
	}
\end{figure}

\paragraph{RNN sizes:} The RNN size, plotted in Figure~\ref{fig:wi_results_hyp_rnnsize}, significantly affects the amount of memory needed to store the neural network since the last layer has a number of weights equal to the RNN size multiplied by the vocabulary size, the vocabulary size being generally large. The last layer's number of weights equals $r \times v$ in inject architectures whilst in merge architectures have $(r + p) \times v$, where $r$ is the RNN size, $p$ is the post-image size and $v$ is the vocabulary size.

Init-inject and merge don't seem to be affected much by the size of the RNN state, although merge does tend to work slightly better on smaller state sizes. This could be because merge is more likely to overfit at its RNN due to it only needing to store linguistic information rather than a mix of visual and linguistic information like the inject architectures. As we'll see, init-inject seems to be stable across all model sizes, which is good as it reaches peak performance with a small number of parameters.

On the other hand pre-inject and par-inject are more sensitive to the RNN state size. In the case of the pre-inject architecture, this could be in order to better remember the first word (the image). The fact that init-inject does not have the same effect (requiring a large RNN state to remember the image in the initial state) might be explained in the next chapter which shows that GRUs are very sensitive to their initial state. Par-inject, as will be seen, requires a large model in general in order to work well.

\begin{figure}
	\centering
	\begin{subfigure}{0.4\textwidth}
		\centering
		\includegraphics[scale=0.5]{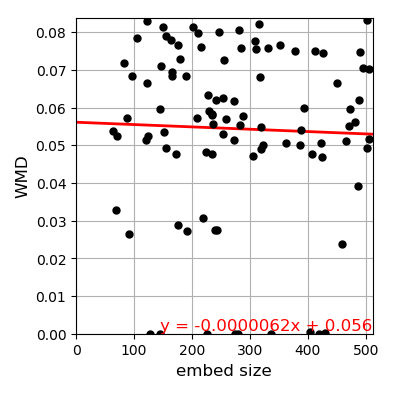}
		\caption{
			\label{fig:wi_results_hyp_embedsize_init}
			The init-inject architecture.
		}
	\end{subfigure}
	\quad
	\begin{subfigure}{0.4\textwidth}
		\centering
		\includegraphics[scale=0.5]{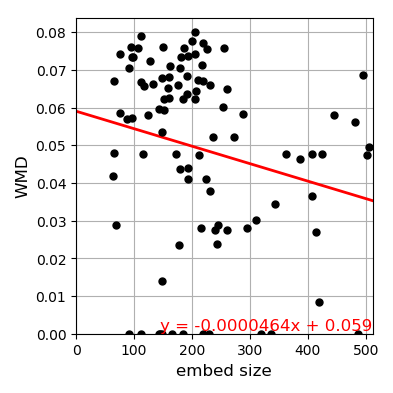}
		\caption{
			\label{fig:wi_results_hyp_embedsize_pre}
			The pre-inject architecture.
		}
	\end{subfigure}
	
	\begin{subfigure}{0.4\textwidth}
			\centering
			\includegraphics[scale=0.5]{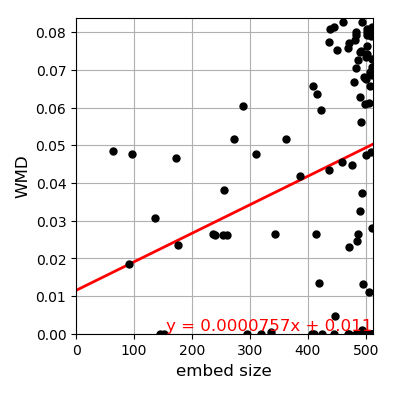}
			\caption{
				\label{fig:wi_results_hyp_embedsize_par}
				The par-inject architecture.
			}
		\end{subfigure}
	\quad
	\begin{subfigure}{0.4\textwidth}
		\centering
		\includegraphics[scale=0.5]{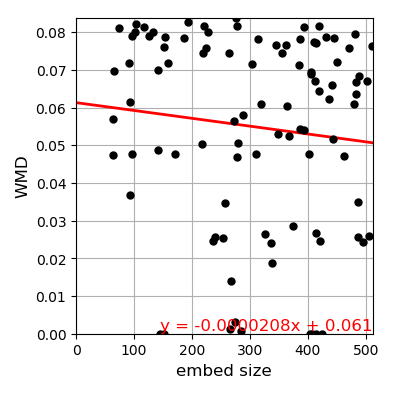}
		\caption{
			\label{fig:wi_results_hyp_embedsize_merge}
			The merge architecture.
		}
	\end{subfigure}
	\caption{
		\label{fig:wi_results_hyp_embedsize}
		Scatter plots of how the embed size relates to the WMD for each architecture. A linear trend line is fitted and shown as well.
	}
\end{figure}

\paragraph{Embed sizes:} The embed size, plotted in Figure~\ref{fig:wi_results_hyp_embedsize}, also significantly affects the amount of memory needed to store the neural network since the embedding matrix has a number of weights equal to the embed size multiplied by the vocabulary size, the vocabulary being generally large. The embedding matrix's number of weights equals $e \times v$, where $e$ is the embed size and $v$ is the vocabulary size.

Again, init-inject is not affected much by the embedding layer size, evidence of its efficiency in parameter use. Merge is negatively correlated, probably due to overfitting, just like for the state size.

The fact that pre-inject requires a large state size but a small word size (and image size, since they have to be equal in pre-inject) could be in order to be able to pack more information in the RNN state by putting smaller items in a larger memory. Par-inject, again, requires a large model to perform well.

\begin{figure}
	\centering
	\begin{subfigure}{0.4\textwidth}
		\centering
		\includegraphics[scale=0.5]{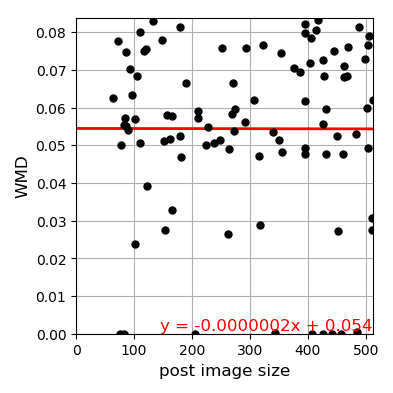}
		\caption{
			\label{fig:wi_results_hyp_postimagesize_init}
			The init-inject architecture.
		}
	\end{subfigure}
	\quad
	\begin{subfigure}{0.4\textwidth}
		\centering
		\includegraphics[scale=0.5]{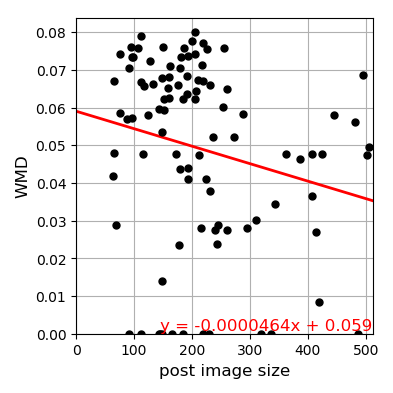}
		\caption{
			\label{fig:wi_results_hyp_postimagesize_pre}
			The pre-inject architecture.
		}
	\end{subfigure}
	
	\begin{subfigure}{0.4\textwidth}
		\centering
		\includegraphics[scale=0.5]{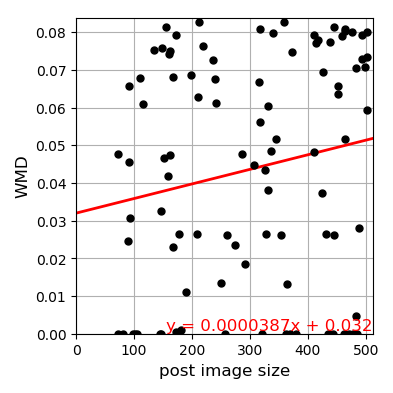}
		\caption{
			\label{fig:wi_results_hyp_postimagesize_par}
			The par-inject architecture.
		}
	\end{subfigure}
	\quad
	\begin{subfigure}{0.4\textwidth}
		\centering
		\includegraphics[scale=0.5]{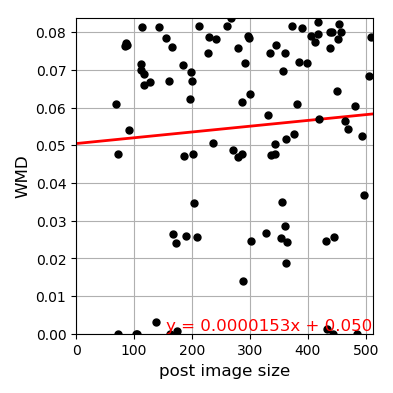}
		\caption{
			\label{fig:wi_results_hyp_postimagesize_merge}
			The merge architecture.
		}
	\end{subfigure}
	\caption{
		\label{fig:wi_results_hyp_postimagesize}
		Scatter plots of how the post-image size relates to the WMD for each architecture. A linear trend line is fitted and shown as well.
	}
\end{figure}

\paragraph{Post-image sizes:} The post-image size, plotted in Figure~\ref{fig:wi_results_hyp_postimagesize}, affects the amount of memory needed to store the neural network if the original image vector size is large, which in this case it is since it is 4\,096-elements long. The post-image layer's number of weights equals $i \times p$, where $i$ is the image vector size and $p$ is the post-image size.

Given that init-inject is not affected much by the state size then it must also not be affect much by the image size, since these two sizes are tied in the init-inject architecture. Similarly, given that pre-inject works worse on large embedding sizes then it must also work worse on large image sizes.

Merge seems to work better on larger image sizes, although it is unclear why this is. Again, par-inject requires more parameters to work well.

\begin{figure}
	\centering
	\begin{subfigure}{0.4\textwidth}
		\centering
		\includegraphics[scale=0.5]{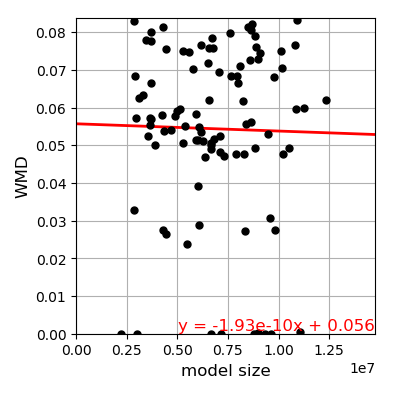}
		\caption{
			\label{fig:wi_results_hyp_modelsize_init}
			The init-inject architecture.
		}
	\end{subfigure}
	\quad
	\begin{subfigure}{0.4\textwidth}
		\centering
		\includegraphics[scale=0.5]{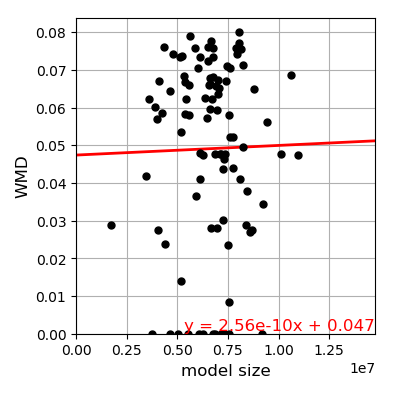}
		\caption{
			\label{fig:wi_results_hyp_modelsize_pre}
			The pre-inject architecture.
		}
	\end{subfigure}
	
	\begin{subfigure}{0.4\textwidth}
		\centering
		\includegraphics[scale=0.5]{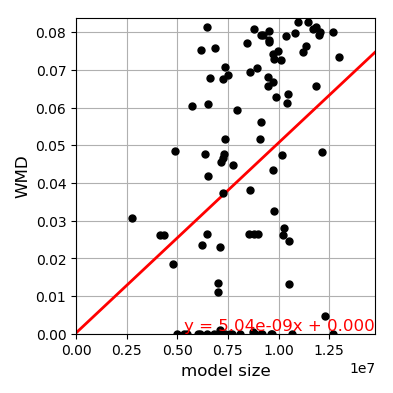}
		\caption{
			\label{fig:wi_results_hyp_modelsize_par}
			The par-inject architecture.
		}
	\end{subfigure}
	\quad
	\begin{subfigure}{0.4\textwidth}
		\centering
		\includegraphics[scale=0.5]{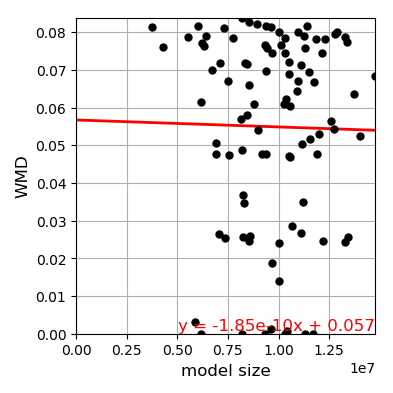}
		\caption{
			\label{fig:wi_results_hyp_modelsize_merge}
			The merge architecture.
		}
	\end{subfigure}
	\caption{
		\label{fig:wi_results_hyp_modelsize}
		Scatter plots of how the full model size (number of parameters) relates to the WMD for each architecture. The vocabulary size of MSCOCO is assumed. A linear trend line is fitted and shown as well.
	}
\end{figure}

\paragraph{Full model sizes:} We finally put all the layers together and see how the WMD changes as the total number of parameters changes. We calculate what the model size will be if the vocabulary size was that of MSCOCO. This is plotted in Figure~\ref{fig:wi_results_hyp_modelsize}.

We can see that par-inject does not work well with small models whilst all the other architectures can work well with different model sizes. Merge is shifted towards larger models in general which means that the merge architecture results in a large model more often than not, although it can work well with smaller models unlike par-inject.

We have a hypothesis for why init- and pre-inject can work well with small models whilst par-inject can't, which is that the former two architectures use shared representations whilst par-inject does not. Due to the fact that the image representation is shared is the RNN state representation in init-inject, as the state representation gets better, so might the image representation. Ditto for pre-inject but with the word representation rather than the state representation. This could result in more effective training at smaller model sizes whereas par-inject would underfit. We do not have a hypothesis to explain why the merge architecture does not also underfit in spite of not having shared representations.

\subsubsection{Interim summary}

\begin{itemize}
	\item The init-inject architecture is stable across layer sizes and can achieve both high and low performance at every model size. This could be because it has a shared representation between the state and the image which might result in better a representation for both.
	
	\item The pre-inject architecture needs a large RNN size but a small image size. Probably in order to better remember information about the image given that it is the first word in the sequence.
	
	\item The par-inject architecture needs to be larger at every layer to increase in performance. Having separate channels for the image and words into the RNN might require more processing in order to be able to create a single representation in the RNN. The shared representations of init- and pre-inject seem to help them work with less parameters.
	
	\item The merge architecture might overfit with a large RNN given that it only needs to store linguistic information in there.
\end{itemize}

\subsection{Experimental results}

We now look at the results produced by the actual experiments after training and evaluating the architectures using the best hyperparameters found in the hyperparameter search (the rank 1 columns in Table~\ref{tbl:wi_hyperparams}).

\paragraph{Probability metrics:} Table~\ref{tbl:wi_results_exp_prb} shows the result of calculating different aggregations of the probability and perplexity of test set sentences according to the trained neural networks. Init-inject and par-inject perform the best at all of these measures, whilst pre-inject and merge seem to perform poorly.

\paragraph{Quality metrics:} Table~\ref{tbl:wi_results_exp_qty} shows the result of calculating different caption quality metrics on the generated sentences for the test set images. The ceiling is there as an anchor point to know what the maximum score for each metric is. It is computed by treating the captions provided with the images in the test set themselves as if they were generated (since there are multiple captions with each image, a random caption was selected). All four architectures perform similarly but, again, init-inject and par-inject outperform pre-inject and merge.

\paragraph{Diversity metrics:} Table~\ref{tbl:wi_results_exp_div} shows the amount of variation in the generated sentences in terms of words used. The ceiling is there as an anchor point to know what the human-written test set sentences would score on each metric (it is computed just like for the quality metrics). The percentage of vocabulary used is a metric for measuring how much of the available vocabulary is exploited when generating sentences as opposed to relying on just high frequency words (which would be exposed to the neural network more often during training). In this case, init-inject and par-inject perform the best with pre-inject performing the worst, which corresponds with the caption quality metrics.

Furthermore, it turns out that the amount of minimum frequency words used correlates with the amount of the vocabulary used, that is, the more words are used, the more rare words are used (even though less than 15\% of the vocabulary is used). This means that a well performing neural network is capable of using words with a frequency of just 5 (the minimum frequency threshold used in the experiments) and so it might be worthwhile to include even rarer words in the vocabulary. It is concerning that, compared to the human written captions, generated sentences use very little of the available vocabulary, hinting that the generated captions tend to be stereotyped and perhaps `robotic' sounding. This observation was in fact made by \citet{Devlin2015} who found that the generated captions in an init-inject architecture are very stereotypical, although unlike what is reported in that paper, none of our generated sentences were found in the training set. All the generated sentences have similar average lengths. Par-inject generated the most unique sentences whilst pre-inject generated the most duplicated sentences as well as using exclusively very frequent words.

\paragraph{Retrieval metrics:} Table~\ref{tbl:wi_results_exp_ret} shows the result of using the trained models to search for a test set image using its corresponding caption. Init-inject performs the best at this task whilst pre-inject performs the worst. Here merge performs very similarly to init-inject and par-inject whilst pre-inject performs very poorly.

\paragraph{Miscellaneous metrics:} Table~\ref{tbl:wi_results_exp_msc} shows miscellaneous results such as the number of parameters in the models and the training time. The smallest architectures are merge and pre-inject, although given how badly pre-inject performed in general it is likely that this is not the optimum size for such an architecture and a more extensive hyperparameter search might have resulted in a bigger model which performs better. The fastest models to train, both in terms of number of epochs and training time, are init-inject and merge. Par-inject is the all-round worst model at both size and training speed.

\begin{table}
	\centering
	\begin{subtable}{\textwidth}
		\centering
		\begin{small}
			\begin{tabular}{l|cccc}
				 &	Init-inject &	Pre-inject &	Par-inject &	Merge \\
				\hline
			Mean prob. &	{1.8e-05} (2.9e-06) &	{1.7e-05} (4.6e-06) &	\textbf{1.9e-05} (7.6e-07) &	{1.0e-05} (1.5e-06) \\
			Median prob. &	\textbf{3.1e-13} (1.9e-14) &	{2.7e-14} (2.9e-15) &	{7.8e-14} (5.4e-15) &	{1.4e-14} (1.5e-15) \\
			Geomean prob. &	\textbf{1.4e-14} (3.6e-16) &	{1.3e-15} (1.3e-16) &	{3.1e-15} (1.5e-16) &	{6.2e-16} (7.0e-17) \\
			Mean pplx &	\textbf{24.21} (0.36) &	{29.94} (0.30) &	{29.00} (0.15) &	{33.29} (0.80) \\
			Median pplx &	\textbf{12.91} (0.04) &	{16.56} (0.17) &	{14.71} (0.12) &	{16.78} (0.19) \\
			Geomean pplx &	\textbf{14.28} (0.02) &	{17.61} (0.15) &	{16.25} (0.08) &	{18.62} (0.16) \\
			\end{tabular}
			\caption{
				\label{tbl:wi_results_exp_prb_flickr8k}
				Results for Flickr8K.
				\vspace{15pt}
			}
		\end{small}
	\end{subtable}
	
	\begin{subtable}{\textwidth}
		\centering
		\begin{small}
			\begin{tabular}{l|cccc}
				 &	Init-inject &	Pre-inject &	Par-inject &	Merge \\
				\hline
			Mean prob. &	{4.4e-06} (9.4e-07) &	{2.9e-06} (7.6e-07) &	\textbf{6.3e-06} (1.0e-06) &	{2.1e-06} (3.1e-07) \\
			Median prob. &	\textbf{2.4e-15} (3.5e-16) &	{1.1e-16} (2.4e-17) &	{1.2e-15} (9.8e-17) &	{1.1e-16} (1.5e-17) \\
			Geomean prob. &	\textbf{1.9e-17} (7.9e-19) &	{7.8e-19} (1.5e-19) &	{9.2e-18} (9.9e-19) &	{7.0e-19} (3.4e-20) \\
			Mean pplx &	\textbf{33.59} (0.74) &	{43.66} (1.15) &	{36.93} (0.69) &	{55.59} (2.28) \\
			Median pplx &	\textbf{15.07} (0.08) &	{19.68} (0.38) &	{15.83} (0.21) &	{19.20} (0.09) \\
			Geomean pplx &	\textbf{17.26} (0.07) &	{22.06} (0.35) &	{18.17} (0.14) &	{22.00} (0.08) \\
			\end{tabular}
			\caption{
				\label{tbl:wi_results_exp_prb_flickr30k}
				Results for Flickr30K.
				\vspace{15pt}
			}
		\end{small}
	\end{subtable}
	
	\begin{subtable}{\textwidth}
		\centering
		\begin{small}
			\begin{tabular}{l|cccc}
				 &	Init-inject &	Pre-inject &	Par-inject &	Merge \\
				\hline
			Mean prob. &	\textbf{5.0e-05} (4.5e-06) &	{3.1e-05} (9.1e-07) &	{3.5e-05} (2.4e-06) &	{1.6e-05} (7.5e-07) \\
			Median prob. &	\textbf{1.9e-11} (6.7e-13) &	{1.2e-12} (6.1e-14) &	{1.3e-11} (7.6e-13) &	{3.1e-12} (2.5e-13) \\
			Geomean prob. &	\textbf{1.5e-12} (5.5e-14) &	{8.5e-14} (3.1e-15) &	{9.4e-13} (2.5e-14) &	{1.6e-13} (6.1e-15) \\
			Mean pplx &	\textbf{19.57} (0.05) &	{29.41} (0.14) &	{21.73} (0.17) &	{31.26} (0.61) \\
			Median pplx &	\textbf{9.20} (0.05) &	{11.76} (0.05) &	{9.45} (0.04) &	{10.73} (0.04) \\
			Geomean pplx &	\textbf{10.54} (0.03) &	{13.56} (0.04) &	{10.98} (0.03) &	{12.78} (0.04) \\
			\end{tabular}
			\caption{
				\label{tbl:wi_results_exp_prb_mscoco}
				Results for MSCOCO.
			}
		\end{small}
	\end{subtable}
	
	\caption{
		\label{tbl:wi_results_exp_prb}
		Results of the probability metrics. Legend: prob. - probability, geomean - geometric mean, pplx - perplexity.
	}
\end{table}

\begin{table}
	\centering
	\begin{subtable}{\textwidth}
		\centering
		\begin{small}
			\begin{tabular}{l|cccc|c}
				 &	Init-inject &	Pre-inject &	Par-inject &	Merge &	Ceiling \\
				\hline
				BLEU-1 &	\textbf{0.605} (0.005) &	{0.602} (0.005) &	{0.591} (0.005) &	{0.602} (0.011) &	{1.000} (0.000) \\
				BLEU-2 &	\textbf{0.421} (0.006) &	{0.413} (0.003) &	{0.410} (0.005) &	{0.411} (0.010) &	{1.000} (0.000) \\
				BLEU-3 &	\textbf{0.285} (0.004) &	{0.277} (0.005) &	{0.274} (0.005) &	{0.269} (0.009) &	{1.000} (0.000) \\
				BLEU-4 &	\textbf{0.191} (0.004) &	{0.183} (0.005) &	{0.181} (0.004) &	{0.174} (0.007) &	{1.000} (0.000) \\
				METEOR &	{0.194} (0.002) &	{0.187} (0.001) &	\textbf{0.196} (0.002) &	{0.190} (0.001) &	{1.000} (0.000) \\
				ROUGE-L &	{0.446} (0.004) &	{0.440} (0.004) &	\textbf{0.446} (0.003) &	{0.439} (0.004) &	{1.000} (0.000) \\
				CIDEr &	{0.474} (0.017) &	{0.441} (0.011) &	\textbf{0.476} (0.012) &	{0.457} (0.011) &	{2.663} (0.007) \\
				SPICE &	\textbf{0.134} (0.002) &	{0.127} (0.001) &	{0.133} (0.003) &	{0.128} (0.002) &	{0.439} (0.002) \\
				WMD &	\textbf{0.140} (0.003) &	{0.137} (0.004) &	{0.137} (0.002) &	{0.137} (0.003) &	{1.000} (0.000) \\
			\end{tabular}
			\caption{
				\label{tbl:wi_results_exp_qty_flickr8k}
				Results for Flickr8K.
				\vspace{15pt}
			}
		\end{small}
	\end{subtable}
	
	\begin{subtable}{\textwidth}
		\centering
		\begin{small}
			\begin{tabular}{l|cccc|c}
				 &	Init-inject &	Pre-inject &	Par-inject &	Merge &	Ceiling \\
				\hline
				BLEU-1 &	{0.602} (0.018) &	{0.604} (0.005) &	{0.608} (0.006) &	\textbf{0.618} (0.003) &	{1.000} (0.000) \\
				BLEU-2 &	{0.410} (0.015) &	{0.406} (0.005) &	{0.416} (0.007) &	\textbf{0.421} (0.002) &	{1.000} (0.000) \\
				BLEU-3 &	{0.275} (0.011) &	{0.270} (0.004) &	{0.281} (0.007) &	\textbf{0.282} (0.002) &	{1.000} (0.000) \\
				BLEU-4 &	{0.186} (0.008) &	{0.179} (0.005) &	{0.189} (0.007) &	\textbf{0.189} (0.002) &	{1.000} (0.000) \\
				METEOR &	{0.174} (0.001) &	{0.167} (0.001) &	\textbf{0.178} (0.001) &	{0.171} (0.000) &	{1.000} (0.000) \\
				ROUGE-L &	{0.419} (0.004) &	{0.413} (0.003) &	\textbf{0.426} (0.003) &	{0.421} (0.001) &	{1.000} (0.000) \\
				CIDEr &	{0.361} (0.011) &	{0.337} (0.005) &	\textbf{0.381} (0.010) &	{0.371} (0.004) &	{2.541} (0.007) \\
				SPICE &	{0.112} (0.002) &	{0.108} (0.001) &	\textbf{0.116} (0.001) &	{0.110} (0.001) &	{0.416} (0.002) \\
				WMD &	{0.121} (0.001) &	{0.118} (0.002) &	\textbf{0.122} (0.002) &	{0.120} (0.002) &	{1.000} (0.000) \\
			\end{tabular}
			\caption{
				\label{tbl:wi_results_exp_qty_flickr30k}
				Results for Flickr30K.
				\vspace{15pt}
			}
		\end{small}
	\end{subtable}
	
	\begin{subtable}{\textwidth}
		\centering
		\begin{small}
			\begin{tabular}{l|cccc|c}
				 &	Init-inject &	Pre-inject &	Par-inject &	Merge &	Ceiling \\
				\hline
				BLEU-1 &	{0.668} (0.004) &	{0.651} (0.002) &	\textbf{0.678} (0.003) &	{0.664} (0.003) &	{1.000} (0.000) \\
				BLEU-2 &	{0.489} (0.006) &	{0.472} (0.002) &	\textbf{0.501} (0.002) &	{0.487} (0.003) &	{1.000} (0.000) \\
				BLEU-3 &	{0.357} (0.006) &	{0.341} (0.002) &	\textbf{0.364} (0.003) &	{0.353} (0.004) &	{1.000} (0.000) \\
				BLEU-4 &	{0.263} (0.006) &	{0.251} (0.002) &	\textbf{0.266} (0.003) &	{0.257} (0.004) &	{1.000} (0.000) \\
				METEOR &	\textbf{0.225} (0.001) &	{0.215} (0.001) &	{0.225} (0.001) &	{0.217} (0.001) &	{1.000} (0.000) \\
				ROUGE-L &	{0.494} (0.003) &	{0.482} (0.001) &	\textbf{0.498} (0.001) &	{0.488} (0.002) &	{1.000} (0.000) \\
				CIDEr &	{0.790} (0.014) &	{0.739} (0.005) &	\textbf{0.806} (0.006) &	{0.764} (0.007) &	{2.715} (0.003) \\
				SPICE &	{0.153} (0.002) &	{0.144} (0.001) &	\textbf{0.155} (0.001) &	{0.148} (0.001) &	{0.431} (0.001) \\
				WMD &	\textbf{0.179} (0.002) &	{0.170} (0.001) &	{0.176} (0.002) &	{0.172} (0.001) &	{1.000} (0.000) \\
			\end{tabular}
			\caption{
				\label{tbl:wi_results_exp_qty_mscoco}
				Results for MSCOCO.
			}
		\end{small}
	\end{subtable}
	
	\caption{
		\label{tbl:wi_results_exp_qty}
		Results of the caption quality metrics. The ceiling is the result of using these metrics on the test set sentences themselves.
	}
\end{table}

\begin{table}
	\centering
	\begin{subtable}{\textwidth}
		\centering
		\begin{small}
			\begin{tabular}{l|cccc|c}
				 &	Init-inject &	Pre-inject &	Par-inject &	Merge &	Ceiling \\
				\hline
				Vocab. used &	{14.3\%} (0.6\%) &	{10.0\%} (0.6\%) &	\textbf{14.3\%} (0.5\%) &	{11.5\%} (0.3\%) &	{47.3\%} (0.3\%) \\
				Min. freq. vocab. &	\textbf{5.0} (0.0) &	{18.2} (4.5) &	{5.2} (0.4) &	{8.0} (2.2) &	{5.0} (0.0) \\
				Mean sent. len. &	{10.1} (0.1) &	{9.4} (0.2) &	{10.6} (0.1) &	{10.0} (0.2) &	{10.9} (0.0) \\
				Reused sents. &	{0.0\%} (0.0\%) &	{0.0\%} (0.0\%) &	{0.0\%} (0.0\%) &	{0.0\%} (0.0\%) &	{0.0\%} (0.0\%) \\
				Unique sents. &	{80.3\%} (1.3\%) &	{71.2\%} (3.8\%) &	\textbf{91.1\%} (1.0\%) &	{86.1\%} (1.8\%) &	{99.9\%} (0.0\%) \\
			\end{tabular}
			\caption{
				\label{tbl:wi_results_exp_div_flickr8k}
				Results for Flickr8K.
				\vspace{15pt}
			}
		\end{small}
	\end{subtable}
	
	\begin{subtable}{\textwidth}
		\centering
		\begin{small}
			\begin{tabular}{l|cccc|c}
				 &	Init-inject &	Pre-inject &	Par-inject &	Merge &	Ceiling \\
				\hline
				Vocab. used &	{5.6\%} (0.2\%) &	{4.0\%} (0.2\%) &	\textbf{6.3\%} (0.3\%) &	{5.2\%} (0.2\%) &	{24.1\%} (0.2\%) \\
				Min. freq. vocab. &	\textbf{7.8} (3.7) &	{77.8} (26.3) &	{10.4} (3.3) &	{10.4} (3.2) &	{5.0} (0.0) \\
				Mean sent. len. &	{11.0} (0.6) &	{10.4} (0.2) &	{11.2} (0.2) &	{10.3} (0.1) &	{12.4} (0.1) \\
				Reused sents. &	{0.0\%} (0.0\%) &	{0.0\%} (0.0\%) &	{0.0\%} (0.0\%) &	{0.0\%} (0.0\%) &	{0.0\%} (0.0\%) \\
				Unique sents. &	{74.2\%} (1.2\%) &	{63.6\%} (3.8\%) &	\textbf{84.2\%} (1.6\%) &	{75.0\%} (1.7\%) &	{100.0\%} (0.0\%) \\
			\end{tabular}
			\caption{
				\label{tbl:wi_results_exp_div_flickr30k}
				Results for Flickr30K.
				\vspace{15pt}
			}
		\end{small}
	\end{subtable}
	
	\begin{subtable}{\textwidth}
		\centering
		\begin{small}
			\begin{tabular}{l|cccc|c}
				 &	Init-inject &	Pre-inject &	Par-inject &	Merge &	Ceiling \\
				\hline
				Vocab. used &	{7.6\%} (0.2\%) &	{5.1\%} (0.1\%) &	\textbf{8.4\%} (0.2\%) &	{7.3\%} (0.2\%) &	{37.2\%} (0.2\%) \\
				Min. freq. vocab. &	{10.2} (3.1) &	{115.8} (35.8) &	\textbf{7.6} (1.7) &	{11.6} (3.3) &	{5.0} (0.0) \\
				Mean sent. len. &	{9.3} (0.1) &	{9.1} (0.0) &	{9.2} (0.1) &	{9.0} (0.0) &	{10.5} (0.0) \\
				Reused sents. &	{0.0\%} (0.0\%) &	{0.0\%} (0.0\%) &	{0.0\%} (0.0\%) &	{0.0\%} (0.0\%) &	{0.0\%} (0.0\%) \\
				Unique sents. &	{43.6\%} (1.1\%) &	{28.9\%} (1.5\%) &	\textbf{54.2\%} (0.6\%) &	{46.9\%} (0.8\%) &	{99.8\%} (0.1\%) \\
			\end{tabular}
			\caption{
				\label{tbl:wi_results_exp_div_mscoco}
				Results for MSCOCO.
			}
		\end{small}
	\end{subtable}
	
	\caption{
		\label{tbl:wi_results_exp_div}
		Results of the diversity metrics. The ceiling is the result of using these metrics on the test set sentences themselves. Legend: vocab. - vocabulary, min. - minimum, freq. - frequency, sent. - sentence, len. - length, min. freq. vocab. - the minimum training set frequency of the words used in generated sentences.
	}
\end{table}

\begin{table}
	\centering
	\begin{subtable}{\textwidth}
		\centering
		\begin{small}
			\begin{tabular}{l|cccc}
				 &	Init-inject &	Pre-inject &	Par-inject &	Merge \\
				\hline
				R@1 &	\textbf{18.4\%} (0.8\%) &	{10.6\%} (0.2\%) &	{15.8\%} (0.6\%) &	{15.7\%} (0.9\%) \\
				R@5 &	\textbf{43.6\%} (0.9\%) &	{32.3\%} (1.2\%) &	{39.6\%} (1.0\%) &	{37.9\%} (1.1\%) \\
				R@10 &	\textbf{57.6\%} (1.3\%) &	{45.4\%} (2.0\%) &	{54.0\%} (1.7\%) &	{50.3\%} (1.3\%) \\
				Median rank &	\textbf{7.4} (0.5) &	{12.8} (1.0) &	{9.0} (0.6) &	{10.4} (0.8) \\
			\end{tabular}
			\caption{
				\label{tbl:wi_results_exp_ret_flickr8k}
				Results for Flickr8K.
				\vspace{15pt}
			}
		\end{small}
	\end{subtable}
	
	\begin{subtable}{\textwidth}
		\centering
		\begin{small}
			\begin{tabular}{l|cccc}
				 &	Init-inject &	Pre-inject &	Par-inject &	Merge \\
				\hline
				R@1 &	\textbf{21.8\%} (0.6\%) &	{12.8\%} (1.0\%) &	{20.5\%} (1.4\%) &	{19.4\%} (1.3\%) \\
				R@5 &	\textbf{48.9\%} (0.8\%) &	{35.1\%} (0.6\%) &	{46.1\%} (1.5\%) &	{44.6\%} (1.4\%) \\
				R@10 &	\textbf{60.3\%} (1.3\%) &	{48.6\%} (1.3\%) &	{58.5\%} (1.4\%) &	{55.5\%} (0.7\%) \\
				Median rank &	\textbf{6.0} (0.0) &	{11.4} (0.8) &	{6.6} (0.5) &	{7.7} (0.4) \\
			\end{tabular}
			\caption{
				\label{tbl:wi_results_exp_ret_flickr30k}
				Results for Flickr30K.
				\vspace{15pt}
			}
		\end{small}
	\end{subtable}
	
	\begin{subtable}{\textwidth}
		\centering
		\begin{small}
			\begin{tabular}{l|cccc}
				 &	Init-inject &	Pre-inject &	Par-inject &	Merge \\
				\hline
				R@1 &	\textbf{13.3\%} (0.2\%) &	{6.8\%} (0.5\%) &	{11.7\%} (0.3\%) &	{11.2\%} (0.2\%) \\
				R@5 &	\textbf{34.4\%} (0.4\%) &	{21.9\%} (0.2\%) &	{31.6\%} (0.5\%) &	{29.8\%} (0.4\%) \\
				R@10 &	\textbf{46.3\%} (0.5\%) &	{33.1\%} (0.2\%) &	{44.0\%} (0.6\%) &	{41.8\%} (0.2\%) \\
				Median rank &	\textbf{12.6} (0.5) &	{22.7} (0.7) &	{14.2} (0.4) &	{16.0} (0.0) \\
			\end{tabular}
			\caption{
				\label{tbl:wi_results_exp_ret_mscoco}
				Results for MSCOCO.
			}
		\end{small}
	\end{subtable}
	
	\caption{
		\label{tbl:wi_results_exp_ret}
		Results of the retrieval metrics. Legend: R@$n$ - recall at $n$.
	}
\end{table}

\begin{table}
	\centering
	\begin{subtable}{\textwidth}
		\centering
		\begin{small}
			\begin{tabular}{l|cccc}
				 &	Init-inject &	Pre-inject &	Par-inject &	Merge \\
				\hline
				Number of params. &	{5199452} (0) &	{3682810} (0) &	{5628326} (0) &	\textbf{3396151} (0) \\
				Number of epochs &	\textbf{10.6} (0.5) &	{16.6} (1.4) &	{22.4} (1.0) &	{14.8} (0.7) \\
				Training time (s) &	{327.2} (14.2) &	{652.6} (52.5) &	{845.4} (38.5) &	\textbf{310.8} (14.4) \\
			\end{tabular}
			\caption{
				\label{tbl:wi_results_exp_msc_flickr8k}
				Results for Flickr8K.
				\vspace{15pt}
			}
		\end{small}
	\end{subtable}
	
	\begin{subtable}{\textwidth}
		\centering
		\begin{small}
			\begin{tabular}{l|cccc}
				 &	Init-inject &	Pre-inject &	Par-inject &	Merge \\
				\hline
				Number of params. &	{9629462} (0) &	\textbf{7073860} (0) &	{10173776} (0) &	{7109471} (0) \\
				Number of epochs &	\textbf{9.0} (0.6) &	{17.8} (3.1) &	{18.8} (2.5) &	{11.0} (0.9) \\
				Training time (s) &	\textbf{3518.4} (243.8) &	{8677.2} (1484.0) &	{8813.2} (1144.6) &	{3611.4} (282.9) \\
			\end{tabular}
			\caption{
				\label{tbl:wi_results_exp_msc_flickr30k}
				Results for Flickr30K.
				\vspace{15pt}
			}
		\end{small}
	\end{subtable}
	
	\begin{subtable}{\textwidth}
		\centering
		\begin{small}
			\begin{tabular}{l|cccc}
				 &	Init-inject &	Pre-inject &	Par-inject &	Merge \\
				\hline
				Number of params. &	{10903205} (0) &	\textbf{8048875} (0) &	{11480711} (0) &	{8177147} (0) \\
				Number of epochs &	\textbf{11.2} (1.5) &	{17.0} (1.1) &	{20.8} (2.0) &	{12.2} (1.0) \\
				Training time (s) &	{9255.2} (1177.2) &	{17372.0} (1113.9) &	{20141.0} (1950.6) &	\textbf{8224.4} (647.7) \\
			\end{tabular}
			\caption{
				\label{tbl:wi_results_exp_msc_mscoco}
				Results for MSCOCO.
			}
		\end{small}
	\end{subtable}
	
	\caption{
		\label{tbl:wi_results_exp_msc}
		Results of the miscellaneous metrics. Legend: params. - Parameters (weights and biases).
	}
\end{table}

\FloatBarrier
\subsubsection{Interim summary}

\begin{itemize}
	\item Init-inject and par-inject perform the best all round. Almost all the best results went to either init-inject or par-inject with a small portion going to merge and a couple going to pre-inject. Init-inject performs best at probability and retrieval metrics (which are related) whilst par-inject performs best at the diversity metrics (but init-inject was best according to the WMD metric). The best quality metrics were shared between init-inject and par-inject.
	
	\item Not surprisingly, the two worst performing models also happen to be the smallest.
	
	\item The WMD quality metric seems to correlate well with the probability and retrieval metrics. Its only mismatch is for the Flickr30K results but then init-inject and par-inject are very close in terms of WMD, geometric mean perplexity, and R@1 metrics.
\end{itemize}

\clearpage


\section{Human evaluation}

We opted to also include a human evaluation of the different captions that were generated. Given that a caption generator is ultimately meant to be used by humans, a human evaluation of the system's output is important although, unfortunately, seldom done in practice due to it being a time-consuming process. \citet{Elliott2013} evaluated their system by asking whether the generated descriptions were grammatical, correctly described the action in the image, and correctly described the scene in the image. \citet{Mitchell2012} also evaluated their system by asking whether the generated descriptions were grammatical and correctly describe the image but included also whether the main aspects were described, whether the order of objects was reasonable, and whether it sounds like a human wrote it. We opted to only ask about accuracy and fluency (grammaticality) in order to speed up the annotation process.

Five annotators were recruited to evaluate the generated captions of 200 images that were randomly selected from the MSCOCO test set. Given that each system was run five times, thereby generating five separate captions for each image, we randomly selected one of the five captions for each of the 200 images and for each of the 4 architectures, resulting in $4 \times 200 \times 1 = 800$ captions, and put them in an online database which were shown to the annotators on their personal computers.

The task of the annotators was to select which captions were a satisfyingly accurate description of the given image (even if the grammar is poor) and which captions were satisfyingly fluent (even if the description has nothing to do with the image). In order to simplify the annotation process and the aggregation of results, these annotations were binary (yes/no) rather than graded. This allows us to simply give the percentage of captions for a given architecture that were annotated as accurate or fluent. The annotators also thought that binary choices were easier and faster to input. Although there is a forced choice, this did not seem to be a problem in preliminary tests.

\begin{figure}
	\centering
	\includegraphics[scale=0.6]{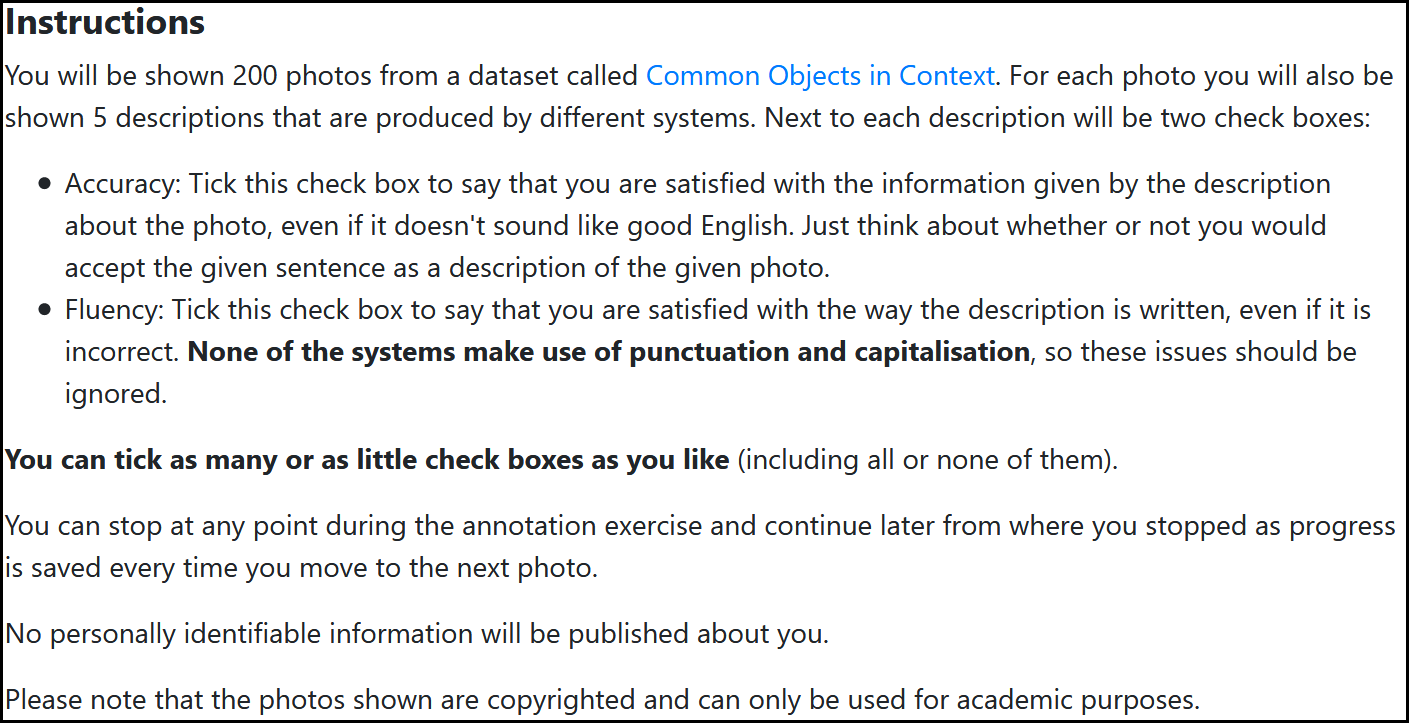}
	\caption{
		\label{fig:wi_humaneval_instructions}
		A screenshot of the instructions presented to the annotators prior to beginning the annotation process.
	}
\end{figure}

\begin{figure}
	\centering
	\includegraphics[scale=0.5]{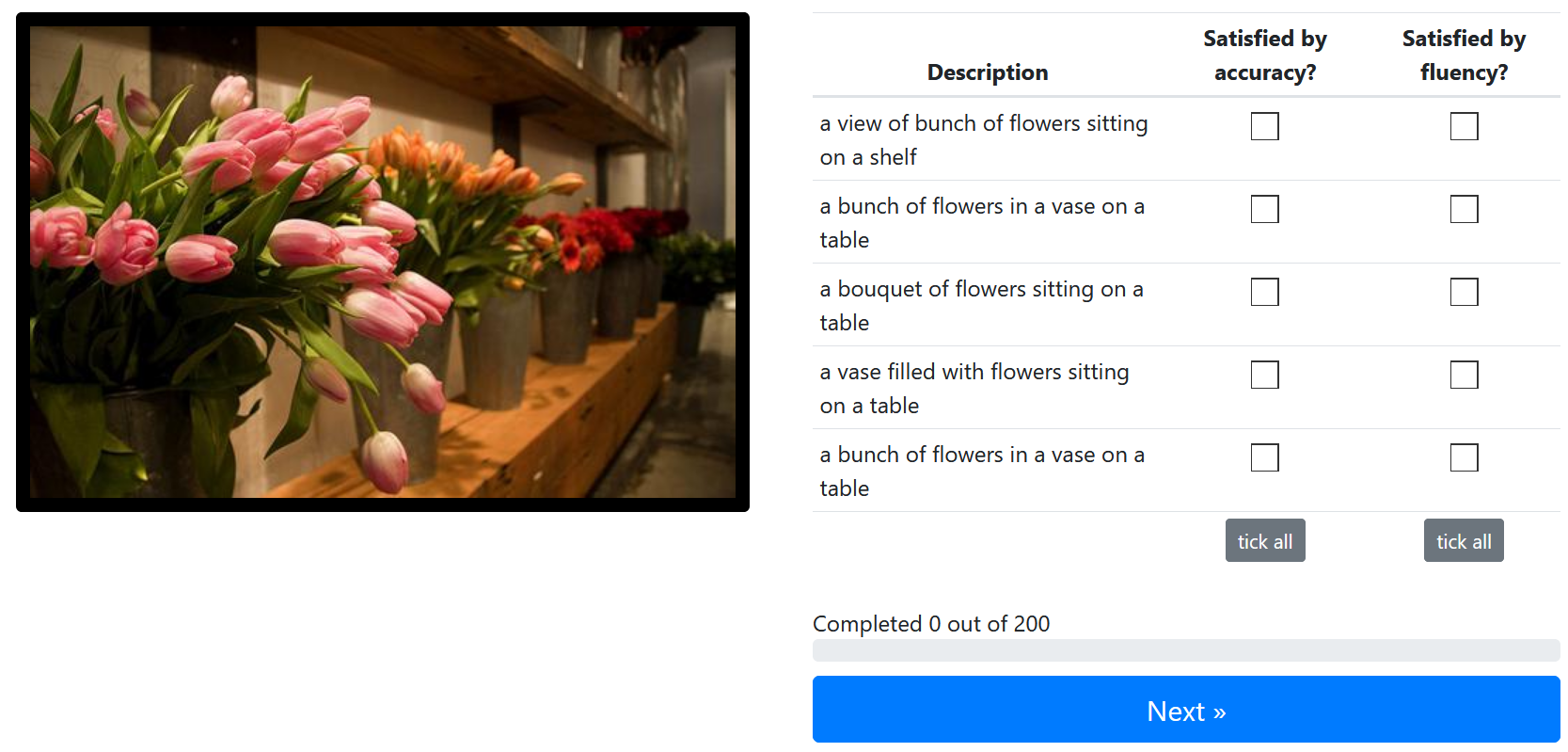}
	\caption{
		\label{fig:wi_humaneval_screenshot}
		A screenshot of one of the questions presented to the annotators during the annotation process.
	}
\end{figure}

The annotators were not informed of which caption was generated by which architecture and the order of the images shown was randomised for each annotator. For each image, the order of the captions shown was also randomised for each user. For reference, apart from the generated captions, one of the human written captions in the test set was also shown among the captions. The annotators were not aware of this and were told that all captions were generated by a computer. Since there are about 5 manually written captions for each image, one was randomly selected for each image. Therefore, the true number of captions in the database was $(4+1) \times 200 \times 1 = 1\,000$ captions. Figure~\ref{fig:wi_humaneval_instructions} shows a screenshot of the instructions presented to the annotators whilst Figure~\ref{fig:wi_humaneval_screenshot} shows a screenshot of the annotation screen with an image and corresponding captions.

All 5 annotators finished annotating all 200 images. The inter-annotator agreement, as measured using Cohen's kappa coefficient, is shown in Table~\ref{tbl:wi_humaneval_results_agreement}. The agreement is generally low, which is evidence that caption evaluation is not an exact science. It is interesting however that human written captions have less inter-annotator agreement than automatically generated ones. To further investigate this, we also measured the percentage of annotations that were unanimous, that is, where all 5 annotators gave the same annotation for a given caption, as shown in Table~\ref{tbl:wi_humaneval_results_unanimouslyagreed}. Here we see that human written captions have less agreement when it comes to fluency but more so on accuracy.

Table~\ref{tbl:wi_humaneval_results_positives} shows the percentage of positive annotations given to the captions of each architecture. We can see that human captions are much more accurate than automatically generated captions, but then are the least fluent. This is again confirmed in Table~\ref{tbl:wi_humaneval_results_unanimouslypositives} which shows the percentage of positive annotations after filtering for only unanimously agreed annotations. Clearly there is something unusual about the human written captions that was not picked up by the neural networks. The human written captions \citep{Chen2015a} (which are the ones provided in the MSCOCO test set) are written by Amazon Mechanical Turk workers who were asked to:
\begin{itemize}
	\item describe all the important parts of the scene,
	\item not to start sentences with `there is',
	\item not to describe unimportant details,
	\item not to describe things that might have happened in the future or past,
	\item not to describe what a person might be saying,
	\item not to give people proper names,
	\item and not to use less than 8 words.
\end{itemize}
Nothing in the rules predicts that the captions should not be fluent. Here are all of the unanimously agreed non-fluent human written descriptions:
\begin{itemize}
	\item a man that is jumping a skateboard outside
	\item a few people are getting of a plane
	\item a large body of water with small boats floating on top of it
	\item people on the street near a sea with waters
	\item a guy showing off a zebra at a building
	\item a cluttered computer desk has a nice chair with it
\end{itemize}
The second one has a spelling mistake whilst the rest sound awkward. On the other hand, typical automatically generated captions sound like:
\begin{itemize}
	\item a bike parked next to a wooden fence
	\item a group of people sitting on a bench
	\item a bird standing on top of a body of water
	\item a group of elephants standing next to each other
	\item a little girl sitting at a table with a cake
\end{itemize}
It is clear that the neural networks have learned a very simple grammatical pattern which is probably the most common in the training set. The advantage of the neural networks is that if they latch onto a fluent pattern then they can use it consistently whereas humans are not consistent in general.

\begin{table}
	\centering
	\begin{tabular}{l|ccccc}
		 &	Human &	Init-inject &	Pre-inject &	Par-inject &	Merge \\
		\hline
		Accuracy &	{0.262} &	{0.433} &	{0.439} &	{0.435} &	\textbf{0.457} \\
		Fluency &	{0.127} &	{0.209} &	{0.169} &	\textbf{0.305} &	{0.284} \\
	\end{tabular}
	\caption{
		\label{tbl:wi_humaneval_results_agreement}
		Results of the amount of inter-annotator agreement using Cohen's kappa coefficient.
	}
\end{table}

\begin{table}
	\centering
	\begin{tabular}{l|ccccc}
		 &	Human &	Init-inject &	Pre-inject &	Par-inject &	Merge \\
		\hline
		Accuracy &	\textbf{63.0\%} &	{49.0\%} &	{51.5\%} &	{55.0\%} &	{55.0\%} \\
		Fluency &	{20.5\%} &	{42.0\%} &	{44.5\%} &	{42.5\%} &	\textbf{45.0\%} \\
	\end{tabular}
	\caption{
		\label{tbl:wi_humaneval_results_unanimouslyagreed}
		Results of the percentage of responses that were in unanimous agreement.
	}
\end{table}

\begin{table}
	\centering
	\begin{tabular}{l|ccccc}
		 &	Human &	Init-inject &	Pre-inject &	Par-inject &	Merge \\
		\hline
		Accuracy &	\textbf{87.1\%} &	{33.7\%} &	{29.4\%} &	{27.4\%} &	{27.7\%} \\
		Fluency &	{65.7\%} &	{76.6\%} &	\textbf{80.8\%} &	{72.9\%} &	{76.2\%} \\
	\end{tabular}
	\caption{
		\label{tbl:wi_humaneval_results_positives}
		Results of the percentage of responses that were positive.
	}
\end{table}

\begin{table}
	\centering
	\begin{tabular}{l|ccccc}
		 &	Human &	Init-inject &	Pre-inject &	Par-inject &	Merge \\
		\hline
		Accuracy &	\textbf{98.4\%} &	{16.3\%} &	{13.6\%} &	{10.0\%} &	{12.7\%} \\
		Fluency &	{85.4\%} &	{94.0\%} &	\textbf{96.6\%} &	{88.2\%} &	{91.1\%} \\
	\end{tabular}
	\caption{
		\label{tbl:wi_humaneval_results_unanimouslypositives}
		Results of the percentage of responses that were positive and in unanimous agreement out of all unanimously agreed responses.
	}
\end{table}

Regarding the results of the automatically generated captions, according to the automatic metrics, namely geometric mean perplexity (Table~\ref{tbl:wi_results_exp_prb_mscoco}), WMD (Table~\ref{tbl:wi_results_exp_qty_mscoco}), and recall at 1 (Table~\ref{tbl:wi_results_exp_ret_mscoco}), which all gave the same ranking in the MSCOCO data, the performance ranking should be init-inject, followed by par-inject, followed by merge, followed by pre-inject. But according to the human evaluation the ranking by accuracy is init-inject, followed by pre-inject, followed by merge, followed by par-inject. Somehow, par-inject, which was ranked as one of the best by the automatic metrics, was deemed as performing the worst by the annotators whilst pre-inject, which was ranked as the worst by the automatic metrics, was deemed as performing very well by the annotators. Although there is agreement on the best architecture, perhaps more work needs to go into designing automatic evaluation measures as there might still be a long way to go before we should trust them as much as we do \citep{Vinyals2017,Reiter2009}. On the other hand, the accuracies are quite similar to each other and only 200 images were used so maybe a larger sample would give a different story.

Although fluent captions do not imply that they also sound pleasant, it is very telling that pre-inject was considered to be the most fluent, even though it has very stereotyped captions according to the diversity metrics. It seems that you can make very fluent sentences using only very frequent words. The second most fluent system was init-inject, which is promising given that it was the most accurate system.

We conclude this section with some examples of generated captions together with their corresponding image in Table~\ref{tbl:wi_humaneval_results_examples}. We show one example for every combination of accuracy and fluency for comparison.\\

\begin{table}
	\centering
	\begin{tabular}{ccp{4cm}c}
		Accurate? &	Fluent? &	Caption &	Image \\
		\hline
		Yes &	Yes &	{a bathroom with a toilet and a sink} &	\raisebox{-\totalheight}{\includegraphics[scale=0.6]{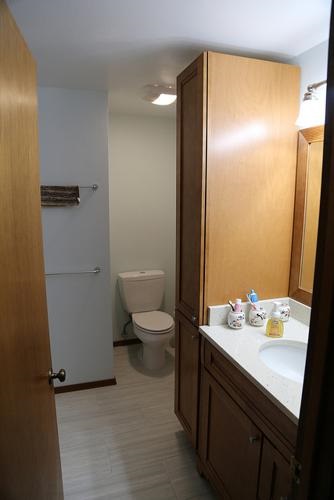}} \\
		Yes &	No &	{a white plate topped with a slice of cake} &	\raisebox{-\totalheight}{\includegraphics[scale=0.6]{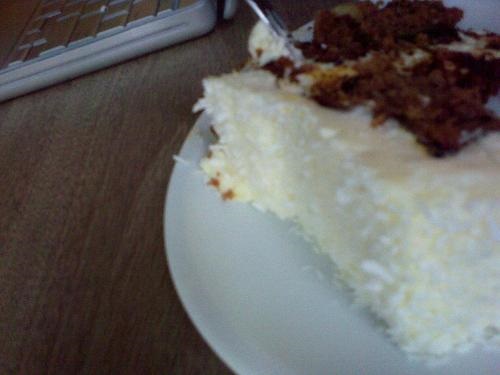}} \\
		No &	Yes &	{a man holding a banana in his hand} &	\raisebox{-\totalheight}{\includegraphics[scale=0.6]{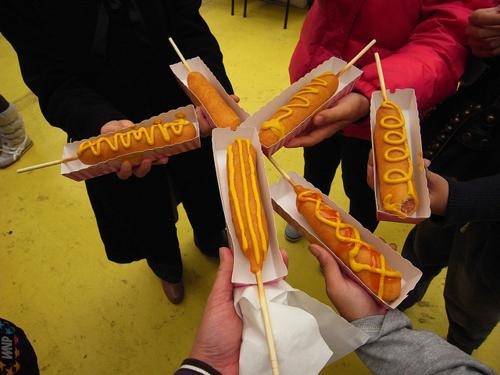}} \\
		No &	No &	{a wooden bench sitting on top of a wooden bench} &	\raisebox{-\totalheight}{\includegraphics[scale=0.6]{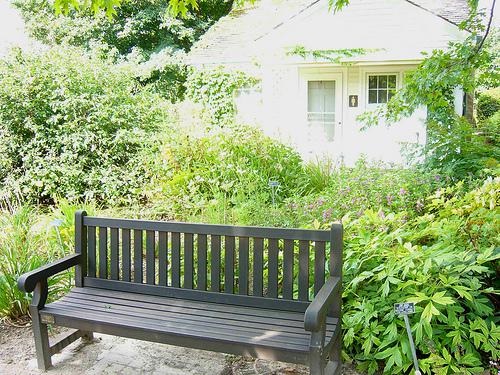}} \\
	\end{tabular}
	\caption{
		\label{tbl:wi_humaneval_results_examples}
		Examples of captions for each combination of unanimous accuracy and fluency annotations. All of these captions were generated by the merge architecture as it was the only one to have an example from each combination.
	}
\end{table}

\subsubsection{Interim summary}

\begin{itemize}
	\item Init-inject is the best architecture overall.
	
	\item Beyond the best architecture, the human evaluation and automatic evaluation are in disagreement. Assuming that the sample size is reliable, this means that the automatic metrics might not be as reliable as we expect them to be.
	
	\item Although using only very high frequency words, pre-inject generated the most fluent sentences of all.
\end{itemize}

\clearpage


\section{Conclusion}

In general, it seems that init-inject performs the best as an architecture, both according to automatic measures and human measures. One caveat here is that these models are highly dependent on the hyperparameters used, as a small change in the hyperparameters might result in substantial change in performance. For example, in a previous version of these experiments \citep{Tanti2018} that used a less comprehensive hyperparameter search, we found that par-inject performed poorly in general and that merge performed the best at the diversity metrics. Given that all the models were subjected to the same hyperparameter search process, we feel that the experiments performed here give a fair comparison nonetheless. Some consistencies that were found across experiments are that init-inject always performs well and merge always has a relatively small number of parameters.

For convenience, here are the architectures sorted by their general performance:
\begin{itemize}
	\item According to automatic metrics: init-inject, par-inject, merge, pre-inject.

	\item According to human annotators: init-inject, pre-inject, merge, par-inject.
\end{itemize}

Whilst state-of-the-art caption generators seem to give the impression that the image captioning task is solved, experiments show that these models might still not be as grounded to vision as we would like them to be. For example \citet{Hodosh2016} set up a binary classification task where an image has to be assigned to one of two captions: one being correct and one being incorrect. The classification is done by using a caption generator to measure the probability of the caption given the image and picking the most probable caption. When the distractor caption was completely different from the correct caption except for mentioning the correct scene, such as `a man at a podium' as opposed to `a woman at a podium', the classification accuracy of the model was almost equal to chance. Similarly, \citet{Shekhar2017} also found that caption generators were not suitably grounded as they could not identify an incorrect caption better than chance.

How to visually ground a language model is therefore still an open question. In order to better understand how to answer this question, it is important to first further analyse the different ways how each architecture grounds itself in vision by probing its internal representation. This is what we discuss in the next chapter.

\clearpage

%% file: tex/chp4_groundedness.tex
\chapterwithfootnote{Groundedness analysis}{An earlier version of the work shown in this chapter has been published \citep{Tanti2019b}.}
\clearpage

\section{Aims}

Whereas the previous chapter focussed on the practical aspects of different caption generator architectures, in this chapter we will investigate the internal representations learned by the different architectures. An understanding of internal representations, which is a task in the field of explainable AI \citep{Samek2018}, can shed light on the extent to which the generator is grounded in visual data and help to explain some of the model's output decisions, which is useful for understanding why a model makes any mistakes it does and for convincing users that it is working correctly.

It is known that not all words in a sentence are given equal importance by a neural language model \citep{Kadar2017}. Rather than measuring the importance of words, as was done by \citet{Kadar2017}, we would like to measure the importance of the image. The main question we address is how sensitive the different caption generators investigated in the previous chapter actually are to the visual input, that is, to what extent the output of these models varies as a function of the image. We address this using sensitivity analysis \citep{Samek2018} and an analysis based on foils \citep{Shekhar2017}. In addressing this question, we hope to achieve a better understanding of the extent to which different caption generation architectures succeed in grounding words in visual features.

\clearpage


\section{Experiments}

Our goal is to measure how much influence visual information has on the output of each implemented architecture described in the previous chapter. We want to see how this influence changes throughout the course of generating a caption. We will use the already-trained models of all architectures, datasets, and runs obtained from the previous chapter to do this.

\begin{figure}
	\centering
	\includegraphics[scale=0.6]{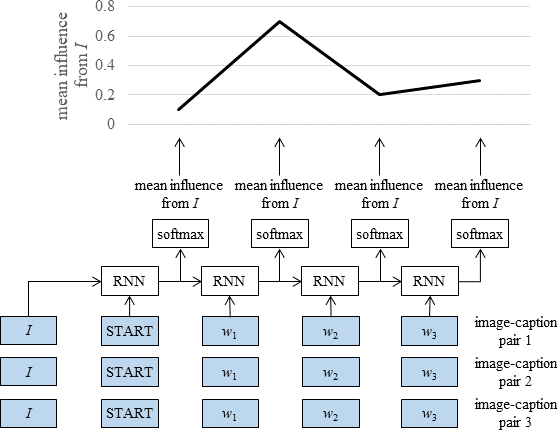}
	\caption{
		\label{fig:ii_visual_influence}
		A visual explanation of how captions from the test set are used to measure visual influence on the output. $I$ is an image, $w$ is a word in a caption, and the illustrated architecture is init-inject. Captions of a given length (three in the diagram) are grouped together and the amount of influence that the image has on the neural network is measured after every word. Note that the first point on the graph is the influence of the image on predicting the first word in the caption and not on the start token.
	}
\end{figure}

The way we use the captions to measure visual influence is illustrated in Figure~\ref{fig:ii_visual_influence}. We take captions from the test sets of Flickr8K, Flickr30K, and MSCOCO and feed each caption to the already-trained caption generators one word at a time to predict the probability of the next word. The words that are predicted by the neural network are not used at any point in this process. Only the words in the test set captions are fed to the neural network, just like when we were predicting the probability of captions. Each time a word is fed, the influence of the image is measured using one of two ways described below. We do this using all captions of a given length so that we can report the mean amount of visual influence at each time step over all corresponding time steps in captions. This allows us to draw a graph showing how average visual influence changes for each word in all captions of a given length. We also consider all five runs of the trained models and take their average visual influence per time step.

\subsection{Visual influence measures}

Although there are numerous existing techniques for measuring the visual influence of a caption generator \citep{Hodosh2016,Shekhar2017,Kadar2017}, these do not allow us to measure the influence of the image on every word but on only some of the words such as nouns. For this reason, we developed two visual influence measures: sensitivity analysis and omission scoring, both of which are explained in detail below.

\subsubsection{Sensitivity analysis}

Sensitivity analysis \citep{Samek2018} involves measuring the absolute gradient of a model's output with respect to its input in order to see how sensitive the output is to different parts of the input. The more sensitive the output is to a part of the input, the more important that part of the input is to produce the given output.

We use this technique to measure how sensitive the softmax layer is to the image at different time steps in the generation process. We do this by computing the absolute value of the partial derivative of the softmax output with respect to the input image vector. The steeper the gradient of the output with respect to a given image, the greater the change would be in the output if the image were different. It is important to note that even though the image might only be input once as an initial state to the RNN (as would happen in the init-inject architecture), the output's gradient with respect to the image will not be the same at every time step.

As we implemented our neural networks in Tensorflow, it is not possible to find the gradient of the whole softmax vector since, at the time of writing, Tensorflow does not allow for computing full Jacobian matrices efficiently. It can only efficiently find the gradient of a scalar (with respect to any shape tensor). Instead, we only take the gradient of a single element in the softmax: the maximum value. The maximum probability element in a softmax vector can be taken to be the output token of the neural network when generating the caption greedily. This gives us a vector with a partial derivative for every element in the image vector. We aggregate these partial derivatives by taking the mean of the absolute values.

\textit{Note that this visual influence measure measures the influence of the image on a single element in the softmax output only (for every time step in the captions).}

\subsubsection{Omission scoring}

Omission scoring \citep{Kadar2017} measures changes in the model's output as some part of an input is removed or replaced by a `foil'. The more the output changes as a result of removing a particular input, the more important the removed input is. This has been done in the image captioning domain, and has yielded datasets such as FOIL-COCO \citep{Shekhar2017}. \citet{Shekhar2017} tested the visual sensitivity of images by replacing words in captions with strategically chosen different words called foils which result in the caption not being faithful to the image content any more. A trained caption generator was then used to see if the model's output probabilities can be used to detect which captions contain the foil word (captions with a foil word should have a smaller probability given the correct image than completely correct captions). The results showed that this is a hard task for many vision-language models, despite being trivial for humans.

We use a similar technique to measure how important the image is to the hidden layer representations in the neural network by using a foil image. A foil image is selected for each correct image by taking the nouns in the correct image's captions and selecting all the other test set images that do not have any caption nouns in common with it. An image whose captions have no nouns in common is likely to have no salient objects in common and so should depict a different scene. Given this subset of images, we then compare each filtered image's feature vector to that of the correct image using cosine distance and use the image that is the most different in feature space from the correct image. This increases the likelihood that the foil images are maximally different from the correct image both in terms of content and feature space.

In order to measure the influence of the image on the caption generators, we measure the cosine distance which is defined as
\begin{align}
\text{cos}(u, v) = \frac{\sum_{i} (u_i \times v_i)}{\sqrt{\sum_{i} {u_i}^2} \times \sqrt{\sum_{i} {v_i}^2}}
\end{align}
where $u$ and $v$ are equally sized vectors. Cosine is used to measure the distance between the internal representation of the model when the correct image is used and the perturbed internal representation when the foil image is used. The greater the distance, the more influence the image has on the model's internal representation and hence the more important the image to the predicted word. For internal representations we use the multimodal vector (RNN hidden state vector for inject architectures or the RNN hidden state vector concatenated with the image vector for the merge architecture) and the softmax output vector. We measure by how much these representations change when perturbed with the foil image for every word in the image's captions.

\textit{Note that this visual influence measure measures the influence of the image on a whole layer rather than on a single element in a layer like the previous measure does.}

\subsection{Caption lengths}

\begin{figure}
	\centering
	\begin{subfigure}{0.3\textwidth}
		\centering
		\includegraphics[scale=0.5]{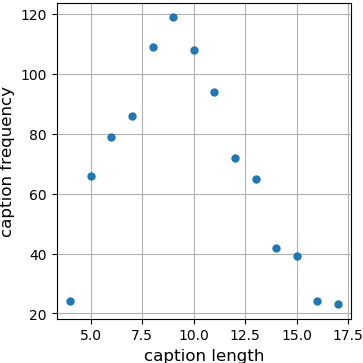}
		\caption{
			\label{fig:ii_sentlen-sentfreqs_flickr8k}
			Dataset: Flickr8K.
		}
	\end{subfigure}
	\quad
	\begin{subfigure}{0.3\textwidth}
		\centering
		\includegraphics[scale=0.5]{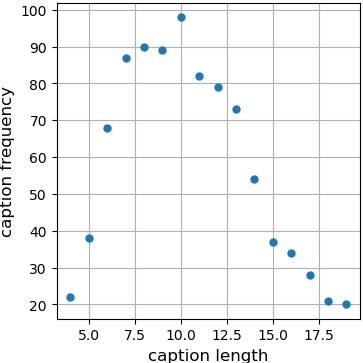}
		\caption{
			\label{fig:ii_sentlen-sentfreqs_flickr30k}
			Dataset: Flickr30K.
		}
	\end{subfigure}
	\quad
	\begin{subfigure}{0.3\textwidth}
		\centering
		\includegraphics[scale=0.5]{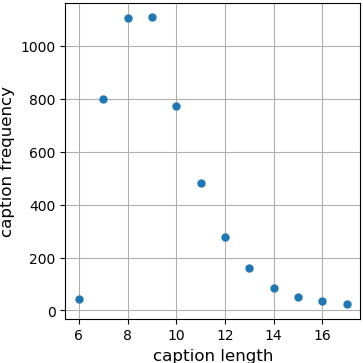}
		\caption{
			\label{fig:ii_sentlen-sentfreqs_mscoco}
			Dataset: MSCOCO.
		}
	\end{subfigure}
	\caption{
		\label{fig:ii_caplen-capfreqs}
		A length-frequency plot of the captions in the test sets of each dataset.
	}
\end{figure}

Since captions are grouped by caption length for these experiments to work (see Figure~\ref{fig:ii_visual_influence}), we opted to work with a short caption length and a long one in order to see the effect of both short and long sentences. It is important that the number of captions with these given lengths is sufficient to create a meaningful average when aggregating results. Therefore, different caption lengths were selected for each dataset according to the frequencies of each caption length in each dataset.

The frequencies of each dataset's caption lengths form a unimodal frequency curve as shown in Figure~\ref{fig:ii_caplen-capfreqs}. We opted to choose the shortest and longest length for which there are at least 50 different captions in the respective test set. The caption lengths we chose were:

\begin{itemize}
	\item Flickr8K: 5 (frequency: 66) and 13 (frequency: 65)
	\item Flickr30K: 6 (frequency: 68) and 14 (frequency: 54)
	\item MSCOCO: 7 (frequency: 800) and 15 (frequency: 51)
\end{itemize}

\clearpage


\section{Results}

\subsection{Nouns}

We start with this subsection in order to make it easier to see a common trend in all the following experiment results. As will be shown, nouns generally turn out to be an important feature to predicting visual influence. It will be shown that when the next word to be predicted is a noun, the image's influence on the output tends to dramatically increase. This makes sense since nouns tend to be visual in nature and the captions were written with the intention of being concrete and conceptual rather than abstract, which biases the humans writing the captions toward naming salient objects. It also makes sense given that the convolutional neural network used to extract image features was trained for the task of object recognition rather than something like action recognition which would be useful for verbs. Therefore, the neural network is likely basing its decision to produce a verb more on linguistic information than visual information. \citet{Wang2018} provides more evidence for this by showing that a bag-of-words vector specifying which objects are in the image is enough information to condition the neural language model in a way that provides even better captions than when using visual features extracted from a CNN.

\begin{figure}
	\centering
	\figuretitle{Noun frequencies}
	\begin{subfigure}{0.4\textwidth}
		\centering
		\includegraphics[scale=0.5]{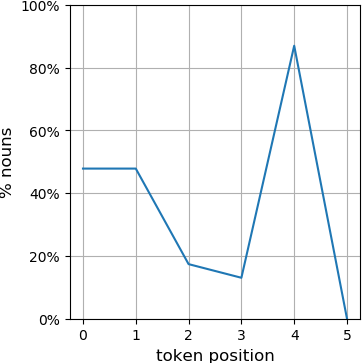}
		\caption{
			\label{fig:ii_tagfreqs_flickr8k_5}
			Dataset: Flickr8K, length: 5.
			\vspace{15pt}
		}
	\end{subfigure}
	\quad
	\begin{subfigure}{0.4\textwidth}
		\centering
		\includegraphics[scale=0.5]{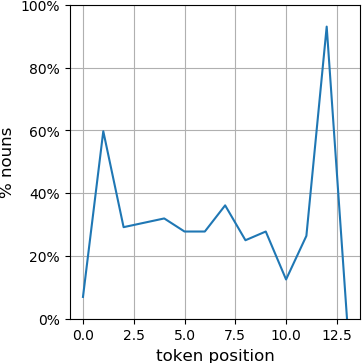}
		\caption{
			\label{fig:ii_tagfreqs_flickr8k_13}
			Dataset: Flickr8K, length: 13.
			\vspace{15pt}
		}
	\end{subfigure}
	
	\begin{subfigure}{0.4\textwidth}
		\centering
		\includegraphics[scale=0.5]{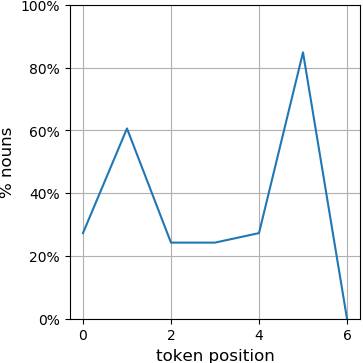}
		\caption{
			\label{fig:ii_tagfreqs_flickr30k_6}
			Dataset: Flickr30K, length: 6.
			\vspace{15pt}
		}
	\end{subfigure}
	\quad
	\begin{subfigure}{0.4\textwidth}
		\centering
		\includegraphics[scale=0.5]{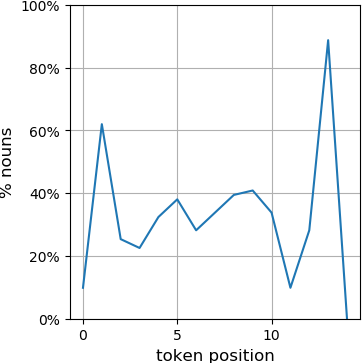}
		\caption{
			\label{fig:ii_tagfreqs_flickr30k_14}
			Dataset: Flickr30K, length: 14.
			\vspace{15pt}
		}
	\end{subfigure}
	
	\begin{subfigure}{0.4\textwidth}
		\centering
		\includegraphics[scale=0.5]{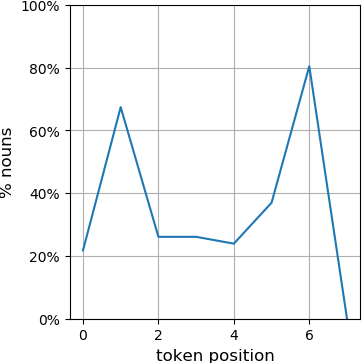}
		\caption{
			\label{fig:ii_tagfreqs_mscoco_7}
			Dataset: MSCOCO, length: 7.
		}
	\end{subfigure}
	\quad
	\begin{subfigure}{0.4\textwidth}
		\centering
		\includegraphics[scale=0.5]{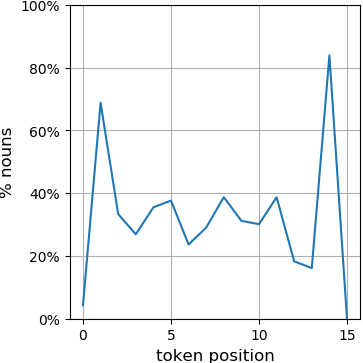}
		\caption{
			\label{fig:ii_tagfreqs_mscococ_15}
			Dataset: MSCOCO, length: 15.
		}
	\end{subfigure}
	\caption{
		\label{fig:ii_tagfreqs}
		The percentage of sentences of the given length that have a noun at the given token position. Note that the last token index is for the end token.
	}
\end{figure}

Therefore, before discussing the results, it is worth noting where nouns tend to occur in the test set captions of each dataset. For this reason, we have shown in Figure~\ref{fig:ii_tagfreqs} what percentage of words are nouns in different token positions in captions. The captions were tagged using NLTK's default tagger\footnote{\texttt{nltk.pos\_tag()}} (v3.2.5) using the universal tagset.

It is clear that nouns tend to occur mainly in two positions of a caption: the second word and the last word, with the last word being the most likely to be a noun out of all token positions. The captions in the datasets have a degree of stereotypical constructions that relate two nouns on either side of the sentence, both of which are preceded by determiners. Here are some examples from MSCOCO:
\begin{itemize}
	\item a dog running in a field
	\item a statue sitting by the road near a door
	\item a man is walking down a path covered in a snow
\end{itemize}
It turns out that the second and last word positions usually require the most visual information to process, as we will show below.

The reason why short Flickr8K captions seem to have the same percentage of words being nouns in the first and second positions is because, in order to be short, they sometimes do away with the first determiner. An example of a short Flickr8K caption is ``dog running in a field''.

\subsection{Sensitivity analysis}

\begin{figure}
	\centering
	\figuretitle{Sensitivity with respect to image}
	\begin{subfigure}{0.4\textwidth}
		\centering
		\includegraphics[scale=0.5]{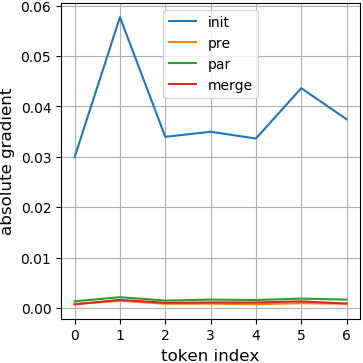}
		\caption{
			\label{fig:ii_results_gradient-wrt-image-max_flickr8k_5}
			Dataset: Flickr8K, length: 5.
			\vspace{15pt}
		}
	\end{subfigure}
	\quad
	\begin{subfigure}{0.4\textwidth}
		\centering
		\includegraphics[scale=0.5]{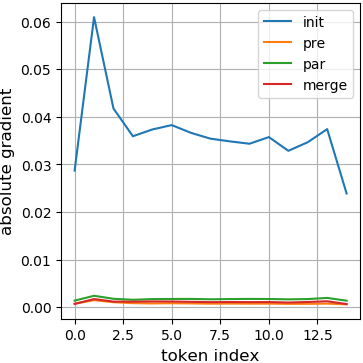}
		\caption{
			\label{fig:ii_results_gradient-wrt-image-max_flickr8k_13}
			Dataset: Flickr8K, length: 13.
			\vspace{15pt}
		}
	\end{subfigure}
	
	\begin{subfigure}{0.4\textwidth}
		\centering
		\includegraphics[scale=0.5]{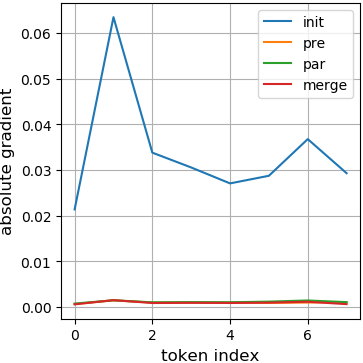}
		\caption{
			\label{fig:ii_results_gradient-wrt-image-max_flickr30k_6}
			Dataset: Flickr30K, length: 6.
			\vspace{15pt}
		}
	\end{subfigure}
	\quad
	\begin{subfigure}{0.4\textwidth}
		\centering
		\includegraphics[scale=0.5]{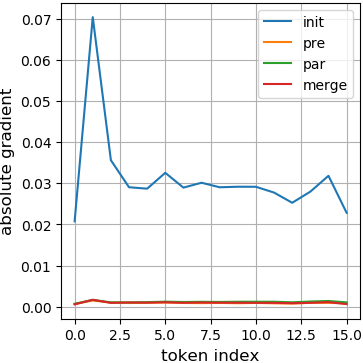}
		\caption{
			\label{fig:ii_results_gradient-wrt-image-max_flickr30k_14}
			Dataset: Flickr30K, length: 14.
			\vspace{15pt}
		}
	\end{subfigure}
	
	\begin{subfigure}{0.4\textwidth}
		\centering
		\includegraphics[scale=0.5]{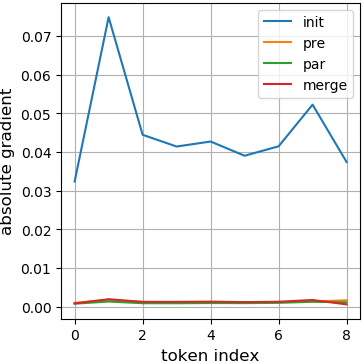}
		\caption{
			\label{fig:ii_results_gradient-wrt-image-max_mscoco_7}
			Dataset: MSCOCO, length: 7.
		}
	\end{subfigure}
	\quad
	\begin{subfigure}{0.4\textwidth}
		\centering
		\includegraphics[scale=0.5]{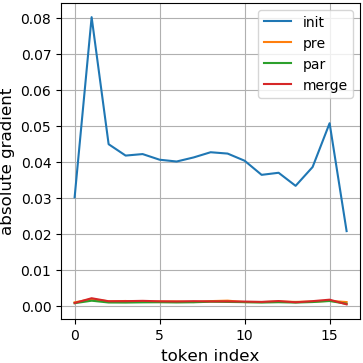}
		\caption{
			\label{fig:ii_results_gradient-wrt-image-max_mscococ_15}
			Dataset: MSCOCO, length: 15.
		}
	\end{subfigure}
	\caption{
		\label{fig:ii_results_gradient-wrt-image-max}
		Results for gradient of maximum probability at each time step with respect to the image. Note that the last token index is for the end token.
	}
\end{figure}

We start by measuring how sensitive the most probable next word in the softmax is to the average image vector element, shown in Figure~\ref{fig:ii_results_gradient-wrt-image-max}.

One feature is prevalent in the charts: init-inject is much more sensitive to the image than all the other architectures. But this is likely to be because init-inject is the only architecture to have the hyperparameter for normalising the image vector being set to true, which means that the image vector numbers will be very small next to unnormalised image vectors. This will make the weights in the layer processing the image vector large in order to compensate for this difference in magnitude, which in turn makes its derivative large.

\begin{figure}
	\centering
	\figuretitle{Sensitivity with respect to post-image}
	\begin{subfigure}{0.4\textwidth}
		\centering
		\includegraphics[scale=0.5]{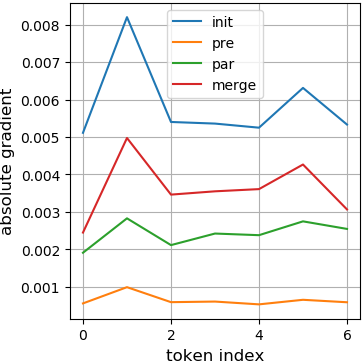}
		\caption{
			\label{fig:ii_results_gradient-wrt-postimage-max_flickr8k_5}
			Dataset: Flickr8K, length: 5.
			\vspace{15pt}
		}
	\end{subfigure}
	\quad
	\begin{subfigure}{0.4\textwidth}
		\centering
		\includegraphics[scale=0.5]{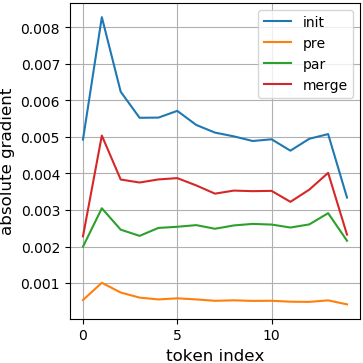}
		\caption{
			\label{fig:ii_results_gradient-wrt-postimage-max_flickr8k_13}
			Dataset: Flickr8K, length: 13.
			\vspace{15pt}
		}
	\end{subfigure}
	
	\begin{subfigure}{0.4\textwidth}
		\centering
		\includegraphics[scale=0.5]{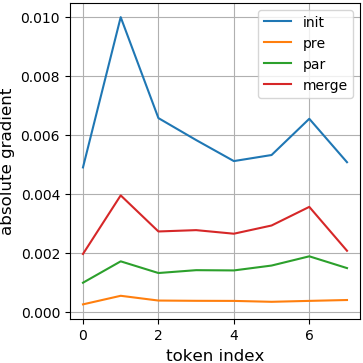}
		\caption{
			\label{fig:ii_results_gradient-wrt-postimage-max_flickr30k_6}
			Dataset: Flickr30K, length: 6.
			\vspace{15pt}
		}
	\end{subfigure}
	\quad
	\begin{subfigure}{0.4\textwidth}
		\centering
		\includegraphics[scale=0.5]{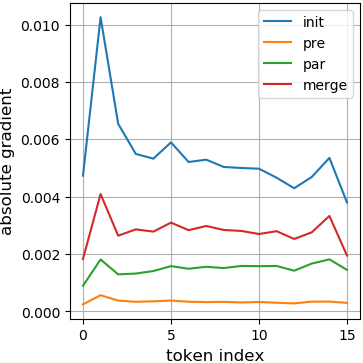}
		\caption{
			\label{fig:ii_results_gradient-wrt-postimage-max_flickr30k_14}
			Dataset: Flickr30K, length: 14.
			\vspace{15pt}
		}
	\end{subfigure}
	
	\begin{subfigure}{0.4\textwidth}
		\centering
		\includegraphics[scale=0.5]{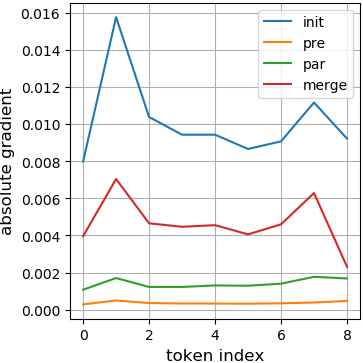}
		\caption{
			\label{fig:ii_results_gradient-wrt-postimage-max_mscoco_7}
			Dataset: MSCOCO, length: 7.
		}
	\end{subfigure}
	\quad
	\begin{subfigure}{0.4\textwidth}
		\centering
		\includegraphics[scale=0.5]{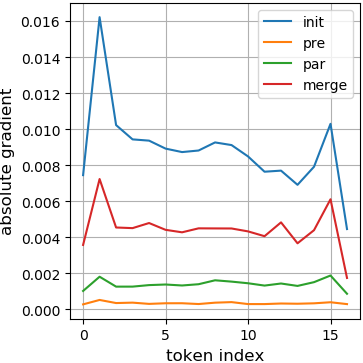}
		\caption{
			\label{fig:ii_results_gradient-wrt-postimage-max_mscococ_15}
			Dataset: MSCOCO, length: 15.
		}
	\end{subfigure}
	\caption{
		\label{fig:ii_results_gradient-wrt-postimage-max}
		Results for gradient of maximum probability at each time step with respect to the post-image. Note that the last token index is for the end token.
	}
\end{figure}

We can neutralise the effect of the weights being multiplied by the image vector by instead measuring the sensitivity with respect to the post-image vector, that is, the vector that comes out after the weights multiplication. This is shown in Figure~\ref{fig:ii_results_gradient-wrt-postimage-max}.

The sensitivity of init-inject is now less extreme when compared to the other architectures but still the highest. We can clearly see two spikes for every architecture in every chart: one for the second word and one for the last word (not including the end token). As mentioned in the previous subsection, this is likely due to these word positions being usually occupied by nouns.

The fact that init-inject is both the most sensitive architecture and also performed the best in the previous chapter does not mean that one predicts the other. Image sensitivity and generated caption quality are only partially related at best, plus the other architectures' sensitivity to the image does not predict their performance.

There also seems to be a downward trend in sensitivity to the image as the word being predicted gets closer to the end of sentence. For example, although the last word is more likely to be a noun than the second word, the second word's sensitivity to the image is usually higher than that of the last word. This is likely due to the caption generator relying more on information from the caption prefix as it gets longer rather than on the visual information. The longer the prefix, the more information is available in the prefix to deduce what the next word should be and thus the image becomes less important. Although this explains the general shape of the curves, it does not explain why init-inject is significantly affected more than merge. This is something we'll investigate further with omission scores in the next subsection.

\FloatBarrier

One reason why init-inject is so sensitive to the image has to do with the equation describing the GRU. As already shown in Equation~\ref{eqn:gru} in Subsection~\ref{sec:recurrent_neural_networks}, the equation describing how the next state is produced from the previous state is:
\begin{equation*}
\matrixsym{S}_t = \matrixsym{G^i}_t \odot \tanh(((\matrixsym{S}_{t-1} \odot \matrixsym{G^r}_t) \concat \matrixsym{X}_{t}) \tensorprod \matrixsym{W^s} + \vectorsym{b^s}) + \matrixsym{G^f}_t \odot \matrixsym{S}_{t-1}
\end{equation*}

\noindent If we abstract away details about parameters and gates then we get the following equation:
\begin{align}
\matrixsym{S}_t &= \tanh(\matrixsym{S}_{t-1} \concat \matrixsym{X}_{t}) + \matrixsym{S}_{t-1}
\end{align}

\noindent where $\matrixsym{S}$ is a matrix of state vectors, $\matrixsym{X}$ is a matrix of input vectors, and $\concat$ is the vector concatenation operator. We can now see how the recurrence relation evolves with every time step in terms of only inputs and the initial state:
\begin{align}
\matrixsym{S}_1 &= \tanh(\matrixsym{S}_{0} \concat \matrixsym{X}_{1}) + \matrixsym{S}_{0} \\
\matrixsym{S}_2 &= \tanh((\tanh(\matrixsym{S}_{0} \concat \matrixsym{X}_{1}) + \matrixsym{S}_{0}) \concat \matrixsym{X}_{2}) + \tanh(\matrixsym{S}_{0} \concat \matrixsym{X}_{1}) + \matrixsym{S}_{0} \\
\dots \nonumber
\end{align}

We can see that as the number of time steps increases, one thing that remains constant is that the initial state is simply added on as is to the equation. It is not bounded by a non-linear function like all the inputs are. Therefore the initial state is free to grow arbitrarily in absolute value before being added in, which means that it can easily dominate the value of the next state. In contrast, the LSTM would only behave similarly to the GRU when it is the cell state that is being used as both a conditioned initial state and a final state. The initial hidden state in the equation is always buried under at least one squashing function, which limits its absolute value, whilst the initial cell state is unbounded. This hints at injecting the image into the cell state of the LSTM rather than the hidden state. We leave confirmation of this hypothesis for future work.\\

\begin{figure}
	\centering
	\figuretitle{Sensitivity with respect to previous token}
	\begin{subfigure}{0.4\textwidth}
		\centering
		\includegraphics[scale=0.5]{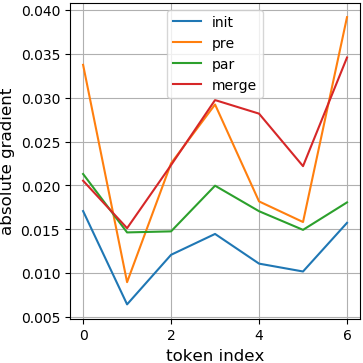}
		\caption{
			\label{fig:ii_results_gradient-wrt-prevtoken-max_flickr8k_5}
			Dataset: Flickr8K, length: 5.
			\vspace{15pt}
		}
	\end{subfigure}
	\quad
	\begin{subfigure}{0.4\textwidth}
		\centering
		\includegraphics[scale=0.5]{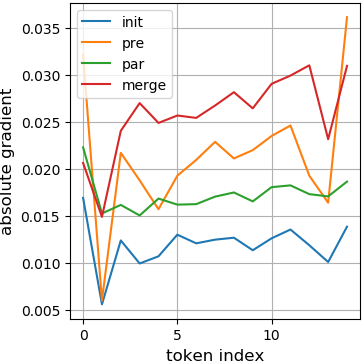}
		\caption{
			\label{fig:ii_results_gradient-wrt-prevtoken-max_flickr8k_13}
			Dataset: Flickr8K, length: 13.
			\vspace{15pt}
		}
	\end{subfigure}
	
	\begin{subfigure}{0.4\textwidth}
		\centering
		\includegraphics[scale=0.5]{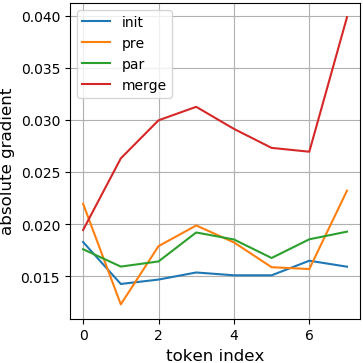}
		\caption{
			\label{fig:ii_results_gradient-wrt-prevtoken-max_flickr30k_6}
			Dataset: Flickr30K, length: 6.
			\vspace{15pt}
		}
	\end{subfigure}
	\quad
	\begin{subfigure}{0.4\textwidth}
		\centering
		\includegraphics[scale=0.5]{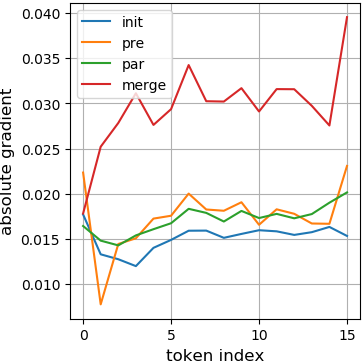}
		\caption{
			\label{fig:ii_results_gradient-wrt-prevtoken-max_flickr30k_14}
			Dataset: Flickr30K, length: 14.
			\vspace{15pt}
		}
	\end{subfigure}
	
	\begin{subfigure}{0.4\textwidth}
		\centering
		\includegraphics[scale=0.5]{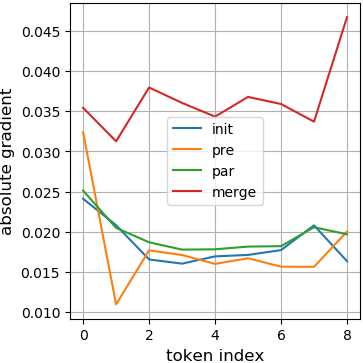}
		\caption{
			\label{fig:ii_results_gradient-wrt-prevtoken-max_mscoco_7}
			Dataset: MSCOCO, length: 7.
		}
	\end{subfigure}
	\quad
	\begin{subfigure}{0.4\textwidth}
		\centering
		\includegraphics[scale=0.5]{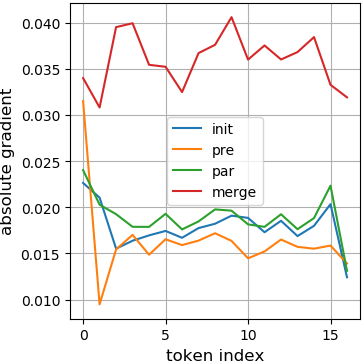}
		\caption{
			\label{fig:ii_results_gradient-wrt-prevtoken-max_mscococ_15}
			Dataset: MSCOCO, length: 15.
		}
	\end{subfigure}
	\caption{
		\label{fig:ii_results_gradient-wrt-prevtoken-max}
		Results for gradient of maximum probability at each time step with respect to the previous token. Note that the last token index is for the end token.
	}
\end{figure}

For completeness' sake, we shall also investigate how sensitive the most probable word is to the preceding word. This is to compare how much influence visual information has on the model as opposed to how much influence linguistic information has on the model. The model's sensitivity to the previous word is shown in Figure~\ref{fig:ii_results_gradient-wrt-prevtoken-max}.

\begin{figure}
	\centering
	\figuretitle{Sensitivity with respect to previous token (language model)}
	\begin{subfigure}{0.42\textwidth}
		\centering
		\includegraphics[scale=0.5]{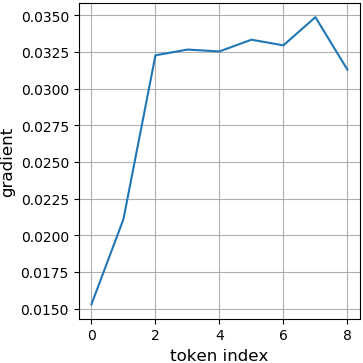}
		\caption{
			\label{fig:ii_langmod_gradient-wrt-prevtoken-max_mscoco_7}
			Dataset: MSCOCO, length: 7.
		}
	\end{subfigure}
	\quad
	\begin{subfigure}{0.42\textwidth}
		\centering
		\includegraphics[scale=0.5]{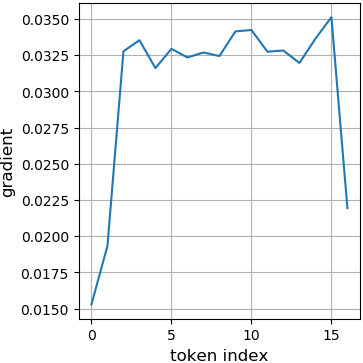}
		\caption{
			\label{fig:ii_langmod_gradient-wrt-prevtoken-max_mscococ_15}
			Dataset: MSCOCO, length: 15.
		}
	\end{subfigure}
	\caption{
		\label{fig:ii_langmod_gradient-wrt-prevtoken-max}
		Results for gradient of maximum probability at each time step with respect to the previous token in a language model (\ie text only). The language model is described in detail in the next chapter (corpus: MSCOCO, corpus size: 300\,000 sentences). Note that the last token index is for the end token.
	}
\end{figure}

Here we can see that, except in Flickr8K, merge is much more sensitive to the previous word in the caption (the one before the word position being predicted) than all the other architectures. This sensitivity probably stems from how similar the merge architecture is to a text-only language model. Given that the RNN hidden state vector is not being hampered by visual information in addition to the words (which happens in inject architectures), there is a cleaner connection from the previous word to the softmax. To give evidence of this, we compare the sensitivity of the merge architecture to the sensitivity of a `blind' language model in order to show that the high sensitivity comes from the lack of visual interference. We train a language model on the text of MSCOCO and measure how sensitive the maximum probability of its softmax is to the previous word. The language model itself will be described in detail in the next chapter. The results are shown in Figure~\ref{fig:ii_langmod_gradient-wrt-prevtoken-max}.

We can see how similar the gradients of the language model and the gradients of the merge architecture are, at least in scale if not in shape.

\FloatBarrier

\subsubsection{Interim summary}

\begin{itemize}
	\item The maximum probability value in the softmax of init-inject is very sensitive to the visual information, with merge coming in second. The reason for init-inject's sensitivity could be due to the way the GRU works. In LSTMs, the same effect can be obtained if the image is injected into the cell state rather than the hidden state.
	
	\item Sensitivity tends to go down as the prefix gets longer. This can be due to the neural network not needing to look at the image as the prefix gets longer as the prefix would be enough information.
	
	\item The previous point is not sufficient to explain why init-inject loses sensitivity more drastically than merge. This is investigated in the next section.
	
	\item The maximum probability value in the softmax of merge is very sensitive to the previous word in the prefix. This sensitivity is similar to the sensitivity of a text-only language model on the previous word.
\end{itemize}

\FloatBarrier

\subsection{Omission scores}

One reason why the sensitivity to the image goes down in init-inject is due to it having a finite memory. The RNN's fixed-size hidden state vector needs to accommodate both the visual information and information about more and more words as the prefix gets longer. Our hypothesis is that since the RNN's memory size is fixed, then it will be harder to remember both the prefix and the image as the prefix gets longer in an inject architecture. A merge architecture on the other hand has an easier time managing its memory as it is only being used to remember words rather than the image as well.

To test this hypothesis we use a second image importance measure that measures the visual influence on a whole layer rather than on a single element in a layer. We will be measuring how much the multimodal vector changes in terms of cosine distance if the image were replaced with a different foil image while keeping the same caption. The multimodal vector is the vector of activations that comes from the layer which contains both information about the image and about the caption prefix. In the case of inject architectures it is the RNN state itself whilst in the case of the merge architecture it is the RNN state concatenated to the post-image vector.

\begin{figure}
	\centering
	\includegraphics[scale=0.5]{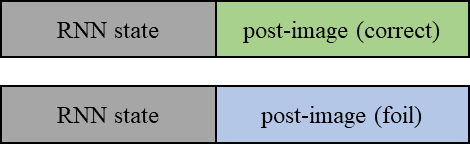}
	\caption{
		\label{fig:ii_merge_cos_vectors}
		A simple illustration showing the original and foil multimodal vector of the merge architecture. Rectangle widths are to scale (227 units vs. 268 units). The top two grey and green rectangles stand for the concatenated multimodal vector resulting from the correct image and the bottom two grey and blue rectangles stand for the multimodal vector resulting from the foil image. Note how the grey area is identical for both vectors, resulting in a cosine distance that should be smaller than if the two vectors had nothing in common.
	}
\end{figure}

Note that in the case of merge, we are measuring the cosine distance between the RNN's vector being concatenated with the correct image vector and the same RNN vector being concatenated with the foil image vector. This means that the merge architecture's multimodal vectors have about half of the elements in their vectors being identical (about 46\% to be more precise). This is shown in Figure~\ref{fig:ii_merge_cos_vectors}. An analysis on randomly generated vectors shows that if the RNN hidden state vector and post-image vector consisted of standard normal random numbers, then the cosine distance between the correct and foil multimodal vectors would be around 0.54 on average. This is in contrast to the cosine distance of the two completely random vectors with nothing in common which would be around 1.0 on average. This means that the merge architecture is expected to be at a disadvantage as it is harder for it to produce a very different multimodal vector than the inject architectures.

\begin{figure}
	\centering
	\figuretitle{Omission with respect to multimodal vector}
	\begin{subfigure}{0.4\textwidth}
		\centering
		\includegraphics[scale=0.5]{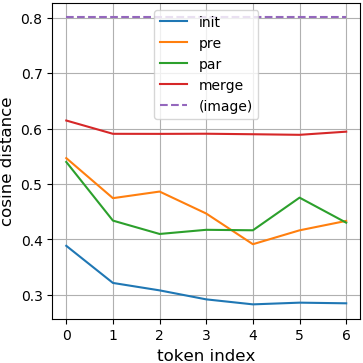}
		\caption{
			\label{fig:ii_results_omission-multimodalvec-cos_flickr8k_5}
			Dataset: Flickr8K, length: 5.
			\vspace{15pt}
		}
	\end{subfigure}
	\quad
	\begin{subfigure}{0.4\textwidth}
		\centering
		\includegraphics[scale=0.5]{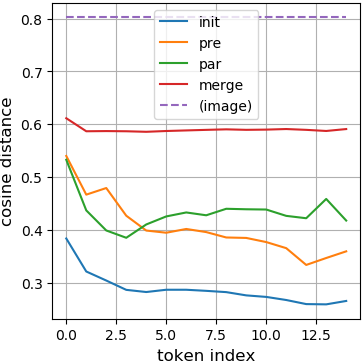}
		\caption{
			\label{fig:ii_results_omission-multimodalvec-cos_flickr8k_13}
			Dataset: Flickr8K, length: 13.
			\vspace{15pt}
		}
	\end{subfigure}
	
	\begin{subfigure}{0.4\textwidth}
		\centering
		\includegraphics[scale=0.5]{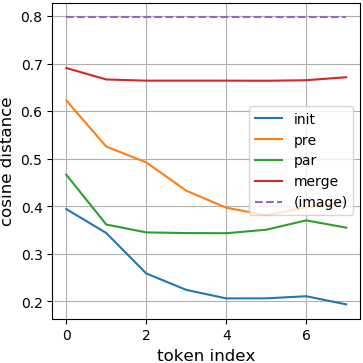}
		\caption{
			\label{fig:ii_results_omission-multimodalvec-cos_flickr30k_6}
			Dataset: Flickr30K, length: 6.
			\vspace{15pt}
		}
	\end{subfigure}
	\quad
	\begin{subfigure}{0.4\textwidth}
		\centering
		\includegraphics[scale=0.5]{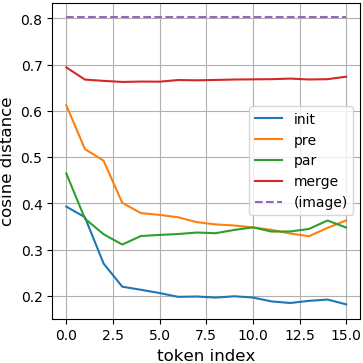}
		\caption{
			\label{fig:ii_results_omission-multimodalvec-cos_flickr30k_14}
			Dataset: Flickr30K, length: 14.
			\vspace{15pt}
		}
	\end{subfigure}
	
	\begin{subfigure}{0.4\textwidth}
		\centering
		\includegraphics[scale=0.5]{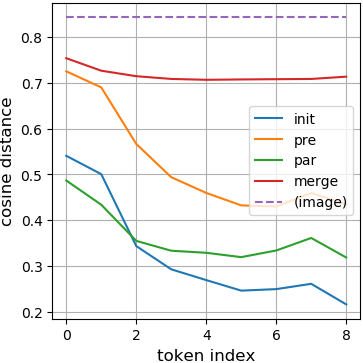}
		\caption{
			\label{fig:ii_results_omission-multimodalvec-cos_mscoco_7}
			Dataset: MSCOCO, length: 7.
		}
	\end{subfigure}
	\quad
	\begin{subfigure}{0.4\textwidth}
		\centering
		\includegraphics[scale=0.5]{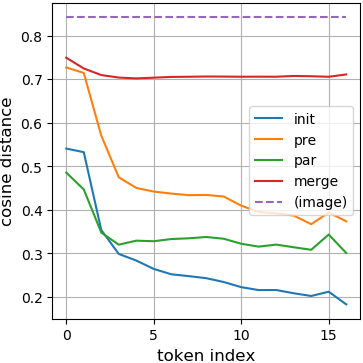}
		\caption{
			\label{fig:ii_results_omission-multimodalvec-cos_mscococ_15}
			Dataset: MSCOCO, length: 15.
		}
	\end{subfigure}
	\caption{
		\label{fig:ii_results_omission-multimodalvec-cos}
		Results for cosine distance between multimodal vectors resulting from the correct image and foil image. Dashed line is the average cosine distance between the correct image and foil image. Note that the last token index is for the end token.
	}
\end{figure}

The results of the omission scoring measure of visual influence are shown in Figure~\ref{fig:ii_results_omission-multimodalvec-cos}. In addition to the multimodal vector, we also show what the average cosine distance between the original images and their corresponding foil images is.
 
One observation that stands out is how the merge architecture's omission scores remain relatively stable and large across datasets. The fact that the merge architecture is expected to have a smaller cosine distance but instead has a relatively large one shows how much the architecture's internal representation is grounded in visual information. The fact that the visual influence remains stable whilst the other architectures seem to lose visual information quickly as we move towards the end of the sentence is also a good sign for the merge architecture.

\FloatBarrier

\begin{figure}
	\centering
	\figuretitle{Omission with respect to output}
	\begin{subfigure}{0.4\textwidth}
		\centering
		\includegraphics[scale=0.5]{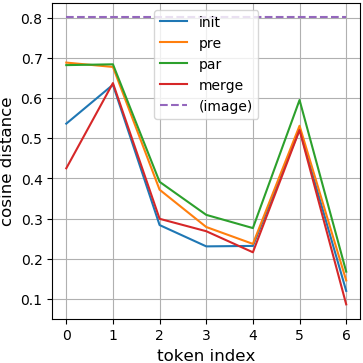}
		\caption{
			\label{fig:ii_results_omission-output-cos_flickr8k_5}
			Dataset: Flickr8K, length: 5.
			\vspace{15pt}
		}
	\end{subfigure}
	\quad
	\begin{subfigure}{0.4\textwidth}
		\centering
		\includegraphics[scale=0.5]{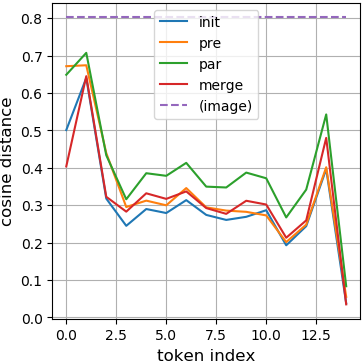}
		\caption{
			\label{fig:ii_results_omission-output-cos_flickr8k_13}
			Dataset: Flickr8K, length: 13.
			\vspace{15pt}
		}
	\end{subfigure}
	
	\begin{subfigure}{0.4\textwidth}
		\centering
		\includegraphics[scale=0.5]{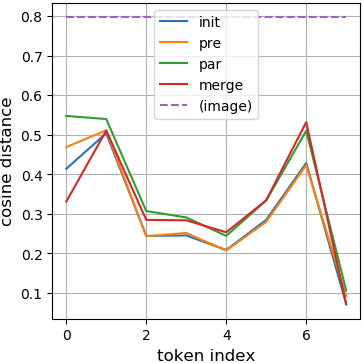}
		\caption{
			\label{fig:ii_results_omission-output-cos_flickr30k_6}
			Dataset: Flickr30K, length: 6.
			\vspace{15pt}
		}
	\end{subfigure}
	\quad
	\begin{subfigure}{0.4\textwidth}
		\centering
		\includegraphics[scale=0.5]{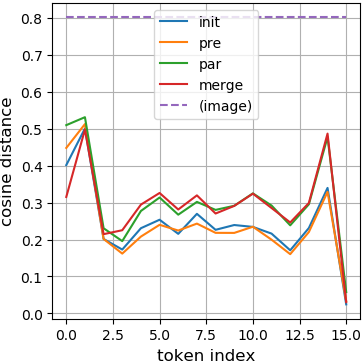}
		\caption{
			\label{fig:ii_results_omission-output-cos_flickr30k_14}
			Dataset: Flickr30K, length: 14.
			\vspace{15pt}
		}
	\end{subfigure}
	
	\begin{subfigure}{0.4\textwidth}
		\centering
		\includegraphics[scale=0.5]{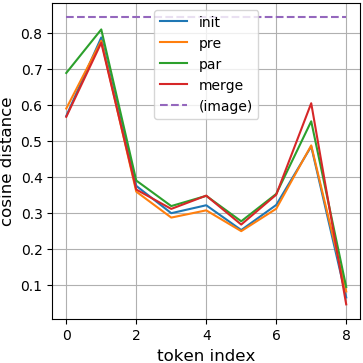}
		\caption{
			\label{fig:ii_results_omission-output-cos_mscoco_7}
			Dataset: MSCOCO, length: 7.
		}
	\end{subfigure}
	\quad
	\begin{subfigure}{0.4\textwidth}
		\centering
		\includegraphics[scale=0.5]{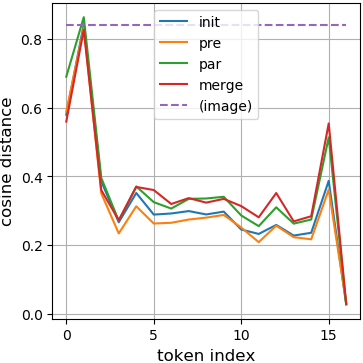}
		\caption{
			\label{fig:ii_results_omission-output-cos_mscococ_15}
			Dataset: MSCOCO, length: 15.
		}
	\end{subfigure}
	\caption{
		\label{fig:ii_results_omission-output-cos}
		Results for cosine distance between output softmaxes resulting from the correct image and foil image. Dashed line is the average cosine distance between the correct image and foil image. Note that the last token index is for the end token.
	}
\end{figure}

For completeness' sake, we also show how the output softmax changes when a foil image is used to see if that changes as much as the multimodal vector. The results of the omission measure applied to the softmax vector are shown in Figure~\ref{fig:ii_results_omission-output-cos}.

\begin{figure}
	\centering
	\figuretitle{Omission with respect to output (JSD)}
	\begin{subfigure}{0.4\textwidth}
		\centering
		\includegraphics[scale=0.5]{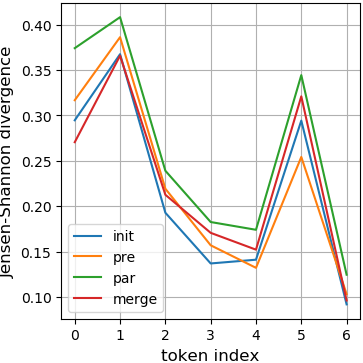}
		\caption{
			\label{fig:ii_results_omission-output-jsd_flickr8k_5}
			Dataset: Flickr8K, length: 5.
			\vspace{15pt}
		}
	\end{subfigure}
	\quad
	\begin{subfigure}{0.4\textwidth}
		\centering
		\includegraphics[scale=0.5]{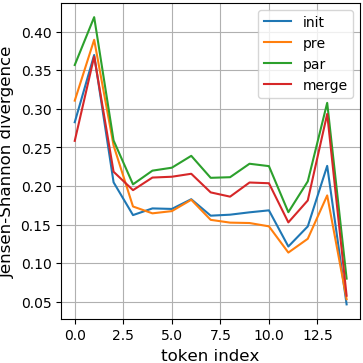}
		\caption{
			\label{fig:ii_results_omission-output-jsd_flickr8k_13}
			Dataset: Flickr8K, length: 13.
			\vspace{15pt}
		}
	\end{subfigure}
	
	\begin{subfigure}{0.4\textwidth}
		\centering
		\includegraphics[scale=0.5]{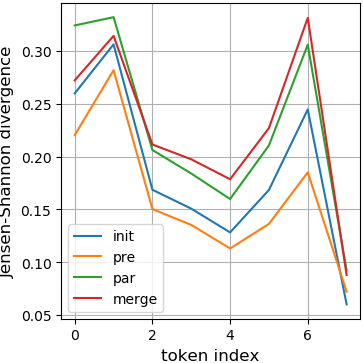}
		\caption{
			\label{fig:ii_results_omission-output-jsd_flickr30k_6}
			Dataset: Flickr30K, length: 6.
			\vspace{15pt}
		}
	\end{subfigure}
	\quad
	\begin{subfigure}{0.4\textwidth}
		\centering
		\includegraphics[scale=0.5]{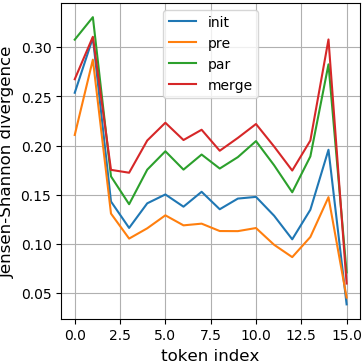}
		\caption{
			\label{fig:ii_results_omission-output-jsd_flickr30k_14}
			Dataset: Flickr30K, length: 14.
			\vspace{15pt}
		}
	\end{subfigure}
	
	\begin{subfigure}{0.4\textwidth}
		\centering
		\includegraphics[scale=0.5]{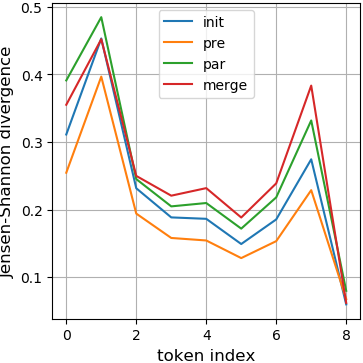}
		\caption{
			\label{fig:ii_results_omission-output-jsd_mscoco_7}
			Dataset: MSCOCO, length: 7.
		}
	\end{subfigure}
	\quad
	\begin{subfigure}{0.4\textwidth}
		\centering
		\includegraphics[scale=0.5]{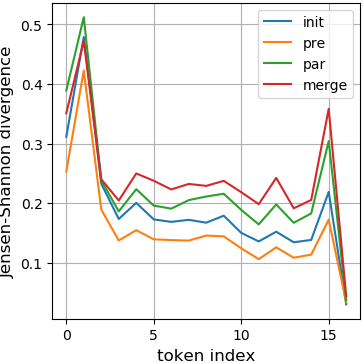}
		\caption{
			\label{fig:ii_results_omission-output-jsd_mscococ_15}
			Dataset: MSCOCO, length: 15.
		}
	\end{subfigure}
	\caption{
		\label{fig:ii_results_omission-output-jsd}
		Results for Jensen-Shannon divergence between output softmaxes resulting from the correct image and foil image. Note that the last token index is for the end token.
	}
\end{figure}

One observation that stands out is that, in the previous figure, the merge architecture has a very different multimodal vector when a foil image is used but here the output vector seems to change just like the other architectures. In fact, all the architectures behave almost identically. The observation is not changed when a different distance measure is used from cosine distance, namely Jensen-Shannon divergence (JSD). JSD is a distance function that is specifically intended for use on probability distributions and is defined as
\begin{align}
\text{JSD}\left(P \middle\| Q\right) &= \frac{1}{2}\left(D\left(P \middle\| \frac{P + Q}{2}\right) + D\left(Q \middle\| \frac{P + Q}{2}\right)\right)\\
D\left(P \middle\| Q\right) &= \sum_x P(x) \log \left(\frac{P(x)}{Q(x)}\right)
\end{align}
where $P$ and $Q$ are equally sized vectors consisting of probabilities that sum to 1. The omission results on the softmax vector using JSD as a distance measure are shown in Figure~\ref{fig:ii_results_omission-output-jsd}.

Although the shapes are still very similar, we can now see a clear ranking in the last peak of each chart. The merge architecture seems to be the most visually influenced architecture, both with respect to the multimodal vector and the output vector. In addition, the merge and par-inject architectures are at the top of the ranking whilst the pre- and init-inject architectures are at the bottom. The charts show that as the number of words being passed into the RNN increases, the distance between these two groups of architectures starts getting wider. This makes sense, as the image is being re-introduced into the model after every time step in the merge and par-inject architectures whereas the other two architectures only see the image once at the beginning.

Seeing the image only once prior to the caption generation process makes it more likely to forget what's in the image as the caption prefix gets longer as there are more and more words that also need to be remembered apart from the image. These two modes of generating descriptions are analogous to a human writing a description of an image by either seeing the image only once and then writing the description from memory or keeping the image visible for the duration of the writing. It may very well be the case that the humans would write different descriptions for the same image given these two different modes of describing.

\begin{figure}
	\centering
	\figuretitle{Omission with respect to logits}
	\begin{subfigure}{0.4\textwidth}
		\centering
		\includegraphics[scale=0.5]{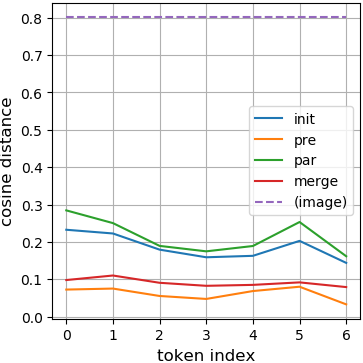}
		\caption{
			\label{fig:ii_results_omission-logits-cos_flickr8k_5}
			Dataset: Flickr8K, length: 5.
			\vspace{15pt}
		}
	\end{subfigure}
	\quad
	\begin{subfigure}{0.4\textwidth}
		\centering
		\includegraphics[scale=0.5]{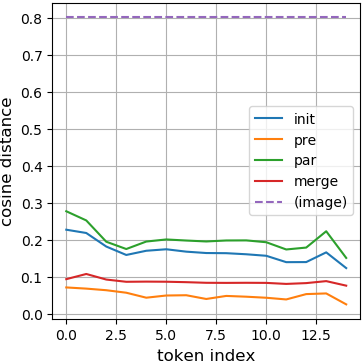}
		\caption{
			\label{fig:ii_results_omission-logits-cos_flickr8k_13}
			Dataset: Flickr8K, length: 13.
			\vspace{15pt}
		}
	\end{subfigure}
	
	\begin{subfigure}{0.4\textwidth}
		\centering
		\includegraphics[scale=0.5]{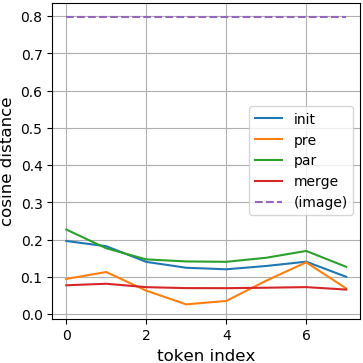}
		\caption{
			\label{fig:ii_results_omission-logits-cos_flickr30k_6}
			Dataset: Flickr30K, length: 6.
			\vspace{15pt}
		}
	\end{subfigure}
	\quad
	\begin{subfigure}{0.4\textwidth}
		\centering
		\includegraphics[scale=0.5]{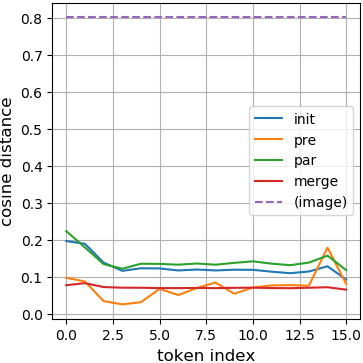}
		\caption{
			\label{fig:ii_results_omission-logits-cos_flickr30k_14}
			Dataset: Flickr30K, length: 14.
			\vspace{15pt}
		}
	\end{subfigure}
	
	\begin{subfigure}{0.4\textwidth}
		\centering
		\includegraphics[scale=0.5]{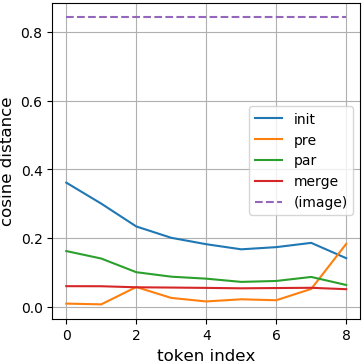}
		\caption{
			\label{fig:ii_results_omission-logits-cos_mscoco_7}
			Dataset: MSCOCO, length: 7.
		}
	\end{subfigure}
	\quad
	\begin{subfigure}{0.4\textwidth}
		\centering
		\includegraphics[scale=0.5]{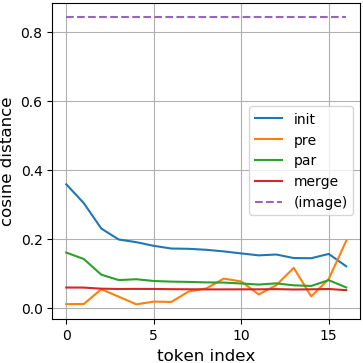}
		\caption{
			\label{fig:ii_results_omission-logits-cos_mscococ_15}
			Dataset: MSCOCO, length: 15.
		}
	\end{subfigure}
	\caption{
		\label{fig:ii_results_omission-logits-cos}
		Results for cosine distance between logits vectors resulting from the correct image and foil image. Dashed line is the average cosine distance between the correct image and foil image. Note that the last token index is for the end token.
	}
\end{figure}

We still need an explanation for why the behaviour of the merge architecture changed so much between the multimodal vector and the softmax output vector. The cause of the anomalous results might be due to the softmax function itself. Applying the omission analysis on the logits of the softmax layer (the values prior to applying the softmax function), shown in Figure~\ref{fig:ii_results_omission-logits-cos}, reveals that the merge architecture's visual influence is still stable at the logits stage, although the relative scale has been changed. This leads us to conclude that the instability originates at the softmax function itself.

\begin{figure}
	\centering
	\figuretitle{Minimum logit}
	\begin{subfigure}{0.4\textwidth}
		\centering
		\includegraphics[scale=0.5]{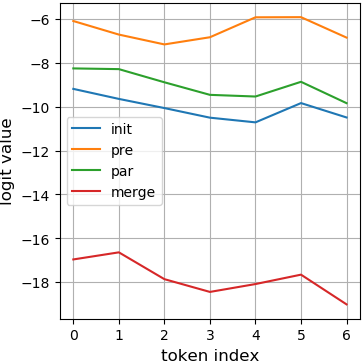}
		\caption{
			\label{fig:ii_min-logit_flickr8k_5}
			Dataset: Flickr8K, length: 5.
			\vspace{15pt}
		}
	\end{subfigure}
	\quad
	\begin{subfigure}{0.4\textwidth}
		\centering
		\includegraphics[scale=0.5]{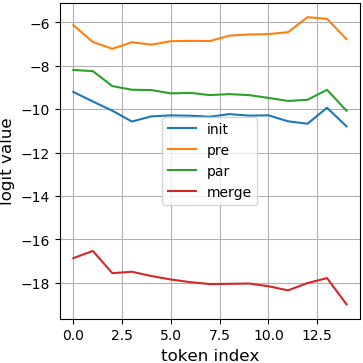}
		\caption{
			\label{fig:ii_min-logit_flickr8k_13}
			Dataset: Flickr8K, length: 13.
			\vspace{15pt}
		}
	\end{subfigure}
	
	\begin{subfigure}{0.4\textwidth}
		\centering
		\includegraphics[scale=0.5]{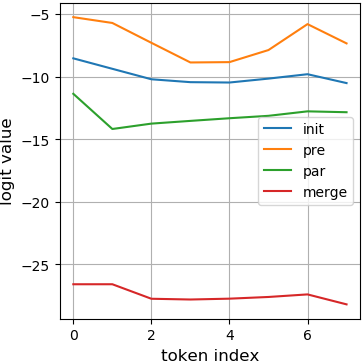}
		\caption{
			\label{fig:ii_min-logit_flickr30k_6}
			Dataset: Flickr30K, length: 6.
			\vspace{15pt}
		}
	\end{subfigure}
	\quad
	\begin{subfigure}{0.4\textwidth}
		\centering
		\includegraphics[scale=0.5]{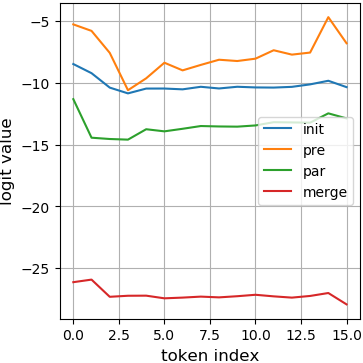}
		\caption{
			\label{fig:ii_min-logit_flickr30k_14}
			Dataset: Flickr30K, length: 14.
			\vspace{15pt}
		}
	\end{subfigure}
	
	\begin{subfigure}{0.4\textwidth}
		\centering
		\includegraphics[scale=0.5]{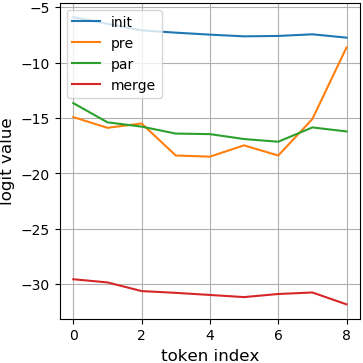}
		\caption{
			\label{fig:ii_min-logit_mscoco_7}
			Dataset: MSCOCO, length: 7.
		}
	\end{subfigure}
	\quad
	\begin{subfigure}{0.4\textwidth}
		\centering
		\includegraphics[scale=0.5]{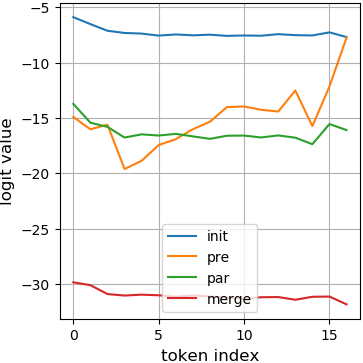}
		\caption{
			\label{fig:ii_min-logit_mscococ_15}
			Dataset: MSCOCO, length: 15.
		}
	\end{subfigure}
	\caption{
		\label{fig:ii_min-logit}
		The minimum logit (produced from correct images) at the given token position. Note that the last token index is for the end token.
	}
\end{figure}

The softmax function eliminates negative numbers in the logits by replacing them with very small positive numbers. Therefore, if the logits vectors have a certain cosine distance due to the numbers in one vector having a different sign (positive or negative) from the corresponding numbers in the other vector or if they are due to differently scaled negative numbers, the softmax function would then eliminate this source of difference. In fact, it turns out that the merge models have the largest magnitude negative numbers in their logits vector, as can be seen from Figure~\ref{fig:ii_min-logit} which shows what the minimum logit value is for each architecture at different time steps.

This gives a tentative explanation for how a stable omission score at the multimodal vector can result in an omission score with large spikes at the output vector. It also helps to keep in mind that, regardless of architecture, the cosine distance between softmax probabilities originating from the correct and foil images cannot remain stable since some word positions do not depend on the image but only on the previous words, such as `a' following `in'. This means that at certain word positions, there must be similar output probabilities (when they do not depend on the image) whilst at other positions there must be different output probabilities (when they do depend on the image).

\FloatBarrier

\subsubsection{Interim summary}

\begin{itemize}
	\item Contrary to what the sensitivity analysis suggests, when the visual influence of the whole layer is measured, init-inject has the least visual influence out of all the architectures.
	
	\item The merge architecture has the greatest and most stable multimodal vector visual omission scores across word positions. This is the case even though the structure of the merge architecture's multimodal vector predicts that it should have a smaller omission score. It also has the most visual influence at the output vector but the amount of influence is not stable across word positions.
	
	\item Whether the image is input just once at the beginning (init- and pre-inject) or at every time step (par-inject and merge) seems to effect how quickly visual influence degrades as the caption gets longer, with the former group degrading the fastest.
	
	\item At the logits vector, merge is still stable but not the greatest and at the softmax vector it is the greatest but not stable.
	
	\item The reason why the behaviour in the merge architecture changes at the output vector is likely because softmax deletes negative numbers from the logits vector and the merge architecture has the largest magnitude negative numbers in the logits vectors of all architectures.
\end{itemize}

\FloatBarrier

\clearpage


\section{Conclusion}

In this chapter we have seen two ways to measure the amount of influence an image has on a caption generator. Sensitivity analysis involves measuring the gradient of the softmax with respect to the image and omission scoring involves measuring the amount of change in a layer's activation values when a different image is used. These two measures are useful for quantifying how much visual information was necessary for each word in the generated caption and are thus useful for partially explaining the behaviour of the neural network.

In general, it seems that the amount of visual information used to predict a word depends significantly on what kind of word is being predicted, such as its part-of-speech. Nouns require more visual information to predict than others, probably because the captions were written to be concrete which makes them tend to mention objects in the image, as well as because the convolution neural network used to extract visual features was trained for object recognition and might be biased toward extracting object-based features. This means that the model is more likely to make mistakes when it comes to verbs and prepositions rather than nouns. In fact, it is possible to get a well-functioning caption generator by only informing the language model of the objects in the image and nothing else \citep{Madhysastha2018}, meaning that the dataset is so stereotyped that you can correctly guess the verbs and prepositions from just the nouns.

It also seems that, although the amount of visual information needed goes down as the word being predicted nears the end of the sentence, it is also likely that inject architectures tend to `forget' visual information since their RNN needs to store both the visual information and an ever growing amount of words as they get generated. The merge architecture on the other hand does not have this problem since the image is kept outside of the RNN and so is not `crowded out' of memory by the words. This does not seem to be an issue in practice however since the init-inject architecture, which suffers the most from this loss of visual information, performs the best in terms of generated captions. It might be because the captions are short enough to avoid losing more than the minimum amount of information needed to produce a coherent caption. It could also be the case that words towards the end of the captions are predictable enough from the prefix alone that the image is not crucial beyond a certain number of words.

For convenience, here are the architectures sorted by their general performance:
\begin{itemize}
	\item According to image sensitivity: init-inject, merge, par-inject, pre-inject.
	
	\item According to omission score at multimodal vector: merge, pre-inject, par-inject, init-inject.
	
	\item According to omission score at output vector: merge, par-inject, init-inject, pre-inject.
\end{itemize}

Contrary to the previous two chapters, the next chapter will not discuss any more comparisons between architectures. Instead we take one architecture and talk about how to pre-train its language-handling part on a corpus of text to boost its performance.

\clearpage

%% file: tex/chp5_transferlearning.tex
\chapterwithfootnote{Transfer learning using the merge architecture}{This chapter has been released separately as a pre-print paper \citep{Tanti2019}.}
\clearpage

\section{Aims}

Up to now we have focussed on comparing the different caption generation architectures in order to tease out any advantages one architecture might have over the others. But there is one practical advantage of the merge architecture over the others which has to do with transfer learning. In this chapter we will investigate what this advantage is and perform experiments with it.

As has been explained in previous chapters, image caption generators make use of a convolutional neural network, typically pre-trained on a separate image-only dataset, in order to extract visual features from images. Using pre-trained neural networks to transform inputs into high level features for other neural networks makes it easier to avoid overfitting \citep{Vinyals2015}. In this chapter we investigate whether the language part of the image caption generator can also be handled by a pre-trained neural network that has been trained on a separate text-only dataset.

The language encoding part of a caption generator is the recurrent neural network (RNN) together with the embedding layer. We collectively call the parameters of these two layers `prefix encoding parameters' because they encode a partially generated caption prefix into a single vector. The source model from which we want to transfer these parameters is a trained neural language model and the target model is an untrained image caption generator.

\begin{figure}
	\centering
	\includegraphics[scale=0.8]{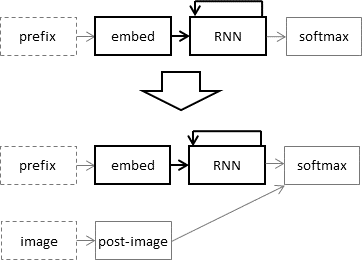}
	\caption{
		\label{fig:lt_architecture}
		The architectures of the language model (top) and caption generator (bottom) with the part that is transferred having a bold outline.
	}
\end{figure}

Not all caption generator architectures allow for this kind of parameter transferring. If the image is provided as an initial state to the RNN, as in the case of the init-inject architecture, then the image would need to be taken into account when training the RNN and hence cannot be trained separately. The merge architecture on the other hand leaves the vision encoding part and language encoding part of the caption generator separate, which allows each to be trained separately. An illustration of what gets transferred is shown in Figure~\ref{fig:lt_architecture}.

\clearpage


\section{Background}

Transfer learning \citep{Pan2010} is the act of exploiting the knowledge gained by a trained model in order to improve another model's learning process in a different task or domain. In NLP tasks, it is common to transfer word embeddings from other tasks, such as word2vec \citep{Mikolov2013} and GloVe \citep{Pennington2014}, in order to have more meaningful word vectors which were optimised using large text corpora. Apart from the word embedding layer, there is also work on transferring the RNN layer as well \citep{Zoph2016,Ramachandran2017,Howard2018,Mou2016}. \citet{Howard2018} conjecture that language modelling is a universal source task for transfer learning in natural language processing. They perform transfer learning by using a language model as an initial sentence encoder for classification tasks such as sentiment analysis, question classification, and topic classification. \citet{Ramachandran2017} also perform transfer learning from the source task of language modelling but transfer to generative tasks like machine translation and abstractive summarisation instead.

As in the present work, \citet{Ramachandran2017} implement an architecture that makes it possible to re-use the unconditioned source language model in a target task that requires a conditioned language model, which we solved by using the merge architecture. In their case, they made it possible to use an init-inject architecture by using two RNNs in series. The first RNN encoded the input text only. The second RNN was initialised with the source sentence vector (for a translation task) but took as input the hidden state vectors of the first. Hence, the first RNN could be pre-trained. In our case, the merge architecture obviates the need for a dual RNN architecture, which is less computationally demanding.

Another important consideration in transfer learning is the relative value of freezing parameters versus fine-tuning during training. \citet{Yosinski2014} performed experiments on transfer learning in CNNs and tried transferring a variable number of layers from the input side of the neural network. With regards to the difference between freezing and fine-tuning, the first half of the layers tend to encode features that when frozen are difficult to exploit by randomly initialised later layers. Fine-tuning allows the transferred and randomly initialised layers to co-operate at reaching a suitable middle-ground representation. This observation has been called fragile co-adaptation. \citet{Mou2016} found similar results for text classification tasks. When transferring the embedding layer and RNN, performance is always better when the transferred parameters are fine-tuned rather than left frozen. \citet{Mou2016} also found that when transferring between similar tasks, such as from a sentiment analysis task to a different sentiment analysis task, performance always improves when both the embedding layer and the RNN are transferred. On the other hand, when transferring between unrelated tasks, such as from a sentiment analysis task to a question classification task, only transferring the embedding layer without the RNN improves performance.

A third consideration in transfer learning is the relationship between the performance of the source model on the source task and the performance of the transferred model on the target task. In particular, we show below that the best performing language model does not transfer as well as lesser-performing models. \citet{Kornblith2018} found similar results with CNNs. Image features produced by different CNNs were used as inputs to logistic regressors that were trained to perform an image classification task. It turns out that the state of the art CNNs do not produce the best fixed image features (according to the final trained logistic regressor's performance) but that it was some of the lesser CNNs that do. On a similar note, \citet{Hessel2015} found something similar with image caption generators. Different pre-trained CNNs were used to extract image features to be used for training image caption generators. It was found that using AlexNet \citep{Krizhevsky2012} results in slightly better captions than using VGG-16 \citep{Simonyan2014}, even though the first CNN has an object recognition top-1 accuracy of 57.1\% whilst the second's accuracy is 75.6\%. These results are evidence that as a neural network becomes better at performing the source task, its internal feature representations become overspecialised for the task and become less useful for performing other tasks.

Finally, the work of \citet{Hendricks2016} is similar to what we have done here. An image caption generator that uses the merge architecture (like we do) was developed in such a way that the vision handling part and the language handling part can be trained separately and then combined together. However, this work focusses on ways to extend the model's vocabulary after training whereas we focus on treating the language handling part of a caption generator in the same way as the vision handling part: using a pre-trained neural network to extract fixed features.

\clearpage


\section{Experiments}

There are several factors that could affect the performance of the final caption generator, and rather than arbitrarily picking one configuration over the others, we set our experiments to vary across three dimensions:
\begin{itemize}
	\item Domain: The corpus training data for the language model varies from data sampled from the image caption domain to data sampled from general (news) text.

	\item Corpus size: The size of the corpus that is used to train the language model is varied from being one tenth to ten times the size of the captions dataset.

	\item Frozen prefix encoding parameters: After transferring the embedding layer and RNN parameters to the caption generator, they can either be frozen or fine-tuned during training of the caption generator.
\end{itemize}

Our experiments consist of comparing the performance of the resulting caption generators with that of the non-transferred merge architecture that was described in the previous chapters. We also compare the performance of the caption generators with that of the language models they are derived from.

In these experiments, we train and evaluate image caption generators on Flickr8K \citep{Hodosh2013}. The rationale for doing so is that this is a relatively small dataset, which enables us to experiment with language models that are trained on corpora up to 10 times the number of sentences of the target corpus using our available hardware resources.

We train the neural language model on one of three different corpora. Each corpus's domain is of a varying degree of similarity to the final caption dataset's domain (Flickr8K). This allows us to see how the performance of the final caption generator changes as its parameters are transferred from more and more distant source domains. The three corpora are
\begin{itemize}
	\item Same captions: An in-domain corpus consisting of the sentences in the Flickr8K dataset itself, which is a performance ceiling since the source corpus cannot have a closer domain to the target corpus than the target corpus itself.
	
	\item Different captions: Another in-domain corpus consisting of the sentences in the MSCOCO \citep{Lin2014} dataset, which is another image captions corpus but which is different and larger than Flickr8K.
	
	\item General text: An out-of-domain corpus consisting of the sentences in the Google one billion token language modelling benchmark corpus\footnote{See: \url{https://github.com/ciprian-chelba/1-billion-word-language-modeling-benchmark}} (LM1B), which is a corpus consisting mostly of news texts.
\end{itemize}

Flickr8K and MSCOCO were both obtained from the distributed versions provided by \citet{Karpathy2015}\footnote{See: \url{http://cs.stanford.edu/people/karpathy/deepimagesent/}}.

We also vary the size of the language model corpus training sets in order to measure the effect of size apart from domain, where sizes are measured in number of sentences. To vary corpus size, a random sample of sentences from one of the above corpora is selected as a subcorpus. Each size of the subcorpus is computed as a multiple of the number of captions in Flickr8K (each caption is one sentence long), where the multiple is an exponent of 10. This allows us to measure how the performance of the caption generator changes as the corpus size is changed on a logarithmic scale, which gives us corpora sizes of 3\,000 sentences ($10^{-1}$ of Flickr8K), 9\,487 sentences ($10^{-0.5}$ of Flickr8K), 30\,000 sentences ($10^{0}$ of Flickr8K), 94\,868 sentences ($10^{0.5}$ of Flickr8K), and 300\,000 sentences ($10^{1}$ of Flickr8K).

Finally, we alternate between using frozen and fine-tuned transferred parameters. Freezing the parameters means that the prefix encoding parameters of the caption generator are not changed during training and are left as they were in the source language model. Fine-tuning the parameters means allowing them to be further optimised whilst training the caption generator.

Table~\ref{tbl:lt_configurations} shows all the different experimental configurations being compared. As a reference for comparing performance, we also train a caption generator on Flickr8K without using transfer learning (`no transfer').

\begin{table}
	\centering
	\begin{tabular}{c|cc|cc}
		 &	\multicolumn{2}{c|}{Language model} &	\multicolumn{2}{c}{Caption generator} \\
		Type &	Dataset &	Size multiple ($10^x$)  &	Dataset &	Frozen/Fine tuned \\
		\hline
		No transfer &	N/A &	N/A &	Flickr8K &	`Fine tuned' only \\
		Same captions &	Flickr8K &	-1, -0.5, 0 &	Flickr8K &	Frozen and fine-tuned \\
		Different captions &	MSCOCO &	-1, -0.5, 0, 0.5, 1 &	Flickr8K &	Frozen and fine-tuned \\
		General text &	LM1B &	-1, -0.5, 0, 0.5, 1 &	Flickr8K &	Frozen and fine-tuned \\
	\end{tabular}
	\caption{
		\label{tbl:lt_configurations}Experimental configurations that were compared. Size multiple refers to the number of sentences in the language model corpus such that a size multiple of $x$ means that the corpus has $10^x$\textsuperscript{th} the number of sentences in Flickr8K ($10^x \times 30\,000$). Frozen/Fine tuned refers to whether the transferred parameters were frozen during training of the caption generator or allowed to be optimised together with the rest of the parameters.
	}
\end{table}

In all these corpora and in all their sizes, only words that occur at least 5 times in their respective training subcorpus were included in the vocabulary, with the remaining words being replaced by the unknown token. The vocabulary is one of the challenges one will encounter when doing transfer learning of language models as it would differ between the language model's training set and the caption generator's training set. We set the vocabulary of the caption generator to be the intersection between the vocabulary extracted from the language model corpus and the vocabulary that would have been extracted from the captions dataset. This means that the final vocabulary of the caption generator would be smaller than that of a non-transferred caption generator, with any word out of the vocabulary being replaced by the unknown token. Note that this is what would happen in practice when an off-the-shelf language model is used to initialise a caption generator. Note also that the language model's vocabulary would not be affected by the caption generator's vocabulary.

All sentences were preprocessed by lowercasing all characters, replacing strings of digits with a `NUM' token, and removing all non-alphanumeric non-space characters. In order to reduce the memory requirements of running these experiments, the LM1B corpus was filtered so that only sentences that have no more than 50 tokens were included.

Early stopping was used on both language model training and caption generator training such that training is stopped on the epoch when the geometric mean of the perplexity on the validation set is worse than it was in the previous epoch. These validation sets were the ones provided with each dataset (Flickr8K, MSCOCO, and LM1B) which means that they are in the same domain as the training set. This is so that the language model would be closer to what would be expected from an off-the-shelf language model. Furthermore, the validation sets were not varied in size with the training sets. It is worth noting that in the case of `same captions', the validation set used for training the language model is the same as the one used whilst training the caption generator.

A GRU \citep{Chung2014} was used as an RNN for both language models and caption generators. Biases were initialised to zero. Image features were extracted from layer `fc7' (the penultimate layer) of the VGG OxfordNet 16-layer convolutional neural network \citep{Simonyan2014}. All of this is similar to what was mentioned in the previous chapters.

\subsection{Hyperparameter tuning}

The hyperparameters of each of the four different experimental configurations (rows shown in Table~\ref{tbl:lt_configurations}), both of the language model and of the derived caption generator, were tuned independently and automatically using Bayesian optimisation. In order to avoid spending too much time on tuning, only the training set size multiple of $10^0$ was used together with frozen prefix encoding parameters when tuning for each configuration. The rest of the variations on the same row in Table~\ref{tbl:lt_configurations} shared the same hyperparameters found. The prefix encoding parameters of the (trained) best language model (found whilst tuning its hyperparameters) are transferred to the caption generator when it is being tuned. The reason for choosing this method is that, in practice, the language model is tuned independently of the caption generation task, while the caption generator itself would be tuned to take advantage of the pre-trained language model. The `no transfer' model is exactly the same as the merge architecture described in the previous chapters.

Just like in the previous chapters, the library Scikit-Optimize\footnote{See: \url{https://scikit-optimize.github.io/}} was used to perform hyperparameter tuning using Bayesian optimisation. As an optimisation cost function, geometric mean of perplexity was used for the language model whilst the Word Mover's Distance (WMD) metric \citep{Kusner2015,Kilickaya2017} was used on the caption generator. This process was performed twice for each hyperparameter combination and the average perplexity or WMD resulting from the two independent train and generation sessions was used as a score for the hyperparameter combination. This makes the score more robust than if the model was only trained and evaluated once. As a reminder, the model, whose purpose is to predict the fitness of a given hyperparameter combination, is a random forest and was initialised using 32 random hyperparameter combinations paired with their evaluated fitness (the perplexity or WMD). The hyperparameter combinations were then optimised by exploring a sequence of 64 candidate hyperparameters that the model suggests will maximise the expected improvement in fitness. The best hyperparameters found for each configuration are shown in Table~\ref{tbl:lt_hyperparams}. Again, the `no transfer' model's hyperparameters were copied over from those of Chapter 3. Refer back to Subsection~\ref{sec:wi_hyperparameter_tuning} for more information regarding each hyperparameter.

\begin{table}
	\centering
	\begin{subtable}{\textwidth}
		\centering
		\begin{small}
			\begin{tabular}{l|cccc}
				 &	No transfer &	Same captions &	Different captions &	General text \\
				\hline
				weight init. method &	N/A &	Xavier &	Normal &	Xavier \\
				max. init. weight &	N/A &	1.72e-01 &	8.25e-02 &	4.45e-01 \\
				embed size &	276 &	502 &	255 &	132 \\
				RNN size &	227 &	201 &	427 &	330 \\
				optimiser &	N/A &	RMSProp &	Adam &	Adam \\
				learning rate &	N/A &	2.49e-03 &	8.34e-04 &	5.98e-03 \\
				weight decay weight &	N/A &	3.72e-10 &	4.21e-08 &	1.45e-05 \\
				embedding dropout rate &	0.01 &	0.01 &	0.07 &	0.03 \\
				RNN dropout rate &	N/A &	0.33 &	0.13 &	0.23 \\
				max. gradient norm &	N/A &	6.96 &	7.54 &	47.90 \\
				minibatch size &	N/A &	210 &	104 &	68 \\
			\end{tabular}
		\end{small}
		\caption{
			\label{tbl:lt_hyperparams_langmod}
			Hyperparameters for the language models.
		}
	\end{subtable}
	\vspace{10pt}

	\begin{subtable}{\textwidth}
		\centering
		\begin{small}
			\begin{tabular}{l|cccc}
				 &	No transfer &	Same captions &	Different captions &	General text \\
				\hline
				weight init. method &	Xavier &	Xavier &	Xavier &	Normal \\
				max. init. weight &	1.96e-01 &	2.43e-03 &	3.06e-04 &	4.52e-05 \\
				post-image size &	268 &	430 &	511 &	307 \\
				post-image activation &	ReLU &	ReLU &	ReLU &	none \\
				optimiser &	Adam &	RMSProp &	Adam &	Adam \\
				learning rate &	2.64e-04 &	2.83e-04 &	4.59e-05 &	1.30e-03 \\
				normalise image &	false &	false &	false &	true \\
				weight decay weight &	3.01e-07 &	2.45e-04 &	1.18e-10 &	2.87e-10 \\
				image dropout rate &	0.02 &	0.06 &	0.13 &	0.20 \\
				post-image dropout rate &	0.21 &	0.29 &	0.01 &	0.31 \\
				RNN dropout rate &	0.28 &	0.41 &	0.18 &	0.01 \\
				max. gradient norm &	685.80 &	366.97 &	841.50 &	153.06 \\
				minibatch size &	237 &	227 &	18 &	162 \\
				beam width &	5 &	4 &	4 &	4 \\
			\end{tabular}
		\end{small}
		\caption{
			\label{tbl:lt_hyperparams_capgen}
			Hyperparameters for the caption generators.
		}
	\end{subtable}
	\vspace{10pt}

	\caption{
		\label{tbl:lt_hyperparams}
		Best hyperparameters found for each experimental configuration.
	}
\end{table}

\clearpage


\section{Results}

Each experiment was run five times, each time using a different randomly chosen subset of the corpus sentences to train the language model (as well as having different initial random weights, minibatches, and other non-deterministic values). The mean of the results for the quality of generated captions using METEOR \citep{Banerjee2005}, CIDEr \citep{Vedantam2015}, SPICE \citep{Anderson2016}, and WMD \citep{Kusner2015,Kilickaya2017} is shown in Table~\ref{tbl:lt_results_exp}.

\begin{table}
	\centering
	\begin{small}
		\begin{tabular}{ccr|cccc}
			Type &	Frozen? &	Size &	METEOR &	CIDEr &	SPICE &	WMD \\
			\hline
			no trans. &	no &	30\,000 &	{\underline{0.190}} (0.001) &	{\underline{0.457}} (0.011) &	{\underline{0.128}} (0.002) &	{\underline{0.137}} (0.003) \\
			\hline
			\hline
			same caps. &	yes &	3\,000 &	{{0.189}} (0.002) &	{{0.467}} (0.022) &	{{0.126}} (0.003) &	{{0.139}} (0.004) \\
			same caps. &	yes &	9\,487 &	{{0.193}} (0.001) &	{{0.476}} (0.015) &	{{0.131}} (0.002) &	{{0.140}} (0.002) \\
			same caps. &	yes &	30\,000 &	{\underline{0.194}} (0.002) &	{{0.476}} (0.010) &	{\underline{0.132}} (0.001) &	{{0.140}} (0.002) \\
			\hline
			same caps. &	no &	3\,000 &	{{0.190}} (0.004) &	{{0.447}} (0.017) &	{{0.127}} (0.003) &	{{0.137}} (0.003) \\
			same caps. &	no &	9\,487 &	{{0.194}} (0.001) &	\textbf{\underline{0.485}} (0.009) &	{{0.130}} (0.002) &	\textbf{\underline{0.141}} (0.002) \\
			same caps. &	no &	30\,000 &	{{0.194}} (0.003) &	{{0.478}} (0.011) &	{{0.131}} (0.003) &	{{0.140}} (0.003) \\
			\hline
			\hline
			diff. caps. &	yes &	3\,000 &	{{0.187}} (0.001) &	{{0.431}} (0.007) &	{{0.124}} (0.001) &	{{0.136}} (0.002) \\
			diff. caps. &	yes &	9\,487 &	{{0.189}} (0.001) &	{{0.451}} (0.006) &	{{0.126}} (0.002) &	{{0.139}} (0.001) \\
			diff. caps. &	yes &	30\,000 &	{{0.191}} (0.001) &	{{0.475}} (0.007) &	{{0.130}} (0.001) &	{{0.140}} (0.002) \\
			diff. caps. &	yes &	94\,868 &	{{0.192}} (0.002) &	{{0.478}} (0.009) &	{{0.132}} (0.002) &	{\underline{0.141}} (0.003) \\
			diff. caps. &	yes &	300\,000 &	{{0.190}} (0.001) &	{{0.469}} (0.010) &	{{0.130}} (0.001) &	{{0.139}} (0.002) \\
			\hline
			diff. caps. &	no &	3\,000 &	{{0.190}} (0.002) &	{{0.438}} (0.005) &	{{0.126}} (0.002) &	{{0.137}} (0.002) \\
			diff. caps. &	no &	9\,487 &	{{0.191}} (0.002) &	{{0.459}} (0.013) &	{{0.129}} (0.002) &	{{0.138}} (0.002) \\
			diff. caps. &	no &	30\,000 &	\textbf{\underline{0.195}} (0.002) &	{{0.482}} (0.010) &	\textbf{\underline{0.133}} (0.002) &	{{0.140}} (0.003) \\
			diff. caps. &	no &	94\,868 &	{{0.194}} (0.002) &	{{0.471}} (0.013) &	{{0.132}} (0.002) &	{{0.137}} (0.002) \\
			diff. caps. &	no &	300\,000 &	{{0.194}} (0.002) &	{\underline{0.482}} (0.008) &	{{0.133}} (0.001) &	{{0.140}} (0.002) \\
			\hline
			\hline
			gen. text &	yes &	3\,000 &	{{0.145}} (0.006) &	{{0.245}} (0.017) &	{{0.071}} (0.005) &	{{0.095}} (0.004) \\
			gen. text &	yes &	9\,487 &	{{0.171}} (0.004) &	{{0.364}} (0.015) &	{{0.113}} (0.003) &	{{0.123}} (0.003) \\
			gen. text &	yes &	30\,000 &	{{0.182}} (0.002) &	{{0.425}} (0.004) &	{{0.122}} (0.001) &	{{0.134}} (0.001) \\
			gen. text &	yes &	94\,868 &	{{0.183}} (0.002) &	{{0.446}} (0.011) &	{{0.125}} (0.002) &	{{0.135}} (0.003) \\
			gen. text &	yes &	300\,000 &	{{0.186}} (0.002) &	{\underline{0.453}} (0.010) &	{{0.127}} (0.001) &	{\underline{0.137}} (0.002) \\
			\hline
			gen. text &	no &	3\,000 &	{{0.156}} (0.003) &	{{0.220}} (0.011) &	{{0.074}} (0.002) &	{{0.097}} (0.002) \\
			gen. text &	no &	9\,487 &	{{0.183}} (0.002) &	{{0.370}} (0.008) &	{{0.118}} (0.003) &	{{0.124}} (0.002) \\
			gen. text &	no &	30\,000 &	{{0.187}} (0.001) &	{{0.419}} (0.011) &	{{0.125}} (0.001) &	{{0.130}} (0.003) \\
			gen. text &	no &	94\,868 &	{{0.187}} (0.001) &	{{0.431}} (0.015) &	{{0.125}} (0.002) &	{{0.133}} (0.004) \\
			gen. text &	no &	300\,000 &	{\underline{0.190}} (0.002) &	{{0.440}} (0.012) &	{\underline{0.128}} (0.002) &	{{0.134}} (0.002) \\
			\hline
		\end{tabular}
	\end{small}
	\caption{
		\label{tbl:lt_results_exp}
		Results for the final generated captions after transfer learning. Underlined values are the best results for each experiment type whilst boldfaced values are the best results across all types. The `no transfer' values were copied from Table~\ref{tbl:wi_results_exp_qty_flickr8k}. Legend: no trans. - no transfer learning, frozen - frozen parameters (vs. fine-tuned), size - corpus size, diff. caps. - different captions, gen. text - general text.
	}
\end{table}

\subsection{Transfer learning versus non-transfer learning}

Transfer learning always improves over non-transfer learning. In fact, for WMD and CIDEr, the best value shown here is better than the values obtained by any architecture shown in Table~\ref{tbl:wi_results_exp_qty_flickr8k} in Chapter~3. Looking at the WMD scores, fine-tuning only improves the caption generator's performance more than freezing for the `same captions' corpus, and even then it is by a minuscule amount, meaning that the prefix encoding parameters are transferable between language models and caption generators as is. Regarding the language model corpus, domain plays an important role: the `general text' corpus (LM1B) never performs better than an in-domain corpus with 9\,487 sentences, even when 300\,000 sentences are used. In fact, when 300\,000 sentences are used to train the `general text' language model, the resulting caption generator performance is on a par with the performance obtained by the `no transfer' model.

It is interesting that simply pre-training the prefix encoding parameters on the text of the same captions dataset that will be used to train the caption generator will improve the performance of the final caption generator. This fact could be of great practical importance when training neural networks, possibly as a form of smart initialisation where the caption generator's prefix encoding parameters are initialised at a sensible point in parameter space. It could be argued that this is instead the result of more effective hyperparameter tuning due to a transferred caption generator having less hyperparameters to optimise: the embedding and RNN sizes are determined and fixed by the source language model whilst the non-transferred caption generator needs to optimise them as well. This is in fact however a practical advantage of transfer learning where the dimensionality of the hyperparameter search space is reduced.

\FloatBarrier

\subsubsection{Interim summary}

\begin{itemize}
	\item Transferring the prefix encoding parameters of a language model to a caption generator without fine tuning, just as is done with a convolutional neural network, is possible and leads to improvements over not using transfer learning.
	
	\item Transfer learning with the merge architecture results in better performance than any other architecture with no transfer learning (on Flickr8K).
	
	\item Transfer learning from a non-captions corpus requires 10 times the number of sentences in a caption dataset in order to achieve equal performance to non-transfer learning on the caption dataset.
	
	\item Transfer learning on the text of the same captions dataset (that the resulting caption generator will be trained on) results in better performance than not using transfer learning.
\end{itemize}

\subsection{Size of language model corpus}

\begin{figure}
	\centering
	\begin{subfigure}{0.3\textwidth}
		\includegraphics[scale=0.5]{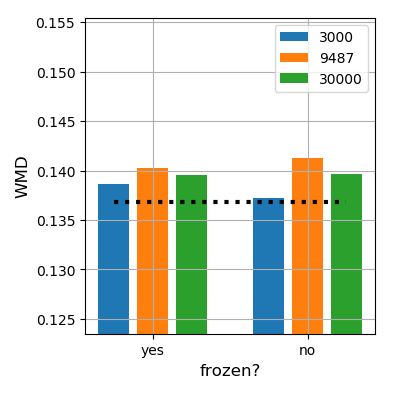}
		\caption{
			\label{results_wmd_samecaps}
			In-domain: same captions
		}
	\end{subfigure}
	\quad
	\begin{subfigure}{0.3\textwidth}
		\includegraphics[scale=0.5]{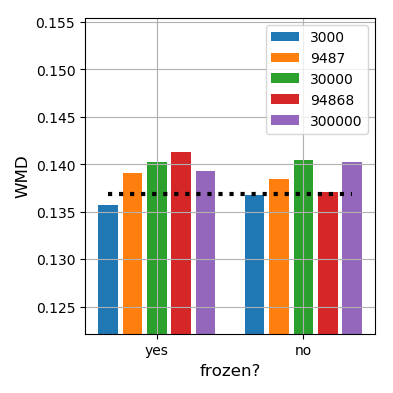}
		\caption{
			\label{results_wmd_diffcaps}
			In-domain: different captions
		}
	\end{subfigure}
	\quad
	\begin{subfigure}{0.3\textwidth}
		\includegraphics[scale=0.5]{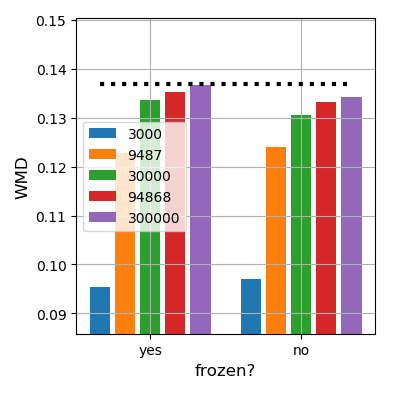}
		\caption{
			\label{results_wmd_gentext}
			Out-of-domain: general text
		}
	\end{subfigure}
	\caption{
		\label{fig:results_wmd}
		The WMD score of the different language models measured on the Flickr8K test set captions after transfer learning. Each colour bar shows the number of sentences that the language model was trained on. The black dotted line indicates what WMD score the `no transfer' model obtained. Note how the largest in-domain corpora do not result in the best WMD. As for the out-of-domain corpus, whilst there is a monotonic increase in WMD with corpus size, it also does not perform as well as the in-domain corpora.
	}
\end{figure}

One important observation in Table~\ref{tbl:lt_results_exp} is that increasing the language model corpus size does not automatically increase the resulting caption generator performance. In the case of the `same captions' and `different captions' models, on most quality metrics, pre-training on part of the language model corpus gives a better performance than pre-training on the largest size. Figure~\ref{fig:results_wmd} shows more clearly how the caption generator's WMD score changes as the language model's corpus size changes.

The reason for the performance getting worse after a certain point in corpus size could be related to the language model's performance. In order to check for this, we measured the geometric mean of the perplexity of the language models on their respective validation set corpus in order to measure the performance of the language model rather than that of the caption generator.

\subsubsection{Comparing perplexities given differently sized vocabularies}

One must be careful when comparing the perplexity of language models with different vocabularies as we have done. This is because the unknown token gives an unfair advantage to smaller vocabularies. To understand this, imagine if, in the extreme case, all the words were omitted from the vocabulary and were all replaced with the unknown token. This would make every word almost perfectly predictable (save for the end of sentence token) and the language model would assign almost 100\% of the probability to the unknown token each time, leading to a perplexity that is almost perfect. Adding words to the vocabulary would then make the prediction more uncertain and thus lead to a worse perplexity, with larger vocabularies resulting in more uncertainty than smaller ones. This means that language models with smaller vocabularies have an unfair advantage over language models with larger vocabularies, regardless of what probabilities they actually output.

To get around this problem and make the unknown token behave more fairly, we divide the probability assigned to the unknown token by the number of different words it replaces in the corpus we are using it on. For example, if the number of different words in the corpus being used to evaluate the language model is 1\,000 and the known vocabulary covers 400 of those word types, then the unknown token will be replacing the remaining 600 word types. Given that the language model does not give any information about those 600 word types, we assume that they are uniformly distributed and assume that the probability assigned by the language model to the unknown token is evenly divided between all those 600 word types. Now, whenever we encounter an unknown token in the corpus, we replace its probability $p$ by $\frac{p}{600}$. This effectively makes the vocabulary size equal to 1\,000 again, with the 600 out-of-vocabulary words dividing the unknown token's probability mass equally among themselves. This eliminates the advantage gained by smaller vocabularies as all vocabularies are effectively of equal size now.

\subsubsection{Perplexity results}

\begin{figure}
	\centering
	\begin{subfigure}{0.3\textwidth}
		\includegraphics[scale=0.5]{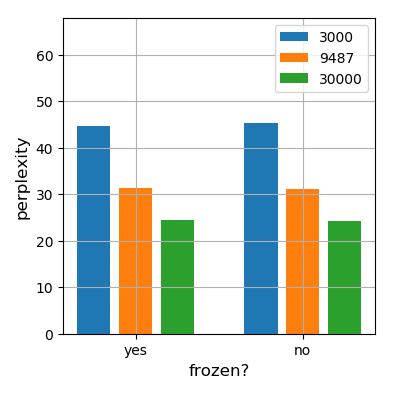}
		\caption{
			\label{lt_results_pplx_samecaps}
			In-domain: same captions
		}
	\end{subfigure}
	\quad
	\begin{subfigure}{0.3\textwidth}
		\includegraphics[scale=0.5]{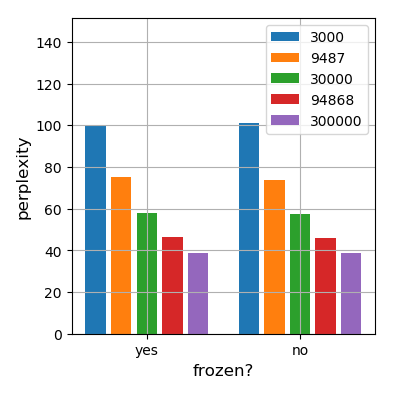}
		\caption{
			\label{lt_results_pplx_diffcaps}
			In-domain: different captions
		}
	\end{subfigure}
	\quad
	\begin{subfigure}{0.3\textwidth}
		\includegraphics[scale=0.5]{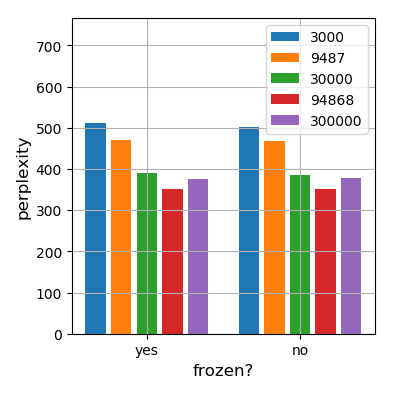}
		\caption{
			\label{lt_results_pplx_gentext}
			Out-of-domain: general text
		}
	\end{subfigure}
	\caption{
		\label{fig:lt_results_pplx}
		The language model perplexity of the different language models measured on their respective validation set corpus. Each colour bar shows the number of sentences that the language model was trained on. Smaller perplexities are better. Note how the perplexity keeps improving when larger corpora are used which means that the reason for the degradation in the transferred caption generator performance is not due to a degradation in the performance of the source language model.
	}
\end{figure}

Figure~\ref{fig:lt_results_pplx} shows the perplexity of every language model on its corresponding validation set corpus. It shows that the language models do not perform worse as the corpus size grows, apart from the general text corpus. The general text language model might start performing worse after a certain corpus size due to the vocabulary being too large for the hyperparameters chosen for the language model (which were tuned using a smaller corpus). On the other hand, the performance of the caption generator that was transferred from the largest general text corpus language model was the best. There is an interesting relationship between the performance of the language model and the performance of the transferred caption generator which we can highlight with a scatter plot.

\begin{figure}
	\centering
	\begin{subfigure}{0.3\textwidth}
		\includegraphics[scale=0.5]{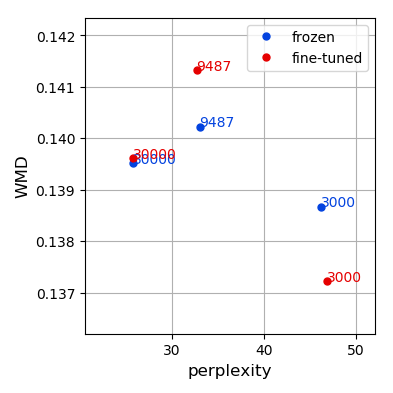}
		\caption{
			\label{lt_results_pplx-wmd_samecaps}
			In-domain: same captions
		}
	\end{subfigure}
	\quad
	\begin{subfigure}{0.3\textwidth}
		\includegraphics[scale=0.5]{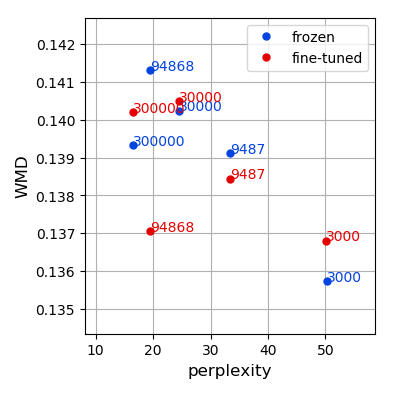}
		\caption{
			\label{lt_results_pplx-wmd_diffcaps}
			In-domain: different captions
		}
	\end{subfigure}
	\quad
	\begin{subfigure}{0.3\textwidth}
		\includegraphics[scale=0.5]{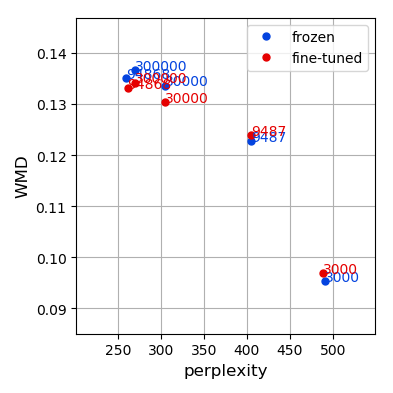}
		\caption{
			\label{lt_results_pplx-wmd_gentext}
			Out-of-domain: general text
		}
	\end{subfigure}
	\caption{
		\label{fig:lt_results_pplx-wmd}
		Scatter plot relating language model perplexity and transferred caption generator WMD score. To the right of each point is the number of sentences used to train the language model. Points that are vertically higher have a better WMD score whilst points that are further to the left have a better perplexity. Note how, in general, the best perplexity does not give the best WMD score.
	}
\end{figure}

Figure~\ref{fig:lt_results_pplx-wmd} shows a scatter plot that illustrates the relationship between language model perplexity and the transferred caption generator WMD. The WMD score starts off correlating with the perplexity until the best perplexity is reached, at which point the WMD score dips. There seems to be an exception in the case of the fine-tuned version of the `different captions' model where an outlier seems to disrupt the trend but everywhere else the trend is preserved.

So why does the WMD score stop correlating with the corpus size after a point? We explain this as a case of overspecialisation. A language model that is too good at language modelling will produce an internal representation of the sentence prefix that is too specialised in language modelling to be also useful (and hence transferable) to the caption generation task. We are thus in agreement with \citet{Kornblith2018} that state of the art neural networks in a source task might not be ideal to perform transfer learning in a target task.

\subsubsection{Interim summary}

\begin{itemize}
	\item The resulting caption generator's WMD performance does not increase monotonically with the corpus size that the language model is trained on.
	
	\item The language model's perplexity improves with every increase in corpus size but this stops correlating with the resulting caption generator's WMD beyond a certain perplexity value. We hypothesise that this is due to the language model becoming overspecialised in language modelling and the internal representations not being generic enough to be useful in other tasks.
\end{itemize}

\subsection{Partial training of language models}

The previous results led us to a new hypothesis that, rather than varying the perplexity of the language model by varying the training corpus sizes, we can instead prematurely stop the language model's training process before peak validation performance is reached and check if this will also lead to better transferability. We tested this hypothesis by partially training the language model for a fixed number of epochs before transferring the prefix encoding parameters and measuring the resulting caption generator's WMD score.

Given a number of epochs $n$, we trained the language model for $n$ epochs and then transferred its prefix encoding parameters to the caption generator. We varied $n$ to be between 0 (no language model training, just transfer the random parameters as is) and 15. For each $n$, we retrain both the language model and caption generator for five times, just like the main experiments described above.

We only accepted a trained language model if its perplexity on the validation set kept improving after every epoch. If not, we started training over again. If after five attempts at re-training a language model the validation perplexity kept peaking before reaching the $n$\textsuperscript{th} epoch, we terminated training there, and recorded at which $n$ this happened. We then continued increasing $n$ without early stopping in order to see if an overfitted language model (overfitted according to the language model's validation set) resulted in a better or worse caption generator. The largest $n$ we reached with early stopping (no overfitting) was 13.

\begin{figure}
	\centering
	\begin{subfigure}{0.3\textwidth}
			\includegraphics[scale=0.5]{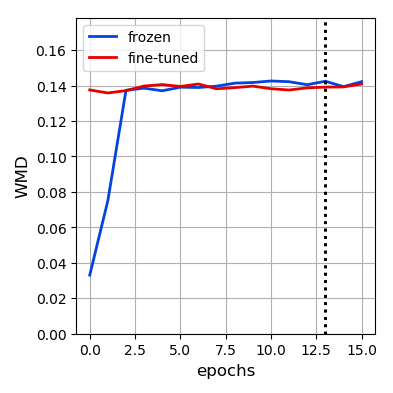}
			\caption{
				\label{lt_results_partialtrain_wmd_samecaps}
				In-domain: same captions. Best WMD (0.142) reached at epoch 10 with frozen prefix encoding parameters. Overfitting occurred after epoch 13.\\
			}
		\end{subfigure}
		\quad
		\begin{subfigure}{0.3\textwidth}
			\includegraphics[scale=0.5]{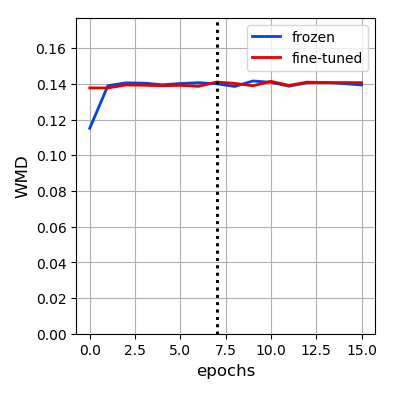}
			\caption{
				\label{lt_results_partialtrain_wmd_diffcaps}
				In-domain: different captions. Best WMD (0.141) reached at epoch 9 with frozen prefix encoding parameters. Overfitting occurred after epoch 7.
			}
		\end{subfigure}
		\quad
		\begin{subfigure}{0.3\textwidth}
			\includegraphics[scale=0.5]{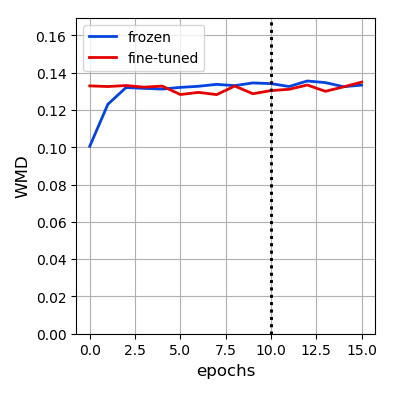}
			\caption{
				\label{lt_results_partialtrain_wmd_gentext}
				Out-of-domain: general text. Best WMD (0.136) reached at epoch 12 with frozen prefix encoding parameters. Overfitting occurred after epoch 10.
			}
		\end{subfigure}
	\caption{
		\label{fig:lt_results_partialtrain_wmd}
		How the WMD score in the transferred caption generator changes as the language model is trained for a varying number of epochs. The dark vertical line shows the last epoch before the language model started overfitting (in terms of perplexity measured on the validation set).
	}
\end{figure}

For each language model corpus, we included both frozen and fine-tuned prefix encoding parameters. Since this experiment takes a long time to complete, we only used one corpus size which is 30\,000 sentences, that is, the full size of Flickr8K, with the other corpora consisting of a random sample of sentences as described before. The results, shown in Figure~\ref{fig:lt_results_partialtrain_wmd}, reveal several interesting points.

After training for two epochs, the transferred frozen parameters are sufficient for generating captions that are close to the best quality. In fact, for the `different captions' and `general text' corpora, using the randomly initialised prefix encoding parameters and leaving them frozen (frozen zero epoch training) does not result in substantially degraded performance.

This can be explained on the grounds that although an RNN has random weights, it does not act non-deterministically. In fact, it still encodes something about the prefix of the partially generated caption; it is just that the encoding is not optimised to work well on the training set. The neural layer that reads and processes the RNN's hidden state vector can still find some useful features about the prefix from the `wild' encoding given by a random RNN. This recalls findings in research on echo state networks \citep{Jaeger2004}, where a randomly initialised simple RNN is frozen during training and only the layer that reads the RNN's hidden state vector is trained. With echo state networks, however, the RNN's random weights are supposed to have a spectral radius that is less than one, whereas here we do not use any such restriction.

Although the `same captions' corpus does result in substantially degraded performance for the frozen zero epoch training version, we determined that this is due to differences in the chosen hyperparameters. In fact, we found that a single hyperparameter change can give substantial improvement to the frozen zero epoch training version: initialising weights using a normal distribution instead of using Xavier initialisation. This results in a WMD of 0.118, exceeding the WMD of 0.115 of frozen zero epoch `different captions'. Xavier initialisation uses a smaller variance in random values which probably results in most of the activations in the RNN's hidden state vector being similar and hence making it harder to accidentally encode enough useful distinct features. That said, the frozen zero epoch training version of the `general text' language model still works relatively well with Xavier initialisation so this explanation is not complete.

Finally, and most importantly, partial training does not seem to have the same effect as limiting the amount of data to train on. Whereas the previous results showed that changing the amount of data results in a predictable change in performance, changing the amount of epochs for which to train the language model seems to result in a more haphazard change in performance.

\subsubsection{Interim summary}

\begin{itemize}
	\item Partial training does not seem to have the same effect as training on less data when it comes to avoiding overspecialisation in the language model.
	
	\item Randomly set prefix encoding parameters work reasonably well if left frozen when training the rest of the caption generator. Randomly determined features can still accidentally include useful features.
\end{itemize}

\clearpage


\section{Conclusion}

In this chapter we have taken advantage of an interesting property of the merge architecture, which is that its RNN and embedding layer can be pre-trained on text-only data. This is done by training text-only language model on a corpus and then transferring its RNN and embedding layer parameters to those of the caption generator. Inject architectures do not allow for this to happen easily as their RNN needs to be trained with visual information present since it is inputted into the RNN.

We have shown that a caption generator benefits from having its embedding layer and RNN transferred from a language model, but only when transferring from an in-domain corpus. When a general text corpus is used rather than captions, the model does not perform better than a non-pre-trained caption generator.

We also found that simply pre-training the embedding layer and RNN on the text of the same captions dataset that it will eventually be trained on after transferring is also beneficial, even if less data is used to pre-train the RNN and embedding layer. Interestingly, this will only work if the language model is exclusively trained on a fraction of the dataset. Beyond a certain number of sentences in the training corpus, the performance of the final caption generator begins to drop. We called this phenomenon overspecialisation, which is when the internal features produced by the language model become overly specialised in the task of language modelling, which interferes with their ability to be useful in other tasks such as the caption generation task. This is not the same as overfitting, which is when the model memorises the training data and does not generalise. An overspecialised model can still generalise, but only in one task.

Furthermore, partial training of the language model does not seem like a reliable way to prevent this overspecialisation from happening as the performance of models that were trained for a fixed number of epochs seems haphazard at best.

\clearpage

%% file: tex/chp6_conclusion.tex
\chapter{Conclusion}
\clearpage

\section{Summary and answers to research questions}

We have determined from the literature on caption generation that there are four main neural network architectures for including visual information into neural language models. These architectures were named init-inject, pre-inject, par-inject, and merge, and are illustrated in Figure~\ref{fig:bg_cnnlm} in Subsection~\ref{sec:combining_cnns_with_nlms}.

We started this thesis with research questions about the theoretical nature of grounding language in vision followed by three more specific questions. We shall now be answering these questions.

\subsection{Research questions 1 and 2}
{\it Should the visual information be encoded together with the words as a single mixed representation?}

\noindent {\it Should it be introduced just once at the beginning only?}\\

Different measures to rank these architectures by their performance agreed that the best performing architecture is the init-inject architecture. This was confirmed by both automatic and human evaluation. This is evidence in favour of mixed representations and one-time image input. There could be a number of reasons for this.

One-time image input architectures like init-inject and pre-inject have their image layer size tied to another layer size such as the embedding layer size or the RNN size. This (1) creates a shared image representation with the RNN state/embedding representation that could result in a better represention for both and (2) it reduces the number of hyperparameters needed to tune them since there would be fewer independent layer sizes to change, which makes it more likely to find a quasi-optimal hyperparameter combination.

Another reason is that, since the image representation is stored inside the RNN's hidden state vector together with the caption prefix information, and since the hidden state vector changes for every word in the prefix, then the image representation gets to evolve with every word in the prefix. This might have an effect similar to attention mechanisms where the image representation changes depending on what has been already generated.

Finally, the absolute values of the initial state of the GRU are not bounded by a squashing function and can be as large as needed. This might help init-inject work better. This can also be the case with LSTM provided that the cell state is what is initialised with the image.

On the other hand, an analysis of the visual influence on the output of the models showed that the visual influence on init-inject goes down quickly as the number of words generated grows. This means that init-inject might suffer when generating long captions. The visual influence of par-inject and merge goes down more slowly meaning that these two architectures could work better when longer captions are generated.

\subsection{Research question 3}
{\it What are the merits of each architecture?}\\

Here is a summary of all the interesting features observed in each architecture throughout the whole thesis:

\paragraph{Init-inject:} The init-inject architecture was found to generally be the best performing architecture according to our experiments. It also has its most probable next word being the most sensitive to the image content, the fact that the image representation is shared by the RNN state representation might allow it to learn to perform well with a small model, and it has fewer hyperparameters to tune due to having tied layer sizes. Finally, it seems that the GRU is well suited for init-inject since it allows its initial state to grow without bound from a squashing function and the LSTM might also be well suited if the cell state is initialised with the image.

\paragraph{Pre-inject:} Although it produced captions using very high frequency words only, the pre-inject architecture happened to generate the most fluent sounding captions. Also, the fact that the image representation is shared by the embedding word representation might allow it to learn to perform well with a small model and it has fewer hyperparameters to tune due to having tied layer sizes.

\paragraph{Par-inject:} The par-inject architecture requires a large model in order to perform well. On the other hand, the fact that the image is re-input after every time step allows it to keep the caption influenced on the image more than the other two inject architectures. It is likely that as the caption gets longer, par-inject will perform better than the other inject architectures as its visual information is continuously refreshed rather than provided once at the beginning only.

\paragraph{Merge:} Although it does not have any representation shared with the that of the image, the merge architecture can still perform well with small models. It also has its most probable next word being the most sensitive to the previous word, its output probabilities as a whole are the most influenced by the image data, and it is likely that as the caption gets longer, merge will perform better than init- and pre-inject as its RNN's hidden state vector does not need to remember any visual features, meaning that it is better able to remember a long caption prefix. Merge is also the only architecture that can copy parameters over from a text-only language model without further fine-tuning, and as a result performs better than all the other architectures.

\subsection{Research question 4}
{\it Are there architectures that are influenced less by visual information than others?}\\

Yes, the merge architecture is the one whose entire output probability distribution gets changed the most by changing the image whilst the init-inject architecture is the architecture whose maximum output probability is most sensitive to the image.

The init-inject architecture is also the one which loses its sensitivity the fastest as the caption gets generated one word at a time. This could be because, as the caption gets longer, the RNN needs to remember all the words that were generated in order for the model to predict the next sensible word. Given that the RNN's hidden state vector is finite, this linguistic information would start to `crowd out' the visual information. It is possible that crowding out the visual information might even be a strategic optimisation given the RNN's finite state size since the generated words might be enough to predict the next word even if the image is partly forgotten.

\subsection{Research question 5}
{\it Is it possible to train the visual and linguistic parts of the merge architecture separately?}\\

Yes, and not only is it possible but the performance improves as well and it performs better than the init-inject architecture. We accomplished this by first pre-training the embedding layer and RNN in a language model and then passing the trained parameters over to the merge architecture. This works even when the same text of the captions dataset is used to pre-train the embedding layer and RNN. This could be due to a number of reasons.

Transfer learning results in fewer independent hyperparameters to tune as the layer sizes in the caption generator will be tied to those of the language model in order to be compatible. This makes it more likely to find a quasi-optimal hyperparameter combination.

It is possible to leave the embedding layer and RNN parameters set randomly during training and the caption generator's performance will not go down drastically. This means that most of the work is done by the rest of the model which is even able to work with a randomly set RNN. Therefore, providing a pre-trained RNN would not disrupt the rest of the model's training and could instead start the model off from a sensible starting point from which to train.

Although pre-training RNNs in language models generally improves the caption generator's performance, it is worth noting that better language models do not necessarily translate into better caption generators. If the language model's performance goes beyond a certain point, the resulting caption generator starts to perform worse than it would with a worse language model. This means that the language model would be overspecialised in language modelling and the internal representations would not be useful in other tasks.

\clearpage

\section{Limitations}

The present work opens up several new questions which, in retrospect, also suggest alternatives to some of the methodological decisions taken in the experiments reported in the previous chapters.

After automatically searching for suitable hyperparameters to use when training each neural network architecture, we only took the best hyperparameter combination found for each architecture and used only that throughout the experiments. We then retrained and evaluated each model using the same hyperparameter combination five times and took the average of the five results. This explores the variation in results from the same model definition but does not explore how sensitive the architecture is to changes in hyperparameters. It would be more informative if the top 5 (different) hyperparameter combinations found were used such that a different one is used for each run.

In the human evaluation of the generated captions it would have been useful to also ask whether a given caption was pleasing as well as accurate and fluent. Although human-written captions were annotated as less fluent than the automatically generated ones, this does not mean that the generated captions would be preferred over the human-written ones. The variation in a human-written caption might be more pleasing to read than the rigorously structured generated captions, even if the latter were more fluent. As the evaluation stands, we know whether the generated captions are accurate and fluent but not if they are desirable.

Also in the human evaluation, the fact that human-written captions were shown together with the automatically generated ones might have influenced the annotators towards being more judgemental towards the less accurate machine generated captions. It would have been better to show one caption per image and to only ask about that caption. It would also have been better if the fluency question was asked separately from the accuracy question and if the image was not shown when asking it. Unfortunately this would have also required more annotators in order to be able to still have multiple responses per question.

There are also some minor details in the experiments that might make more sense had they been changed but this was realised after all the experiments were complete. These are:

\begin{itemize}
	\item Rather than use a uniform distribution as a possible hyperparameter for initialising random weights, it would have been better to use only normal distributions and use the standard deviation of the normal distribution to control the range of weights instead of clipping them.
	
	\item Rather than replacing all strings of digits with a single pseudo-word, it would have been better to have treated each digit as a separate token in order to avoid losing information whilst still avoiding low frequency strings of digits.
\end{itemize}

\clearpage

\section{Future work}

The work presented here opens up several avenues for further research. Here we have listed some ideas we would like to pursue in the future.

In the present work we have investigated each individual architecture in isolation, but in the literature several systems consist of combining two or three architectures at once such as init-inject together with par-inject together with merge \citep{Xu2015}. An analysis of the effect of combining these architectures together would be useful in order to investigate whether emergent properties develop where the whole is greater than the sum of its parts.

As stated in the introduction, we do not use attention mechanisms in the present work because that would typically exclude init-inject and pre-inject architectures. It would be interesting to see if the use of attention mechanisms works better on merge rather than par-inject, but now that we know that init-inject gives the best performance, perhaps it is also worth considering the use of init-inject in attention mechanisms by using the less efficient basic language model training method described in Subsection~\ref{sec:bg_basic_lm}.

In chapter 5 we performed transfer learning on the merge architecture by pre-training the RNN and embedding layer on a text corpus and then evaluating the resulting caption generator on the small Flickr8K dataset. Flickr8K was used in order to train on text corpora that are up to 10 times larger than the captions dataset without requiring more hardware than we had available. Nonetheless, an interesting observation was that even by just training the RNN and embedding layer on the text part of Flickr8K itself, better performance would be obtained than when not using transfer learning. We would now like to see whether this would also happen if the captions dataset was the large MSCOCO dataset and the text corpus was the text part of the MSCOCO dataset itself.

It would also be interesting to confirm what the effect of using an LSTM instead of a GRU would be and to confirm that using init-inject on the cell state would work better than on the hidden state.

Finally, we would like to apply the experiments performed in the present work on tasks other than image caption generation. It would be interesting to see whether the same results hold for other conditioned language models such as those used in machine translation.

\clearpage

%% file: ms.bbl
\begin{thebibliography}{133}
\providecommand{\natexlab}[1]{#1}
\providecommand{\url}[1]{\texttt{#1}}
\expandafter\ifx\csname urlstyle\endcsname\relax
  \providecommand{\doi}[1]{doi: #1}\else
  \providecommand{\doi}{doi: \begingroup \urlstyle{rm}\Url}\fi

\bibitem[Anderson et~al.(2016)Anderson, Fernando, Johnson, and
  Gould]{Anderson2016}
Peter Anderson, Basura Fernando, Mark Johnson, and Stephen Gould.
\newblock Spice: Semantic propositional image caption evaluation.
\newblock In \emph{ECCV}, 2016.

\bibitem[Antol et~al.(2015)Antol, Agrawal, Lu, Mitchell, Batra,
  Lawrence~Zitnick, and Parikh]{Antol2015}
Stanislaw Antol, Aishwarya Agrawal, Jiasen Lu, Margaret Mitchell, Dhruv Batra,
  C.~Lawrence~Zitnick, and Devi Parikh.
\newblock Vqa: Visual question answering.
\newblock In \emph{ICCV}, December 2015.

\bibitem[Bahdanau et~al.(2014)Bahdanau, Cho, and Bengio]{Bahdanau2014}
Dzmitry Bahdanau, Kyunghyun Cho, and Yoshua Bengio.
\newblock Neural machine translation by jointly learning to align and
  translate.
\newblock \emph{CoRR}, September 2014.

\bibitem[Banerjee and Lavie(2005)]{Banerjee2005}
Satanjeev Banerjee and Alon Lavie.
\newblock {METEOR}: {An} automatic metric for {MT} evaluation with improved
  correlation with human judgments.
\newblock In \emph{Proc. Workshop on Intrinsic and Extrinsic Evaluation
  Measures for Machine Translation And/or Summarization}, volume~29, pages
  65--72, 2005.

\bibitem[Bengio et~al.(2015)Bengio, Vinyals, Jaitly, and Shazeer]{Bengio2015}
Samy Bengio, Oriol Vinyals, Navdeep Jaitly, and Noam Shazeer.
\newblock Scheduled sampling for sequence prediction with recurrent neural
  networks.
\newblock In C.~Cortes, N.~D. Lawrence, D.~D. Lee, M.~Sugiyama, and R.~Garnett,
  editors, \emph{NIPS}, pages 1171--1179. Curran Associates, Inc., 2015.
\newblock URL
  \url{http://papers.nips.cc/paper/5956-scheduled-sampling-for-sequence-prediction-with-recurrent-neural-networks}.

\bibitem[Bengio et~al.(2003)Bengio, Ducharme, Vincent, and Jauvin]{Bengio2003}
Yoshua Bengio, R{\'{e}}jean Ducharme, Pascal Vincent, and Christian Jauvin.
\newblock A neural probabilistic language model.
\newblock \emph{JMLR}, 3:\penalty0 1137--1155, February 2003.
\newblock \doi{10.1162/153244303322533223}.

\bibitem[Bernardi et~al.(2016)Bernardi, Cakici, Elliott, Erdem, Erdem,
  Ikizler-cinbis, Keller, Muscat, and Plank]{Bernardi2016}
Raffaella Bernardi, Ruket Cakici, Desmond Elliott, Aykut Erdem, Erkut Erdem,
  Nazli Ikizler-cinbis, Frank Keller, Adrian Muscat, and Barbara Plank.
\newblock {Automatic} {Description} {Generation} from {Images}: {A} {Survey} of
  {Models}, {Datasets}, and {Evaluation} {Measures}.
\newblock \emph{JAIR}, 55:\penalty0 409--442, 2016.

\bibitem[Chen and Zitnick(2014)]{Chen2014}
Xinlei Chen and C.~Lawrence Zitnick.
\newblock Learning a recurrent visual representation for image caption
  generation.
\newblock \emph{CoRR}, 1411.5654, 2014.

\bibitem[Chen and Zitnick(2015)]{Chen2015}
Xinlei Chen and C.~Lawrence Zitnick.
\newblock Mind's eye: A recurrent visual representation for image caption
  generation.
\newblock In \emph{Proc. CVPR}, 2015.

\bibitem[Chen et~al.(2015)Chen, Fang, Lin, Vedantam, Gupta, Dollar, and
  Zitnick]{Chen2015a}
Xinlei Chen, Hao Fang, Tsung-yi Lin, Ramakrishna Vedantam, Saurabh Gupta, Piotr
  Dollar, and C.~Lawrence Zitnick.
\newblock Microsoft coco captions: Data collection and evaluation server.
\newblock \emph{CoRR}, April 2015.

\bibitem[Choromanska et~al.(2015)Choromanska, Henaff, Mathieu, Arous, and
  Lecun]{Choromanska2015}
Anna Choromanska, Mikael Henaff, Michael Mathieu, Gerard~Ben Arous, and Yann
  Lecun.
\newblock {The Loss Surfaces of Multilayer Networks}.
\newblock In Guy Lebanon and S.~V.~N. Vishwanathan, editors, \emph{AISTATS},
  volume~38 of \emph{Proceedings of Machine Learning Research}, pages 192--204,
  San Diego, California, Usa, 2015. Pmlr.
\newblock URL \url{http://proceedings.mlr.press/v38/choromanska15.html}.

\bibitem[Chung et~al.(2014)Chung, G{\"{u}}l{\c{c}}ehre, Cho, and
  Bengio]{Chung2014}
Junyoung Chung, {\c{C}}aglar G{\"{u}}l{\c{c}}ehre, Kyunghyun Cho, and Yoshua
  Bengio.
\newblock {Empirical} {Evaluation} of {Gated} {Recurrent} {Neural} {Networks}
  on {Sequence} {Modeling}.
\newblock \emph{CoRR}, 1412.3555, 2014.

\bibitem[Collobert et~al.(2011)Collobert, Weston, Bottou, Karlen, Kavukcuoglu,
  and Kuksa]{Collobert2011}
Ronan Collobert, Jason Weston, L\'eon Bottou, Michael Karlen, Koray
  Kavukcuoglu, and Pavel Kuksa.
\newblock Natural language processing (almost) from scratch.
\newblock \emph{JMLR}, 12:\penalty0 2493--2537, August 2011.
\newblock URL \url{http://leon.bottou.org/papers/collobert-2011}.

\bibitem[Deng et~al.(2009)Deng, Dong, Socher, Li, Li, and Fei-fei]{Deng2009}
Jia Deng, Wei Dong, Richard Socher, Li-jia Li, Kai Li, and Li~Fei-fei.
\newblock {ImageNet}: A large-scale hierarchical image database.
\newblock In \emph{Proc. CVPR}, 2009.

\bibitem[Devlin et~al.(2015)Devlin, Cheng, Fang, Gupta, Deng, He, Zweig, and
  Mitchell]{Devlin2015}
Jacob Devlin, Hao Cheng, Hao Fang, Saurabh Gupta, Li~Deng, Xiaodong He,
  Geoffrey Zweig, and Margaret Mitchell.
\newblock Language models for image captioning: The quirks and what works.
\newblock In \emph{IJCNLP}, pages 100--105, Beijing, China, July 2015.
\newblock URL \url{http://aclweb.org/anthology/P/P15/P15-2017.pdf}.

\bibitem[Devlin et~al.(2018)Devlin, Chang, Lee, and Toutanova]{Devlin2018}
Jacob Devlin, Ming{-}wei Chang, Kenton Lee, and Kristina Toutanova.
\newblock {BERT:} pre-training of deep bidirectional transformers for language
  understanding.
\newblock \emph{CoRR}, Abs/1810.04805, 2018.
\newblock URL \url{http://arxiv.org/abs/1810.04805}.

\bibitem[Donahue et~al.(2015)Donahue, Hendricks, Guadarrama, Rohrbach,
  Venugopalan, Saenko, and Darrell]{Donahue2015}
Jeff Donahue, Lisa~Anne Hendricks, Sergio Guadarrama, Marcus Rohrbach,
  Subhashini Venugopalan, Kate Saenko, and Trevor Darrell.
\newblock {Long-term} {Recurrent} {Convolutional} {Networks} for {Visual}
  {Recognition} and {Description}.
\newblock In \emph{Proc. CVPR}, 2015.

\bibitem[Eggensperger et~al.(2015)Eggensperger, Hutter, Hoos, and
  Leyton-brown]{Eggensperger2015}
Katharina Eggensperger, Frank Hutter, Holger Hoos, and Kevin Leyton-brown.
\newblock Efficient benchmarking of hyperparameter optimizers via surrogates.
\newblock In \emph{AAAI}, 2015.
\newblock URL
  \url{https://www.aaai.org/ocs/index.php/AAAI/AAAI15/paper/view/9993}.

\bibitem[Eldan and Shamir(2016)]{Eldan2016}
Ronen Eldan and Ohad Shamir.
\newblock The power of depth for feedforward neural networks.
\newblock In Vitaly Feldman, Alexander Rakhlin, and Ohad Shamir, editors,
  \emph{COLT}, volume~49 of \emph{Proceedings of Machine Learning Research},
  pages 907--940, Columbia University, New York, New York, Usa, 2016. Pmlr.
\newblock URL \url{http://proceedings.mlr.press/v49/eldan16.html}.

\bibitem[Elliott and Keller(2013)]{Elliott2013}
Desmond Elliott and Frank Keller.
\newblock Image description using visual dependency representations.
\newblock In \emph{EMNLP}, pages 1292--1302, Seattle, Washington, Usa, October
  2013. Association for Computational Linguistics.
\newblock URL \url{http://www.aclweb.org/anthology/D13-1128}.

\bibitem[Elman(1990)]{Elman1990}
Jeffrey~L. Elman.
\newblock Finding structure in time.
\newblock \emph{Cognitive Science}, 14\penalty0 (2):\penalty0 179--211, March
  1990.
\newblock \doi{10.1207/s15516709cog1402_1}.

\bibitem[Frank and Bod(2011)]{Frank2011}
Stefan~L. Frank and Rens Bod.
\newblock Insensitivity of the human sentence-processing system to hierarchical
  structure.
\newblock \emph{Psychological Science}, 22\penalty0 (6):\penalty0 829--834, May
  2011.
\newblock \doi{10.1177/0956797611409589}.

\bibitem[Fukui et~al.(2016)Fukui, Park, Yang, Rohrbach, Darrell, and
  Rohrbach]{Fukui2016}
Akira Fukui, Dong~Huk Park, Daylen Yang, Anna Rohrbach, Trevor Darrell, and
  Marcus Rohrbach.
\newblock Multimodal compact bilinear pooling for visual question answering and
  visual grounding.
\newblock In \emph{EMNLP}, pages 457--468, Austin, Texas, November 2016.
  Association for Computational Linguistics.
\newblock \doi{10.18653/v1/D16-1044}.
\newblock URL \url{https://www.aclweb.org/anthology/D16-1044}.

\bibitem[Gers et~al.(2002)Gers, Schraudolph, and Schmidhuber]{Gers2002}
Felix~A. Gers, Nicol~N. Schraudolph, and J{\"{u}}rgen Schmidhuber.
\newblock Learning precise timing with {LSTM} recurrent networks.
\newblock \emph{JMLR}, 3:\penalty0 115--143, 2002.
\newblock URL \url{http://www.jmlr.org/papers/v3/gers02a.html}.

\bibitem[Glorot and Bengio(2010)]{Glorot2010}
Xavier Glorot and Yoshua Bengio.
\newblock Understanding the difficulty of training deep feedforward neural
  networks.
\newblock In \emph{PMLR}, volume~9, pages 249--256, 2010.

\bibitem[Goodfellow et~al.(2014)Goodfellow, Pouget-abadie, Mirza, Xu,
  Warde-farley, Ozair, Courville, and Bengio]{Goodfellow2014}
Ian Goodfellow, Jean Pouget-abadie, Mehdi Mirza, Bing Xu, David Warde-farley,
  Sherjil Ozair, Aaron Courville, and Yoshua Bengio.
\newblock Generative adversarial nets.
\newblock In Z.~Ghahramani, M.~Welling, C.~Cortes, N.~D. Lawrence, and K.~Q.
  Weinberger, editors, \emph{NIPS}, pages 2672--2680. Curran Associates, Inc.,
  2014.
\newblock URL
  \url{http://papers.nips.cc/paper/5423-generative-adversarial-nets.pdf}.

\bibitem[Goodfellow et~al.(2016)Goodfellow, Bengio, and
  Courville]{Goodfellow2016}
Ian Goodfellow, Yoshua Bengio, and Aaron Courville.
\newblock \emph{Deep Learning}.
\newblock MIT Press, 2016.
\newblock \url{http://www.deeplearningbook.org}.

\bibitem[Graves et~al.(2016)Graves, Wayne, Reynolds, Harley, Danihelka,
  Grabska-barwi{\'{n}}ska, Colmenarejo, Grefenstette, Ramalho, Agapiou, Badia,
  Hermann, Zwols, Ostrovski, Cain, King, Summerfield, Blunsom, Kavukcuoglu, and
  Hassabis]{Graves2016}
Alex Graves, Greg Wayne, Malcolm Reynolds, Tim Harley, Ivo Danihelka, Agnieszka
  Grabska-barwi{\'{n}}ska, Sergio~G{\'{o}}mez Colmenarejo, Edward Grefenstette,
  Tiago Ramalho, John Agapiou, Adri{\`{a}}~Puigdom{\`{e}}nech Badia,
  Karl~Moritz Hermann, Yori Zwols, Georg Ostrovski, Adam Cain, Helen King,
  Christopher Summerfield, Phil Blunsom, Koray Kavukcuoglu, and Demis Hassabis.
\newblock Hybrid computing using a neural network with dynamic external memory.
\newblock \emph{Nature}, 538\penalty0 (7626):\penalty0 471--476, October 2016.
\newblock \doi{10.1038/nature20101}.

\bibitem[Gr{\"o}nroos et~al.(2018)Gr{\"o}nroos, Huet, Kurimo, Laaksonen,
  Merialdo, Pham, Sj{\"o}berg, Sulubacak, Tiedemann, Troncy, and
  V{\'a}zquez]{Groenroos2018}
Stig-arne Gr{\"o}nroos, Benoit Huet, Mikko Kurimo, Jorma Laaksonen, Bernard
  Merialdo, Phu Pham, Mats Sj{\"o}berg, Umut Sulubacak, J{\"o}rg Tiedemann,
  Raphael Troncy, and Ra{\'u}l V{\'a}zquez.
\newblock The memad submission to the wmt18 multimodal translation task.
\newblock In \emph{WMT}, pages 603--611, Belgium, Brussels, October 2018.
  Association for Computational Linguistics.
\newblock URL \url{https://www.aclweb.org/anthology/W18-6439}.

\bibitem[Gu et~al.(2017)Gu, Wang, Cai, and Chen]{Gu2017}
Jiuxiang Gu, Gang Wang, Jianfei Cai, and Tsuhan Chen.
\newblock An empirical study of language {CNN} for image captioning.
\newblock In \emph{ICCV}, October 2017.

\bibitem[Gupta et~al.(2012)Gupta, Verma, and Jawahar]{Gupta2012}
Ankush Gupta, Yashaswi Verma, and C.~V. Jawahar.
\newblock Choosing linguistics over vision to describe images.
\newblock In \emph{AAAI}, Aaai'12, pages 606--612, Toronto, Ontario, Canada,
  2012. Aaai Press.
\newblock URL \url{http://dl.acm.org/citation.cfm?id=2900728.2900815}.

\bibitem[Harnad(1990)]{Harnad1990}
Stevan Harnad.
\newblock The symbol grounding problem.
\newblock \emph{Physica D}, 42:\penalty0 335--346, 1990.

\bibitem[He et~al.(2016)He, Zhang, Ren, and Sun]{He2016}
Kaiming He, Xiangyu Zhang, Shaoqing Ren, and Jian Sun.
\newblock Deep residual learning for image recognition.
\newblock In \emph{CVPR}. {IEEE}, June 2016.
\newblock \doi{10.1109/cvpr.2016.90}.

\bibitem[Hendricks et~al.(2016)Hendricks, Venugopalan, Rohrbach, Mooney,
  Saenko, and Darrell]{Hendricks2016}
Lisa~Anne Hendricks, Subhashini Venugopalan, Marcus Rohrbach, Raymond Mooney,
  Kate Saenko, and Trevor Darrell.
\newblock {Deep} {Compositional} {Captioning}: {Describing} {Novel} {Object}
  {Categories} without {Paired} {Training} {Data}.
\newblock In \emph{Proc. CVPR}, 2016.

\bibitem[Hessel et~al.(2015)Hessel, Savva, and Wilber]{Hessel2015}
Jack Hessel, Nicolas Savva, and Michael~J. Wilber.
\newblock Image representations and new domains in neural image captioning.
\newblock In \emph{VL}, pages 29--39, Lisbon, Portugal, September 2015.
\newblock \doi{10.18653/v1/W15-2807}.

\bibitem[Hochreiter and Schmidhuber(1997)]{Hochreiter1997}
Sepp Hochreiter and J{\"u}rgen Schmidhuber.
\newblock {Long} short-term memory.
\newblock \emph{Neural Computation}, 9\penalty0 (8):\penalty0 1735--1780, 1997.

\bibitem[Hochreiter et~al.(2000)Hochreiter, Bengio, Frasconi, and
  Schmidhuber]{Hochreiter2000}
Sepp Hochreiter, Yoshua Bengio, Paolo Frasconi, and J{\"{u}}rgen Schmidhuber.
\newblock Gradient flow in recurrent nets: The difficulty of learning longterm
  dependencies.
\newblock In \emph{A Field Guide to Dynamical Recurrent Networks}, pages
  237--244. {IEEE}, 2000.
\newblock \doi{10.1109/9780470544037.ch14}.

\bibitem[Hodosh and Hockenmaier(2016)]{Hodosh2016}
Micah Hodosh and Julia Hockenmaier.
\newblock Focused evaluation for image description with binary forced-choice
  tasks.
\newblock In \emph{VL}, pages 19--28, Berlin, Germany, August 2016. Association
  for Computational Linguistics.
\newblock \doi{10.18653/v1/W16-3203}.
\newblock URL \url{https://www.aclweb.org/anthology/W16-3203}.

\bibitem[Hodosh et~al.(2013)Hodosh, Young, and Hockenmaier]{Hodosh2013}
Micah Hodosh, Peter Young, and Julia Hockenmaier.
\newblock {Framing} {Image} {Description} as a {Ranking} {Task}: {Data},
  {Models} and {Evaluation} {Metrics}.
\newblock \emph{JAIR}, 47\penalty0 (1):\penalty0 853--899, 2013.
\newblock \doi{10.1109/cvprw.2013.51}.

\bibitem[Hornik et~al.(1989)Hornik, Stinchcombe, and White]{Hornik1989}
Kurt Hornik, Maxwell Stinchcombe, and Halbert White.
\newblock Multilayer feedforward networks are universal approximators.
\newblock \emph{Neural Networks}, 2\penalty0 (5):\penalty0 359--366, January
  1989.
\newblock \doi{10.1016/0893-6080(89)90020-8}.

\bibitem[Hossain et~al.(2019)Hossain, Sohel, Shiratuddin, and
  Laga]{Hossain2019}
M.~D.~Zakir Hossain, Ferdous Sohel, Mohd~Fairuz Shiratuddin, and Hamid Laga.
\newblock A comprehensive survey of deep learning for image captioning.
\newblock \emph{ACM Comput. Surv.}, 51\penalty0 (6):\penalty0 118:1--118:36,
  February 2019.
\newblock ISSN 0360-0300.
\newblock \doi{10.1145/3295748}.

\bibitem[Howard and Ruder(2018)]{Howard2018}
Jeremy Howard and Sebastian Ruder.
\newblock Universal language model fine-tuning for text classification.
\newblock In \emph{Proc. ACL}, pages 328--339, Melbourne, Australia, 2018.
  Association for Computational Linguistics.
\newblock URL \url{http://aclweb.org/anthology/P18-1031}.

\bibitem[Iyyer et~al.(2015)Iyyer, Manjunatha, Boyd-graber, and
  Hal~Daum{\'{e}}]{Iyyer2015}
Mohit Iyyer, Varun Manjunatha, Jordan Boyd-graber, and I.~I.~I.
  Hal~Daum{\'{e}}.
\newblock Deep unordered composition rivals syntactic methods for text
  classification.
\newblock In \emph{IJCNLP}. Association for Computational Linguistics, 2015.
\newblock \doi{10.3115/v1/p15-1162}.

\bibitem[Jaeger and Haas(2004)]{Jaeger2004}
Herbert Jaeger and Harald Haas.
\newblock Harnessing nonlinearity: Predicting chaotic systems and saving energy
  in wireless communication.
\newblock \emph{Science}, 304\penalty0 (5667):\penalty0 78--80, 2004.
\newblock ISSN 0036-8075.
\newblock \doi{10.1126/science.1091277}.
\newblock URL \url{http://science.sciencemag.org/content/304/5667/78}.

\bibitem[Johnson et~al.(2015)Johnson, Krishna, Stark, Li, Shamma, Bernstein,
  and Fei-fei]{Johnson2015}
Justin Johnson, Ranjay Krishna, Michael Stark, Li-jia Li, David~A. Shamma,
  Michael~S. Bernstein, and Li~Fei-fei.
\newblock Image retrieval using scene graphs.
\newblock In \emph{CVPR}. {IEEE}, June 2015.
\newblock \doi{10.1109/cvpr.2015.7298990}.

\bibitem[K{\'{a}}d{\'{a}}r et~al.(2017)K{\'{a}}d{\'{a}}r, Chrupa{\l}a, and
  Alishahi]{Kadar2017}
{\'{A}}kos K{\'{a}}d{\'{a}}r, Grzegorz Chrupa{\l}a, and Afra Alishahi.
\newblock Representation of linguistic form and function in recurrent neural
  networks.
\newblock \emph{Computational Linguistics}, 43\penalty0 (4):\penalty0 761--780,
  December 2017.
\newblock \doi{10.1162/coli_a_00300}.

\bibitem[Karpathy and Fei-fei(2015)]{Karpathy2015}
Andrej Karpathy and Li~Fei-fei.
\newblock {Deep} {Visual}-{Semantic} {Alignments} for {Generating} {Image}
  {Descriptions}.
\newblock In \emph{Proc. CVPR}, 2015.
\newblock \doi{10.1109/cvpr.2015.7298932}.

\bibitem[K{\i}l{\i}{\c{c}}kaya et~al.(2017)K{\i}l{\i}{\c{c}}kaya, Erdem,
  Ikizler-cinbis, and Erdem]{Kilickaya2017}
Mert K{\i}l{\i}{\c{c}}kaya, Aykut Erdem, Nazli Ikizler-cinbis, and Erkut Erdem.
\newblock Re-evaluating automatic metrics for image captioning.
\newblock In \emph{EACL}. Association for Computational Linguistics, 2017.
\newblock \doi{10.18653/v1/e17-1019}.

\bibitem[Kingma and Ba(2014)]{P.Kingma2014}
P.~Diederik Kingma and Jimmy Ba.
\newblock {Adam}: {A} {Method} for {Stochastic} {Optimization}.
\newblock \emph{CoRR}, 1412.6980, 2014.

\bibitem[Kiros et~al.(2014{\natexlab{a}})Kiros, Salakhutdinov, and
  Zemel]{Kiros2014}
Ryan Kiros, Ruslan Salakhutdinov, and Richard~S. Zemel.
\newblock {Unifying} visual-semantic embeddings with multimodal neural language
  models.
\newblock \emph{CoRR}, 1411.2539, 2014{\natexlab{a}}.

\bibitem[Kiros et~al.(2014{\natexlab{b}})Kiros, Salakhutdinov, and
  Zemel]{Kiros2014a}
Ryan Kiros, Ruslan Salakhutdinov, and Richard~S. Zemel.
\newblock Multimodal neural language models.
\newblock In \emph{Proc. ICML}, pages 595--603, 2014{\natexlab{b}}.

\bibitem[Kornblith et~al.(2018)Kornblith, Shlens, and Le]{Kornblith2018}
Simon Kornblith, Jonathon Shlens, and Quoc~V. Le.
\newblock Do better imagenet models transfer better?
\newblock \emph{CoRR}, Abs/1805.08974, 2018.
\newblock URL \url{http://arxiv.org/abs/1805.08974}.

\bibitem[Krause et~al.(2016)Krause, Johnson, Krishna, and Fei-fei]{Krause2016}
Jonathan Krause, Justin Johnson, Ranjay Krishna, and Li~Fei-fei.
\newblock A hierarchical approach for generating descriptive image paragraphs.
\newblock \emph{CoRR}, 1611.06607, 2016.

\bibitem[Krizhevsky et~al.(2012)Krizhevsky, Sutskever, and
  Hinton]{Krizhevsky2012}
Alex Krizhevsky, Ilya Sutskever, and Geoffrey~E. Hinton.
\newblock Imagenet classification with deep convolutional neural networks.
\newblock In F.~Pereira, C.~J.~C. Burges, L.~Bottou, and K.~Q. Weinberger,
  editors, \emph{NIPS}, pages 1097--1105. Curran Associates, Inc., 2012.
\newblock URL
  \url{http://papers.nips.cc/paper/4824-imagenet-classification-with-deep-convolutional-neural-networks.pdf}.

\bibitem[Kulkarni et~al.(2011)Kulkarni, Premraj, Dhar, Li, Choi, Berg, and
  Berg]{Kulkarni2011}
Girish Kulkarni, Visruth Premraj, Sagnik Dhar, Siming Li, Yejin Choi,
  Alexander~C. Berg, and Tamara~L. Berg.
\newblock Baby talk: Understanding and generating simple image descriptions.
\newblock In \emph{CVPR}. {IEEE}, June 2011.
\newblock \doi{10.1109/cvpr.2011.5995466}.

\bibitem[Kusner et~al.(2015)Kusner, Sun, Kolkin, and Weinberger]{Kusner2015}
Matt Kusner, Yu~Sun, Nicholas Kolkin, and Kilian Weinberger.
\newblock From word embeddings to document distances.
\newblock In Francis Bach and David Blei, editors, \emph{ICML}, volume~37 of
  \emph{Proceedings of Machine Learning Research}, pages 957--966, Lille,
  France, July 2015. Pmlr.
\newblock URL \url{http://proceedings.mlr.press/v37/kusnerb15.html}.

\bibitem[Kuznetsova et~al.(2014)Kuznetsova, Ordonez, Berg, and
  Choi]{Kuznetsova2014}
Polina Kuznetsova, Vicente Ordonez, Tamara Berg, and Yejin Choi.
\newblock Treetalk: Composition and compression of trees for image
  descriptions.
\newblock \emph{TACL}, 2:\penalty0 351--362, 2014.
\newblock ISSN 2307-387X.
\newblock URL \url{https://transacl.org/ojs/index.php/tacl/article/view/367}.

\bibitem[Lai(2018)]{Lai2018}
Alice~Yingming Lai.
\newblock \emph{{Textual entailment from image caption denotation}}.
\newblock Phd thesis, University of Illinois at Urbana-champaign, 2018.

\bibitem[Lamb et~al.(2016)Lamb, Goyal, Zhang, Zhang, Courville, and
  Bengio]{Lamb2016}
Alex~M. Lamb, Anirudh Goyal, Ying Zhang, Saizheng Zhang, Aaron~C. Courville,
  and Yoshua Bengio.
\newblock Professor forcing: A new algorithm for training recurrent networks.
\newblock In D.~D. Lee, M.~Sugiyama, U.~V. Luxburg, I.~Guyon, and R.~Garnett,
  editors, \emph{NIPS}, pages 4601--4609. Curran Associates, Inc., 2016.
\newblock URL
  \url{http://papers.nips.cc/paper/6099-professor-forcing-a-new-algorithm-for-training-recurrent-networks.pdf}.

\bibitem[Le et~al.(2015)Le, Jaitly, and Hinton]{Le2015}
Quoc~V. Le, Navdeep Jaitly, and Geoffrey~E. Hinton.
\newblock A simple way to initialize recurrent networks of rectified linear
  units.
\newblock \emph{CoRR}, April 2015.

\bibitem[LeCun et~al.(1998)LeCun, Bottou, Bengio, and Haffner]{Lecun1998}
Y.~LeCun, L.~Bottou, Y.~Bengio, and P.~Haffner.
\newblock Gradient-based learning applied to document recognition.
\newblock \emph{Proceedings of the IEEE}, 86\penalty0 (11):\penalty0
  2278--2324, 1998.
\newblock \doi{10.1109/5.726791}.

\bibitem[LeCun et~al.(2015)LeCun, Bengio, and Hinton]{LeCun2015}
Yann LeCun, Yoshua Bengio, and Geoffrey Hinton.
\newblock Deep learning.
\newblock \emph{Nature}, 521\penalty0 (7553):\penalty0 436--444, May 2015.
\newblock \doi{10.1038/nature14539}.

\bibitem[Lin and Och(2004)]{Lin2004}
Chin-yew Lin and Franz~Josef Och.
\newblock Automatic evaluation of machine translation quality using longest
  common subsequence and skip-bigram statistics.
\newblock In \emph{Proc. ACL}, 2004.

\bibitem[Lin et~al.(2014)Lin, Maire, Belongie, Hays, Perona, Ramanan,
  Doll{\'{a}}r, and Zitnick]{Lin2014}
Tsung-yi Lin, Michael Maire, Serge Belongie, James Hays, Pietro Perona, Deva
  Ramanan, Piotr Doll{\'{a}}r, and C.~Lawrence Zitnick.
\newblock Microsoft {COCO}: Common objects in context.
\newblock In \emph{Proc. ECCV}, pages 740--755, 2014.
\newblock \doi{10.1007/978-3-319-10602-1_48}.

\bibitem[Liu et~al.(2017)Liu, Sun, Wang, Wang, and Yuille]{Liu2017}
Chang Liu, Fuchun Sun, Changhu Wang, Feng Wang, and Alan Yuille.
\newblock Mat: A multimodal attentive translator for image captioning.
\newblock In \emph{IJCAI}, pages 4033--4039, 2017.
\newblock \doi{10.24963/ijcai.2017/563}.

\bibitem[Liu et~al.(2016)Liu, Zhu, Ye, Guadarrama, and Murphy]{Liu2016}
Siqi Liu, Zhenhai Zhu, Ning Ye, Sergio Guadarrama, and Kevin Murphy.
\newblock Optimization of image description metrics using policy gradient
  methods.
\newblock \emph{CoRR}, 1612.00370, 2016.

\bibitem[Lu et~al.(2017)Lu, Xiong, Parikh, and Socher]{Lu2016}
Jiasen Lu, Caiming Xiong, Devi Parikh, and Richard Socher.
\newblock Knowing when to look: Adaptive attention via a visual sentinel for
  image captioning.
\newblock In \emph{CVPR}, pages 3242--3250, 2017.
\newblock \doi{10.1109/CVPR.2017.345}.

\bibitem[Ma and Han(2016)]{Ma2016}
Shubo Ma and Yahong Han.
\newblock Describing images by feeding {LSTM} with structural words.
\newblock In \emph{Proc. ICME}, 2016.

\bibitem[Madhysastha et~al.(2018)Madhysastha, Wang, and
  Specia]{Madhysastha2018}
Pranava Madhysastha, Josiah Wang, and Lucia Specia.
\newblock The role of image representations in vision to language tasks.
\newblock \emph{NLE}, 24\penalty0 (3):\penalty0 415--439, 2018.
\newblock \doi{10.1017/S1351324918000116}.

\bibitem[Mansimov et~al.(2016)Mansimov, Parisotto, Ba, and
  Salakhutdinov]{Mansimov2016}
Elman Mansimov, Emilio Parisotto, Lei~Jimmy Ba, and Ruslan Salakhutdinov.
\newblock Generating images from captions with attention.
\newblock In \emph{ICLR}, San Juan, Puerto Rico, May 2016.
\newblock URL \url{http://arxiv.org/abs/1511.02793}.

\bibitem[Mao et~al.(2014)Mao, Xu, Yang, Wang, and Yuille]{Mao2014}
Junhua Mao, Wei Xu, Yi~Yang, Jiang Wang, and Alan~L. Yuille.
\newblock {Explain} images with multimodal recurrent neural networks.
\newblock \emph{Proc. NIPS}, 2014.

\bibitem[Mao et~al.(2015{\natexlab{a}})Mao, Xu, Yang, Wang, Huang, and
  Yuille]{Mao2015}
Junhua Mao, Wei Xu, Yi~Yang, Jiang Wang, Zhiheng Huang, and Alan Yuille.
\newblock {Deep} {Captioning} with {Multimodal} {Recurrent} {Neural} {Networks}
  (m-{RNN}).
\newblock \emph{Proc. ICLR}, 2015{\natexlab{a}}.

\bibitem[Mao et~al.(2015{\natexlab{b}})Mao, Xu, Yang, Wang, Huang, and
  Yuille]{Mao2015a}
Junhua Mao, Wei Xu, Yi~Yang, Jiang Wang, Zhiheng Huang, and Alan Yuille.
\newblock {Learning} like a {Child}: {Fast} {Novel} {Visual} {Concept}
  {Learning} from {Sentence} {Descriptions} of {Images}.
\newblock In \emph{Proc. ICCV}, 2015{\natexlab{b}}.

\bibitem[Mason and Charniak(2014)]{Mason2014}
Rebecca Mason and Eugene Charniak.
\newblock Domain-specific image captioning.
\newblock In \emph{CoNLL}, pages 11--20, Ann Arbor, Michigan, June 2014.
  Association for Computational Linguistics.
\newblock URL \url{http://www.aclweb.org/anthology/W14-1602}.

\bibitem[Mikolov et~al.(2013)Mikolov, Chen, Corrado, and Dean]{Mikolov2013}
Tomas Mikolov, Kai Chen, Greg Corrado, and Jeffrey Dean.
\newblock {Efficient} {Estimation} of {Word} {Representations} in {Vector}
  {Space}.
\newblock \emph{CoRR}, 1301.3781, 2013.

\bibitem[Mitchell et~al.(2012)Mitchell, Dodge, Goyal, Yamaguchi, Stratos, Han,
  Mensch, Berg, Berg, and Daume~Iii]{Mitchell2012}
Margaret Mitchell, Jesse Dodge, Amit Goyal, Kota Yamaguchi, Karl Stratos,
  Xufeng Han, Alyssa Mensch, Alex Berg, Tamara Berg, and Hal Daume~Iii.
\newblock Midge: Generating image descriptions from computer vision detections.
\newblock In \emph{EACL}, pages 747--756, Avignon, France, April 2012.
  Association for Computational Linguistics.
\newblock URL \url{http://www.aclweb.org/anthology/E12-1076}.

\bibitem[Mnih and Hinton(2007)]{Mnih2007}
Andriy Mnih and Geoffrey Hinton.
\newblock Three new graphical models for statistical language modelling.
\newblock In \emph{ICML}, pages 641--648, 2007.

\bibitem[Mou et~al.(2016)Mou, Meng, Yan, Li, Xu, Zhang, and Jin]{Mou2016}
Lili Mou, Zhao Meng, Rui Yan, Ge~Li, Yan Xu, Lu~Zhang, and Zhi Jin.
\newblock How transferable are neural networks in nlp applications?
\newblock In \emph{EMNLP}, pages 479--489, Austin, Texas, November 2016.
  Association for Computational Linguistics.
\newblock URL \url{https://aclweb.org/anthology/D16-1046}.

\bibitem[Nallapati et~al.(2016)Nallapati, Zhou, Dos~Santos, Gulcehre, and
  Xiang]{Nallapati2016}
Ramesh Nallapati, Bowen Zhou, Cicero Dos~Santos, Caglar Gulcehre, and Bing
  Xiang.
\newblock Abstractive text summarization using sequence-to-sequence rnns and
  beyond.
\newblock In \emph{CoNLL}, pages 280--290, Berlin, Germany, August 2016. Acl.

\bibitem[Nina and Rodriguez(2015)]{Nina2015}
Oliver Nina and Andres Rodriguez.
\newblock Simplified {LSTM} unit and search space probability exploration for
  image description.
\newblock In \emph{Proc. ICICS}, 2015.

\bibitem[Ordonez et~al.(2011)Ordonez, Kulkarni, and Berg]{Ordonez2011}
Vicente Ordonez, Girish Kulkarni, and Tamara~L. Berg.
\newblock Im2text: Describing images using 1 million captioned photographs.
\newblock In J.~Shawe-taylor, R.~S. Zemel, P.~L. Bartlett, F.~Pereira, and
  K.~Q. Weinberger, editors, \emph{NIPS}, pages 1143--1151. Curran Associates,
  Inc., 2011.
\newblock URL
  \url{http://papers.nips.cc/paper/4470-im2text-describing-images-using-1-million-captioned-photographs.pdf}.

\bibitem[Oruganti et~al.(2016)Oruganti, Sah, Pillai, and Ptucha]{Oruganti2016}
Ram~Manohar Oruganti, Shagan Sah, Suhas Pillai, and Raymond Ptucha.
\newblock Image description through fusion based recurrent multi-modal
  learning.
\newblock In \emph{Proc. ICIP}, 2016.

\bibitem[Pan and Yang(2010)]{Pan2010}
S.~J. Pan and Q.~Yang.
\newblock A survey on transfer learning.
\newblock \emph{TKDE}, 22\penalty0 (10):\penalty0 1345--1359, October 2010.
\newblock ISSN 1041-4347.
\newblock \doi{10.1109/TKDE.2009.191}.

\bibitem[Papineni et~al.(2002)Papineni, Roukos, Ward, and Zhu]{Papineni2002}
Kishore Papineni, Salim Roukos, Todd Ward, and Wei-jing Zhu.
\newblock {BLEU}: {A} method for automatic evaluation of machine translation.
\newblock In \emph{Proc. ACL}, pages 311--318, 2002.

\bibitem[Pascanu et~al.(2013)Pascanu, Mikolov, and Bengio]{Pascanu2013}
Razvan Pascanu, Tomas Mikolov, and Yoshua Bengio.
\newblock On the difficulty of training recurrent neural networks.
\newblock In Sanjoy Dasgupta and David Mcallester, editors, \emph{ICML},
  volume~28 of \emph{Proceedings of Machine Learning Research}, pages
  1310--1318, Atlanta, Georgia, Usa, June 2013. Pmlr.
\newblock URL \url{http://proceedings.mlr.press/v28/pascanu13.html}.

\bibitem[Pennington et~al.(2014)Pennington, Socher, and
  Manning]{Pennington2014}
Jeffrey Pennington, Richard Socher, and Christopher Manning.
\newblock Glove: Global vectors for word representation.
\newblock In \emph{EMNLP}. Association for Computational Linguistics, 2014.
\newblock \doi{10.3115/v1/d14-1162}.

\bibitem[Ramachandran et~al.(2017)Ramachandran, Liu, and Le]{Ramachandran2017}
Prajit Ramachandran, Peter Liu, and Quoc Le.
\newblock Unsupervised pretraining for sequence to sequence learning.
\newblock In \emph{EMNLP}, pages 383--391, Copenhagen, Denmark, September 2017.
  Association for Computational Linguistics.
\newblock URL \url{https://www.aclweb.org/anthology/D17-1039}.

\bibitem[Reiter and Belz(2009)]{Reiter2009}
Ehud Reiter and Anja Belz.
\newblock An investigation into the validity of some metrics for automatically
  evaluating natural language generation systems.
\newblock \emph{Computational Linguistics}, 35\penalty0 (4):\penalty0 529--558,
  2009.
\newblock \doi{10.1162/coli.2009.35.4.35405}.

\bibitem[Rennie et~al.(2017)Rennie, Marcheret, Mroueh, Ross, and
  Goel]{Rennie2017}
Steven~J. Rennie, Etienne Marcheret, Youssef Mroueh, Jarret Ross, and Vaibhava
  Goel.
\newblock Self-critical sequence training for image captioning.
\newblock In \emph{CVPR}, pages 1179--1195, Honolulu, HI, USA, July 2017.
\newblock \doi{10.1109/CVPR.2017.131}.

\bibitem[Roy(2005)]{Roy2005}
Deb Roy.
\newblock Semiotic schemas: A framework for grounding language in action and
  perception.
\newblock \emph{Artificial Intelligence}, 167\penalty0 (1-2):\penalty0
  170--205, 2005.

\bibitem[Rumelhart et~al.(1986)Rumelhart, Hinton, and Williams]{Rumelhart1986}
David~E. Rumelhart, Geoffrey~E. Hinton, and Ronald~J. Williams.
\newblock Learning representations by back-propagating errors.
\newblock \emph{Nature}, 323\penalty0 (6088):\penalty0 533--536, October 1986.
\newblock \doi{10.1038/323533a0}.

\bibitem[Russakovsky et~al.(2015)Russakovsky, Deng, Su, Krause, Satheesh, Ma,
  Huang, Karpathy, Khosla, Bernstein, Berg, and Fei-fei]{Russakovsky2015}
Olga Russakovsky, Jia Deng, Hao Su, Jonathan Krause, Sanjeev Satheesh, Sean Ma,
  Zhiheng Huang, Andrej Karpathy, Aditya Khosla, Michael Bernstein,
  Alexander~C. Berg, and Li~Fei-fei.
\newblock {ImageNet Large Scale Visual Recognition Challenge}.
\newblock \emph{IJCV}, 115\penalty0 (3):\penalty0 211--252, 2015.
\newblock \doi{10.1007/s11263-015-0816-y}.

\bibitem[Samek et~al.(2018)Samek, Wiegand, and M{\"u}ller]{Samek2018}
Wojciech Samek, Thomas Wiegand, and Klaus-robert M{\"u}ller.
\newblock Explainable artificial intelligence: Understanding, visualizing and
  interpreting deep learning models.
\newblock \emph{ITU Journal: ICT Discoveries}, 1\penalty0 (1):\penalty0 39--48,
  2018.
\newblock URL \url{https://www.itu.int/en/journal/001/Pages/05.aspx}.

\bibitem[Shekhar et~al.(2017)Shekhar, Pezzelle, Klimovich, Herbelot, Nabi,
  Sangineto, and Bernardi]{Shekhar2017}
Ravi Shekhar, Sandro Pezzelle, Yauhen Klimovich, Aur\'{e}lie Herbelot, Moin
  Nabi, Enver Sangineto, and Raffaella Bernardi.
\newblock Foil it! find one mismatch between image and language caption.
\newblock In \emph{Proc. ACL}, pages 255--265, Vancouver, Canada, July 2017.
  Association for Computational Linguistics.
\newblock URL \url{http://aclweb.org/anthology/P17-1024}.

\bibitem[Siegelmann and Sontag(1995)]{Siegelmann1995}
H.~T. Siegelmann and E.~D. Sontag.
\newblock On the computational power of neural nets.
\newblock \emph{Journal of Computer and System Sciences}, 50\penalty0
  (1):\penalty0 132--150, February 1995.
\newblock \doi{10.1006/jcss.1995.1013}.

\bibitem[Simonyan and Zisserman(2014)]{Simonyan2014}
Karen Simonyan and Andrew Zisserman.
\newblock {Very} {Deep} {Convolutional} {Networks} for {Large}-{Scale} {Image}
  {Recognition}.
\newblock \emph{CoRR}, 1409.1556, 2014.

\bibitem[Socher et~al.(2014)Socher, Karpathy, Le, Manning, and Ng]{Socher2014}
Richard Socher, Andrej Karpathy, Quoc~V. Le, Christopher~D. Manning, and
  Andrew~Y. Ng.
\newblock Grounded compositional semantics for finding and describing images
  with sentences.
\newblock \emph{TACL}, 2:\penalty0 207--218, 2014.

\bibitem[Song and Yoo(2016)]{Song2016}
Mingoo Song and Chang~D. Yoo.
\newblock Multimodal representation: Kneser-{N}ey smoothing/skip-gram based
  neural language model.
\newblock In \emph{Proc. ICIP}, 2016.

\bibitem[Specia et~al.(2016)Specia, Frank, Sima{'}an, and Elliott]{Specia2016}
Lucia Specia, Stella Frank, Khalil Sima{'}an, and Desmond Elliott.
\newblock A shared task on multimodal machine translation and crosslingual
  image description.
\newblock In \emph{WMT}, pages 543--553, Berlin, Germany, August 2016.
  Association for Computational Linguistics.
\newblock \doi{10.18653/v1/W16-2346}.
\newblock URL \url{https://www.aclweb.org/anthology/W16-2346}.

\bibitem[Srivastava et~al.(2014)Srivastava, Hinton, Krizhevsky, Sutskever, and
  Salakhutdinov]{Srivastava2014}
Nitish Srivastava, Geoffrey Hinton, Alex Krizhevsky, Ilya Sutskever, and Ruslan
  Salakhutdinov.
\newblock Dropout: A simple way to prevent neural networks from overfitting.
\newblock \emph{JMLR}, 15:\penalty0 1929--1958, 2014.
\newblock URL \url{http://jmlr.org/papers/v15/srivastava14a.html}.

\bibitem[Sukhbaatar et~al.(2015)Sukhbaatar, Szlam, Weston, and
  Fergus]{Sukhbaatar2015}
Sainbayar Sukhbaatar, Arthur Szlam, Jason Weston, and Rob Fergus.
\newblock End-to-end memory networks.
\newblock In C.~Cortes, N.~D. Lawrence, D.~D. Lee, M.~Sugiyama, and R.~Garnett,
  editors, \emph{NIPS}, pages 2440--2448. Curran Associates, Inc., 2015.
\newblock URL
  \url{http://papers.nips.cc/paper/5846-end-to-end-memory-networks.pdf}.

\bibitem[Sutskever et~al.(2014)Sutskever, Vinyals, and Le]{Sutskever2014}
Ilya Sutskever, Oriol Vinyals, and Quoc~V. Le.
\newblock Sequence to sequence learning with neural networks.
\newblock In Z.~Ghahramani, M.~Welling, C.~Cortes, N.~D. Lawrence, and K.~Q.
  Weinberger, editors, \emph{NIPS}, pages 3104--3112. Curran Associates, Inc.,
  2014.

\bibitem[Szegedy et~al.(2015)Szegedy, Liu, Jia, Sermanet, Reed, Anguelov,
  Erhan, Vanhoucke, and Rabinovich]{Szegedy2015}
Christian Szegedy, Wei Liu, Yangqing Jia, Pierre Sermanet, Scott Reed, Dragomir
  Anguelov, Dumitru Erhan, Vincent Vanhoucke, and Andrew Rabinovich.
\newblock Going deeper with convolutions.
\newblock In \emph{CVPR}. {IEEE}, June 2015.
\newblock \doi{10.1109/cvpr.2015.7298594}.

\bibitem[Tanti et~al.(2017)Tanti, Gatt, and Camilleri]{Tanti2017}
Marc Tanti, Albert Gatt, and Kenneth Camilleri.
\newblock What is the role of recurrent neural networks (rnns) in an image
  caption generator?
\newblock In \emph{INLG}, pages 51--60, Santiago De Compostela, Spain,
  September 2017. Association for Computational Linguistics.
\newblock URL \url{http://www.aclweb.org/anthology/W17-3506}.

\bibitem[Tanti et~al.(2018)Tanti, Gatt, and Camilleri]{Tanti2018}
Marc Tanti, Albert Gatt, and Kenneth~P. Camilleri.
\newblock Where to put the image in an image caption generator.
\newblock \emph{NLE}, 24\penalty0 (3):\penalty0 467--489, April 2018.
\newblock \doi{10.1017/S1351324918000098}.
\newblock URL
  \url{https://www.cambridge.org/core/journals/natural-language-engineering/article/where-to-put-the-image-in-an-image-caption-generator/A5B0ACFFE8E4AEAA5840DC61F93153F3\#fndtn-information}.

\bibitem[Tanti et~al.(2019{\natexlab{a}})Tanti, Gatt, and Camilleri]{Tanti2019}
Marc Tanti, Albert Gatt, and Kenneth~P. Camilleri.
\newblock Transfer learning from language models to image caption generators:
  Better models may not transfer better.
\newblock \emph{CoRR}, January 2019{\natexlab{a}}.
\newblock URL \url{https://arxiv.org/abs/1901.01216}.

\bibitem[Tanti et~al.(2019{\natexlab{b}})Tanti, Gatt, and
  Camilleri]{Tanti2019b}
Marc Tanti, Albert Gatt, and Kenneth~P. Camilleri.
\newblock Quantifying the amount of visual information used by neural caption
  generators.
\newblock In Laura Leal-taix{\'e} and Stefan Roth, editors, \emph{ECCV}, pages
  124--132. Springer International Publishing, 2019{\natexlab{b}}.
\newblock ISBN 978-3-030-11018-5.
\newblock \doi{10.1007/978-3-030-11018-5_11}.

\bibitem[Tanti et~al.(2019{\natexlab{c}})Tanti, Gatt, and Muscat]{Tanti2019a}
Marc Tanti, Albert Gatt, and Adrian Muscat.
\newblock Pre-gen metrics: Predicting caption quality metrics without
  generating captions.
\newblock In Laura Leal-taix{\'e} and Stefan Roth, editors, \emph{ECCV}, pages
  114--123. Springer International Publishing, 2019{\natexlab{c}}.
\newblock ISBN 978-3-030-11018-5.
\newblock \doi{10.1007/978-3-030-11018-5_10}.

\bibitem[Vaswani et~al.(2017)Vaswani, Shazeer, Parmar, Uszkoreit, Jones, Gomez,
  Kaiser, and Polosukhin]{Vaswani2017}
Ashish Vaswani, Noam Shazeer, Niki Parmar, Jakob Uszkoreit, Llion Jones,
  Aidan~N. Gomez, \l.~Ukasz Kaiser, and Illia Polosukhin.
\newblock Attention is all you need.
\newblock In I.~Guyon, U.~V. Luxburg, S.~Bengio, H.~Wallach, R.~Fergus,
  S.~Vishwanathan, and R.~Garnett, editors, \emph{NIPS}, pages 5998--6008.
  Curran Associates, Inc., 2017.
\newblock URL
  \url{http://papers.nips.cc/paper/7181-attention-is-all-you-need.pdf}.

\bibitem[Vedantam et~al.(2015)Vedantam, Zitnick, and Parikh]{Vedantam2015}
Ramakrishna Vedantam, C.~Lawrence Zitnick, and Devi Parikh.
\newblock {CIDEr}: Consensus-based image description evaluation.
\newblock In \emph{Proc. CVPR}, 2015.

\bibitem[Venugopalan et~al.(2017)Venugopalan, Hendricks, Rohrbach, Mooney,
  Darrell, and Saenko]{Venugopalan2017}
Subhashini Venugopalan, Lisa~Anne Hendricks, Marcus Rohrbach, Raymond Mooney,
  Trevor Darrell, and Kate Saenko.
\newblock Captioning images with diverse objects.
\newblock In \emph{CVPR}, pages 5753--5761, 2017.
\newblock URL
  \url{http://www.cs.utexas.edu/users/ai-lab/pub-view.php?PubID=127614}.

\bibitem[Vinyals et~al.(2015)Vinyals, Toshev, Bengio, and Erhan]{Vinyals2015}
Oriol Vinyals, Alexander Toshev, Samy Bengio, and Dumitru Erhan.
\newblock {Show} and tell: {A} neural image caption generator.
\newblock In \emph{Proc. CVPR}, 2015.

\bibitem[Vinyals et~al.(2017)Vinyals, Toshev, Bengio, and Erhan]{Vinyals2017}
Oriol Vinyals, Alexander Toshev, Samy Bengio, and Dumitru Erhan.
\newblock Show and tell: Lessons learned from the 2015 {MSCOCO} image
  captioning challenge.
\newblock \emph{TPAMI}, 39\penalty0 (4):\penalty0 652--663, April 2017.
\newblock \doi{10.1109/tpami.2016.2587640}.

\bibitem[Vu et~al.(2018)Vu, Greco, Erofeeva, Jafaritazehjan, Linders, Tanti,
  Testoni, Bernardi, and Gatt]{Vu2018}
Hoa Vu, Claudio Greco, Aliia Erofeeva, Somayeh Jafaritazehjan, Guido Linders,
  Marc Tanti, Alberto Testoni, Raffaella Bernardi, and Albert Gatt.
\newblock Grounded textual entailment.
\newblock In \emph{COLING}, pages 2354--2368, Santa Fe, New Mexico, Usa, August
  2018. Association for Computational Linguistics.
\newblock URL \url{https://www.aclweb.org/anthology/C18-1199}.

\bibitem[Wang et~al.(2016{\natexlab{a}})Wang, Yang, Bartz, and
  Meinel]{Wang2016a}
Cheng Wang, Haojin Yang, Christian Bartz, and Christoph Meinel.
\newblock Image captioning with deep bidirectional {LSTMs}.
\newblock In \emph{ACMMM}. {ACM} Press, 2016{\natexlab{a}}.
\newblock \doi{10.1145/2964284.2964299}.

\bibitem[Wang et~al.(2018)Wang, Madhyastha, and Specia]{Wang2018}
Josiah Wang, Pranava~Swaroop Madhyastha, and Lucia Specia.
\newblock Object counts! bringing explicit detections back into image
  captioning.
\newblock In \emph{NAACL}, pages 2180--2193, New Orleans, Louisiana, June 2018.
  Association for Computational Linguistics.
\newblock \doi{10.18653/v1/N18-1198}.
\newblock URL \url{https://www.aclweb.org/anthology/N18-1198}.

\bibitem[Wang et~al.(2016{\natexlab{b}})Wang, Song, Yang, and Luo]{Wang2016}
Minsi Wang, Li~Song, Xiaokang Yang, and Chuanfei Luo.
\newblock A parallel-fusion {RNN}-{LSTM} architecture for image caption
  generation.
\newblock In \emph{Proc. ICIP}, 2016{\natexlab{b}}.

\bibitem[Werbos(1990)]{Werbos1990}
P.~J. Werbos.
\newblock Backpropagation through time: What it does and how to do it.
\newblock \emph{Proceedings of the IEEE}, 78\penalty0 (10):\penalty0
  1550--1560, 1990.
\newblock \doi{10.1109/5.58337}.

\bibitem[Williams and Zipser(1989)]{Williams1989}
R.~J. Williams and D.~Zipser.
\newblock A learning algorithm for continually running fully recurrent neural
  networks.
\newblock \emph{Neural Computation}, 1\penalty0 (2):\penalty0 270--280, June
  1989.
\newblock ISSN 0899-7667.
\newblock \doi{10.1162/neco.1989.1.2.270}.

\bibitem[Wu et~al.(2015)Wu, Shen, Van Den~Hengel, Liu, and Dick]{Wu2015}
Qi~Wu, Chunhua Shen, Anton Van Den~Hengel, Lingqiao Liu, and Anthony~R. Dick.
\newblock Image captioning with an intermediate attributes layer.
\newblock \emph{CoRR}, 1506.01144, 2015.

\bibitem[Xie et~al.(2019)Xie, Lai, Doran, and Kadav]{Xie2019}
Ning Xie, Farley Lai, Derek Doran, and Asim Kadav.
\newblock {Visual Entailment: A Novel Task for Fine-Grained Image
  Understanding}.
\newblock \emph{CoRR}, 1901.06706v1, 2019.

\bibitem[Xu et~al.(2015)Xu, Ba, Kiros, Cho, Courville, Salakhutdinov, Zemel,
  and Bengio]{Xu2015}
Kelvin Xu, Jimmy Ba, Ryan Kiros, Kyunghyun Cho, Aaron~C. Courville, Ruslan
  Salakhutdinov, Richard~S. Zemel, and Yoshua Bengio.
\newblock {Show}, {Attend} and {Tell}: {Neural} {Image} {Caption} {Generation}
  with {Visual} {Attention}.
\newblock In \emph{Proc. ICML}, 2015.

\bibitem[Xu et~al.(2017)Xu, Zhang, Huang, Zhang, Gan, Huang, and He]{Xu2017}
Tao Xu, Pengchuan Zhang, Qiuyuan Huang, Han Zhang, Zhe Gan, Xiaolei Huang, and
  Xiaodong He.
\newblock Attngan: Fine-grained text to image generation with attentional
  generative adversarial networks.
\newblock \emph{CoRR}, Abs/1711.10485, 2017.
\newblock URL \url{http://arxiv.org/abs/1711.10485}.

\bibitem[Yao et~al.(2017)Yao, Pan, Li, Qiu, and Mei]{Yao2017}
Ting Yao, Yingwei Pan, Yehao Li, Zhaofan Qiu, and Tao Mei.
\newblock Boosting image captioning with attributes.
\newblock In \emph{ICCV}, pages 4904--4912, Venice, Italy, October 2017.
\newblock \doi{10.1109/ICCV.2017.524}.

\bibitem[Yosinski et~al.(2014)Yosinski, Clune, Bengio, and
  Lipson]{Yosinski2014}
Jason Yosinski, Jeff Clune, Yoshua Bengio, and Hod Lipson.
\newblock How transferable are features in deep neural networks?
\newblock In Z.~Ghahramani, M.~Welling, C.~Cortes, N.~D. Lawrence, and K.~Q.
  Weinberger, editors, \emph{NIPS}, pages 3320--3328. Curran Associates, Inc.,
  2014.
\newblock URL
  \url{http://papers.nips.cc/paper/5347-how-transferable-are-features-in-deep-neural-networks.pdf}.

\bibitem[You et~al.(2016)You, Jin, Wang, Fang, and Luo]{You2016}
Quanzeng You, Hailin Jin, Zhaowen Wang, Chen Fang, and Jiebo Luo.
\newblock Image captioning with semantic attention.
\newblock In \emph{Proc. CVPR}, 2016.

\bibitem[Young et~al.(2014)Young, Lai, Hodosh, and Hockenmaier]{Young2014}
Peter Young, Alice Lai, Micah Hodosh, and Julia Hockenmaier.
\newblock From image descriptions to visual denotations: {New} similarity
  metrics for semantic inference over event descriptions.
\newblock \emph{TACL}, 2:\penalty0 67--78, 2014.

\bibitem[Zeiler(2012)]{Zeiler2012}
Matthew~D. Zeiler.
\newblock Adadelta: An adaptive learning rate method.
\newblock \emph{CoRR}, December 2012.

\bibitem[Zhang et~al.(2016{\natexlab{a}})Zhang, Bengio, Hardt, Recht, and
  Vinyals]{Zhang2016}
Chiyuan Zhang, Samy Bengio, Moritz Hardt, Benjamin Recht, and Oriol Vinyals.
\newblock Understanding deep learning requires rethinking generalization.
\newblock \emph{CoRR}, Abs/1611.03530, 2016{\natexlab{a}}.
\newblock URL \url{http://arxiv.org/abs/1611.03530}.

\bibitem[Zhang et~al.(2016{\natexlab{b}})Zhang, Lu, and Lapata]{Zhang2016a}
Xingxing Zhang, Liang Lu, and Mirella Lapata.
\newblock Top-down tree long short-term memory networks.
\newblock In \emph{NAACL}. Association for Computational Linguistics,
  2016{\natexlab{b}}.
\newblock \doi{10.18653/v1/n16-1035}.

\bibitem[Zhou et~al.(2016)Zhou, Xu, Koch, and Corso]{Zhou2016}
Luowei Zhou, Chenliang Xu, Parker Koch, and Jason~J. Corso.
\newblock Image caption generation with text-conditional semantic attention.
\newblock \emph{CoRR}, 1606.04621, 2016.

\bibitem[Zilly et~al.(2017)Zilly, Srivastava, Koutn\'{\i}k, and
  Schmidhuber]{Zilly2017}
Julian~Georg Zilly, Rupesh~Kumar Srivastava, Jan Koutn\'{\i}k, and J{\"u}rgen
  Schmidhuber.
\newblock Recurrent highway networks.
\newblock In Doina Precup and Yee~Whye Teh, editors, \emph{ICML}, volume~70 of
  \emph{Proceedings of Machine Learning Research}, pages 4189--4198,
  International Convention Centre, Sydney, Australia, 2017. Pmlr.
\newblock URL \url{http://proceedings.mlr.press/v70/zilly17a.html}.

\bibitem[Zoph et~al.(2016)Zoph, Yuret, May, and Knight]{Zoph2016}
Barret Zoph, Deniz Yuret, Jonathan May, and Kevin Knight.
\newblock Transfer learning for low-resource neural machine translation.
\newblock In \emph{EMNLP}, pages 1568--1575, Austin, Texas, November 2016.
  Association for Computational Linguistics.
\newblock URL \url{https://aclweb.org/anthology/D16-1163}.

\end{thebibliography}
